\documentclass[twoside,11pt]{article}

\usepackage{blindtext}

\usepackage{jmlr2e}

\usepackage{lastpage}
\usepackage{tikz, tikz-cd}
\usepackage{times}
\usepackage[english]{babel}
\usepackage{subcaption}
\usepackage[mathscr]{euscript}
\usepackage{lscape}
\usepackage{indentfirst}
\usepackage{latexsym}
\usepackage{multirow}
\usepackage{enumitem}
\usepackage{epstopdf}
\usepackage{tabls}
\usepackage{wrapfig}
\usepackage{arydshln}
\usepackage{float}
\usepackage{caption}
\usepackage{booktabs} 
\usepackage{supertabular}
\usepackage{makecell}
\usepackage{algorithm}
\usepackage[noend]{algpseudocode}
\usepackage{amsthm,amsmath,amsfonts,bm,mathtools,amssymb}

\def\eqref#1{equation~\ref{#1}}

\def\1{\bm{1}}

\def\rva{{\mathbf{a}}}
\def\rvb{{\mathbf{b}}}

\def\rvz{{\mathbf{z}}}

\def\vzero{{\bm{0}}}

\def\vmu{{\bm{\mu}}}

\def\vsigma{\bm{\sigma}}
\def\vrho{\bm{\rho}}

\def\ve{{\bm{e}}}

\DeclareMathAlphabet{\mathsfit}{\encodingdefault}{\sfdefault}{m}{sl}
\SetMathAlphabet{\mathsfit}{bold}{\encodingdefault}{\sfdefault}{bx}{n}

\def\gC{{\mathcal{C}}}

\def\gG{{\mathcal{G}}}

\def\gI{{\mathcal{I}}}

\def\gN{{\mathcal{N}}}

\def\gT{{\mathcal{T}}}

\def\gX{{\mathcal{X}}}
\def\gY{{\mathcal{Y}}}
\def\gZ{{\mathcal{Z}}}

\def\sP{{\mathbb{P}}}

\def\sR{{\mathbb{R}}}

\def\tdX{{\Tilde{X}}}
\def\tdZ{{\Tilde{Z}}}
\def\tde{{\Tilde{e}}}
\def\tdg{{\Tilde{g}}}
\def\tdz{{\Tilde{z}}}
\def\tdx{{\Tilde{x}}}
\def\tdf{{\Tilde{f}}}

\newcommand{\E}{\mathbb{E}}

\newcommand{\normltwo}{{\|\cdot\|_2}}

\newcommand{\id}{\mathrm{id}}

\usepackage{pifont}
\newcommand{\cmark}{\ding{51}}%
\newcommand{\xmark}{\ding{55}}%

\newcommand{\littletaller}{\mathchoice{\vphantom{\big|}}{}{}{}}
\newcommand\restr[2]{{
  \left.\kern-\nulldelimiterspace 
  #1 
  \littletaller 
  \right|_{#2} 
  }}
  
\newcommand\corestr[2]{{
  \left.\kern-\nulldelimiterspace 
  #1 
  \right|^{\raisebox{-0.5ex}{\scriptsize $#2$}}
  }}

\DeclareMathOperator*{\argmax}{arg\,max}
\DeclareMathOperator*{\argmin}{arg\,min}

\DeclareMathOperator{\supp}{supp}

\DeclareMathOperator{\var}{Var}

\DeclareMathOperator{\rank}{rank}

\newcommand{\closure}[1]{\operatorname{cl}(#1)}
\newcommand{\interior}[1]{\operatorname{int}(#1)}

\newcommand{\mdata}{\sP_{r}}

\newcommand{\ie}{\textit{i.e.}}
\newcommand{\eg}{\textit{e.g.}}
\newcommand{\etc}{\textit{etc.}}
\newcommand{\wrt}{\textit{w.r.t.\,}}

\ShortHeadings{Least Volume Analysis}{Least Volume Analysis}
\firstpageno{1}

\begin{document}

\title{Least Volume Analysis}

\author{\name Qiuyi Chen  \email qiuyi.chen@gevernova.com \\
       \addr GE Vernova Advanced Research Center\\
       Niskayuna, NY 12309, USA
       \AND
       \name Cashen Diniz  \email cdiniz@ethz.ch \\
       \addr Department of Mechanical and Process Engineering\\
       ETH Z\"urich\\
       8092 Z\"urich, Switzerland
       \AND
       \name Mark Fuge \email mafuge@ethz.ch \\
       \addr Department of Mechanical and Process Engineering\\
       ETH Z\"urich\\
       8092 Z\"urich, Switzerland}

\editor{My editor}

\maketitle

\begin{abstract}
This paper introduces \emph{Least Volume} (LV)—a simple yet effective regularization method inspired by geometric intuition—that reduces the number of latent dimensions required by an autoencoder without prior knowledge of the dataset’s intrinsic dimensionality. We show that its effectiveness depends on the Lipschitz continuity of the decoder, prove that Principal Component Analysis (PCA) is a linear special case, and demonstrate that LV induces a PCA-like importance ordering in nonlinear models. We extend LV to non-Euclidean settings as \emph{Generalized Least Volume} (GLV), enabling the integration of label information into the latent representation. To support implementation, we also develop an accompanying \emph{Dynamic Pruning} algorithm. 
We evaluate LV on several benchmark problems, demonstrating its effectiveness in dimension reduction. Leveraging this, we reveal the role of low-dimensional latent spaces in data sampling and disentangled representation, and use them to probe the varying topological complexity of various datasets. GLV is further applied to labeled datasets, where it induces a contrastive learning effect in representations of discrete labels. On a continuous-label airfoil dataset, it produces representations that lead to smooth changes in aerodynamic performance, thereby stabilizing downstream optimization.  
\end{abstract}


\section{Introduction}
\label{sec:introduction}
Learning data representation is crucial to machine learning~\citep{bengio2013representation}. On one hand, a good representation can distill the primary features from the data samples, thus enhancing downstream tasks such as classification~\citep{krizhevsky2017imagenet, simonyan2014very, he2016deep}. On the other hand, when the data representation lies in a low-dimensional latent space $Z$ and can be mapped backward to the data samples via some decoder $g$, we can considerably facilitate generative tasks by training generative models in $Z$~\citep{ramesh2021zero, rombach2022high}. 

But what makes a data representation\textemdash\ie, $\gZ \coloneqq e(\gX)\subset Z$ of a dataset $\gX$\textemdash good?
Often, a low-dimensional $Z$ is preferred. It is frequently hypothesized that a real world dataset $\gX$ in high-dimensional data space $X$ only resides on a low-dimensional manifold~\citep{fefferman2016testing} (or at least most part of $\gX$ is locally Euclidean of low dimension). Formally speaking, we make the following definition and assumption about a dataset $\gX$: 
\begin{definition}[Dataset]
\label{def:dataset}
    Let $\sP_r$ be the \emph{probability measure of a data distribution} in a data space $X$. 
    We reserve the term \emph{dataset} (without any modifier) for the set $\gX$ of all valid data points that can be drawn from $\sP_r$, which satisfies $\closure{\gX} = \supp\sP_r$. 
\end{definition}
\begin{assumption}[Smooth Manifold Hypothesis]
    A dataset $\gX$ is a countable union of \emph{embedded submanifolds} $\gX^{i}$ with or without boundary in a smooth manifold data space $X$, \ie, ${\gX = \bigcup_i\gX^i\subseteq X}$.
    \label{asp:smooth_manifold}
\end{assumption}
\noindent If $\gX$ satisfies Assumption~\ref{asp:smooth_manifold}, then due to the \emph{rank theorem}~\cite[Theorem~4.12]{lee2012smooth}, $\gX$'s low-dimensionality will be inherited by its latent set $\gZ$ through an at least piecewise smooth and constant rank encoder $e$, which means that $\gZ$ is low-dimensional even if $Z$ is high-dimensional.

Therefore, for a $\gZ\subset Z$ retaining sufficient information about $\gX$, a low-dimensional $Z$ can provide several advantages. First, it can facilitate downstream tasks by aligning its latent dimensions more with the informative dimensions of $\gZ$ and alleviating the curse of dimensionality. In addition, it can increase the robustness of tasks such as data generation. Specifically, if a subset $U\subseteq\gZ$ constitutes an $n$-D manifold that can be embedded in an $n$-D Euclidean space, then due to the \emph{invariance of domain}~\citep[Corollary~19.9]{bredon2013topology}, $U$ is \emph{open} in $Z$ as long as $Z$ is $n$-D. Thanks to the \emph{basis criterion}~\cite[Lemma~2.10]{lee2010introduction}, this openness means for a conditional generative model trained in such $Z$, if its one prediction $\hat{z}$ is not far away from its target $z\in U$, then $\hat{z}$ will also fall inside $U\subseteq\gZ$, thus still be mapped back into $\gX$ by $g$ rather than falling outside it. Moreover, in the case where $\gZ$ is a manifold that cannot embed in similar dimensional Euclidean space, or where $\gZ$ is not even a manifold, retrieving the least dimensional $Z$ needed to embed $\gZ$ may pave the way for studying the complexity and topology of $\gZ$ and $\gX$. 

Additionally, people often hope the latent representation can indicate the importance of each dimension~\citep{rippel2014learning, pham2022pca}, so that the data variations along the principal dimensions can be easily studied, and trade-offs can be easily made when it is necessary to strip off trivial dimensions due to computational cost. This is why PCA~\citep{pearson1901liii}, despite being a century-old linear method, is still widely applied to different areas. 
However, such importance is more often than not implicitly dependent on the data space's \emph{Euclidean distance}, but this distance is sometimes misleading in evaluating the data variation's significance. For instance, in airfoil designs, under certain flow conditions, a small shape variation (\wrt the Euclidean distance over the spline parameter space) at the leading or trailing edge can lead to a significant change in its lift or drag coefficients, and we usually care more about the airfoil's performance than about its aesthetics. In scenarios alike, we have to construct data distances other than the Euclidean distance, and the importance indicator of the data representation must adapt to this change. 

To automatically learn a \emph{both} low-dimensional and importance-indicating nonlinear latent representation unrestricted to Euclidean data spaces, we introduce the \emph{Least Volume} problem, which is based on the intuition that packing a flat paper into a box consumes much less space than a curved one. This paper's contributions are:
\begin{enumerate}

\item We introduce \emph{Least Volume} (LV) regularization for autoencoders (AEs) that can compress the latent set into a low-dimensional latent subspace spanned by the latent space's standard basis. 
In addition, we transform LV into \emph{Generalized Least Volume} (GLV) that can be applied on both labeled datasets and unlabeled datasets endowed with non-Euclidean distance. 

\item We develop the \emph{Dynamic Pruning} (DP) algorithm for automatically removing dummy latent dimensions during the training of least volume autoencoders. This makes the result of LV insensitive to the latent space's initial dimension and allows more effective and thorough dimension reduction than our original work~\citep{qiuyi2024compressing}. 

\item We prove that PCA is a special case of Least Volume applied to a linear autoencoder, and show that it induces a similar importance-ordering effect based on Euclidean distance.

\item We apply Least Volume to several benchmark problems, including a synthetic dataset with known intrinsic dimensionality, MNIST, and CIFAR-10, and show that the volume penalty outperforms traditional regularizers such as Lasso in compressing the latent space. Additionally, we apply LV with Dynamic Pruning to both unlabeled datasets and datasets with labels removed, such as UIUC Airfoils, MNIST, and CelebA. Leveraging LV’s effectiveness in latent dimension reduction, we highlight the importance of minimizing latent dimensionality and provide a simple yet rigorous definition of disentangled representation derived from our findings. Finally, we analyze the topological structures of multiple datasets to reveal their distinct characteristics and complexity.

\item We apply GLV to two labeled datasets: MNIST and UIUC airfoils (with simulated aerodynamics performance values). We show that applying GLV on the categorical dataset MNIST can induce an effect similar to contrastive learning in the latent space. When applied to the UIUC dataset with continuous labels, it makes the aerodynamic performance values change less abruptly in the latent space, such that latent space become more stable for adjoint optimization.
We make the code public on GitHub\footnote{\url{https://github.com/IDEALLab/Generalized-Least-Volume}} to ensure reproducibility.
\end{enumerate}

\paragraph{Outline}
The first non-experimental part of this paper comprises five sections. Section~\ref{sec:methodology} first introduces the intuition and mathematical formulation of \emph{Least Volume} for retrieving datasets' \emph{embedding dimensions} (Def.~\ref{def:dimensions}) with continuous autoencoders. Section~\ref{sec:theory} then theoretically analyzes different components and aspects of LV, highlighting its close connection to PCA. Based on these findings, Section~\ref{sec:glv} uses isometry to extend LV to \emph{Generalized Least Volume}\textemdash applicable to labeled datasets and unlabeled datasets with non-Euclidean distances. Section~\ref{sec:implementation} provides the implementations\textemdash specifically the \emph{na\"ive volume reduction} and the superior \emph{Dynamic Pruning} algorithm\textemdash for solving the LV and GLV problems on autoencoders. Section~\ref{sec:related} reviews related works. 

The second part contains experiment results and discussions. Section~\ref{sec:experiments} empirically validates the expected properties and performance of LV. Section~\ref{sec:exp_ulva} takes advantage of LV's dimension reduction capability to compress different datasets and sheds light on the results from a topological perspective. Section~\ref{sec:exp_slva} demonstrates GLV's effect on labeled datasets with discrete and continuous labels. Section~10 summarizes the key findings in this paper and underscores the limitations.  

\section{Least Volume Problems}
\label{sec:methodology}
Let us begin with a few concepts. Mathematically, \emph{homeomorphism} in general topology~\citep[\S2.1]{lee2010introduction} charaterizes if a set $\gZ$ is a deformed replica of another set $\gX$: 
\begin{definition}[Homeomorphism]
    If $\gX$ and $\gZ$ are topological spaces, a \emph{homeomorphism} from $\gX$ to $\gZ$ is a \emph{continuous bijective map $f: \gX\to \gZ$ with continuous inverse}. We say that $\gX$ and $\gZ$ are \emph{homeomorphic} to each other (denoted by $\gX \simeq \gZ$) if there is a homeomorphism between $\gX$ and $\gZ$.
    \label{def:homeomorphism}
\end{definition}
\noindent This enables the definition of the \emph{embedding dimension} of a dataset:
\begin{definition}[Topological Embedding]
    A map $f: \gX\to Z$ is called a \emph{topological embedding} if its domain $\gX$ is homeomorphic to its image $f(\gX)\subseteq Z$. 
    \label{def:topo_embedding}
\end{definition}
\begin{definition}[Intrinsic and Embedding Dimensions]
    The \emph{intrinsic dimension} of a dataset $\gX$ satisfying Assumption~\ref{asp:smooth_manifold} is defined by $\dim\gX\coloneqq \max_i(\dim\gX^i)$. Its \emph{embedding dimension} is the dimension $m$ of the least dimensional Euclidean space $\sR^m$ to which we can establish a topological embedding $\phi: \gX \to \sR^m$.  
    Due to the \emph{topological invariance of dimension}~\citep[Theorem~13.22]{lee2010introduction}, intrinsic dimension $\leq$ embedding dimension.
    \label{def:dimensions} 
\end{definition}

This section proposes a formulation that intends to \emph{automatically reduce the latent dimension of an autoencoder to the embedding dimension of the given dataset}.
As to be proved later in Theorem~\ref{thm3:ae}, if a continuous autoencoder can reconstruct the dataset $\gX$ perfectly, then its latent set $\gZ$ is a homeomorphic copy of it. Concretely speaking, if through the lens of the manifold hypothesis we conceive the dataset as an elastic curved surface in the high dimensional data space, then the latent set can be regarded as an intact ``flattened'' version of it tucked into the low dimensional latent space by the encoder. Therefore, the task of finding the least dimensional latent space can be imagined as the continuation of this flattening process, in which we keep compressing this elastic latent surface onto latent hyperplanes of even lower dimensionality until it cannot be flattened anymore, reaching the embedding dimension of the dataset. Thereafter, we can extract the final hyperplane as the desired least dimensional latent space. Ideally, we prefer a hyperplane that is either perpendicular or parallel to each latent coordinate axis, such that we can extract it with ease. 

\subsection{Volume Minimization}
\label{sec:vol_pen}
To flatten the latent set $\gZ$ and align it with the latent coordinate axes, we want the latent code's standard deviation (STD) vector $\pmb{\sigma}(\gZ)$ to be as sparse as possible, which bespeaks the compression of the dataset onto a latent subspace of the least dimension. 
The common penalty for promoting sparsity is the $L_1$ norm. However, $\|\pmb{\sigma}\|_1$ does not necessarily lead to flattening (see \S\ref{sec:pedago}).

\begin{figure}
    \centering
    \includegraphics[width=\textwidth]{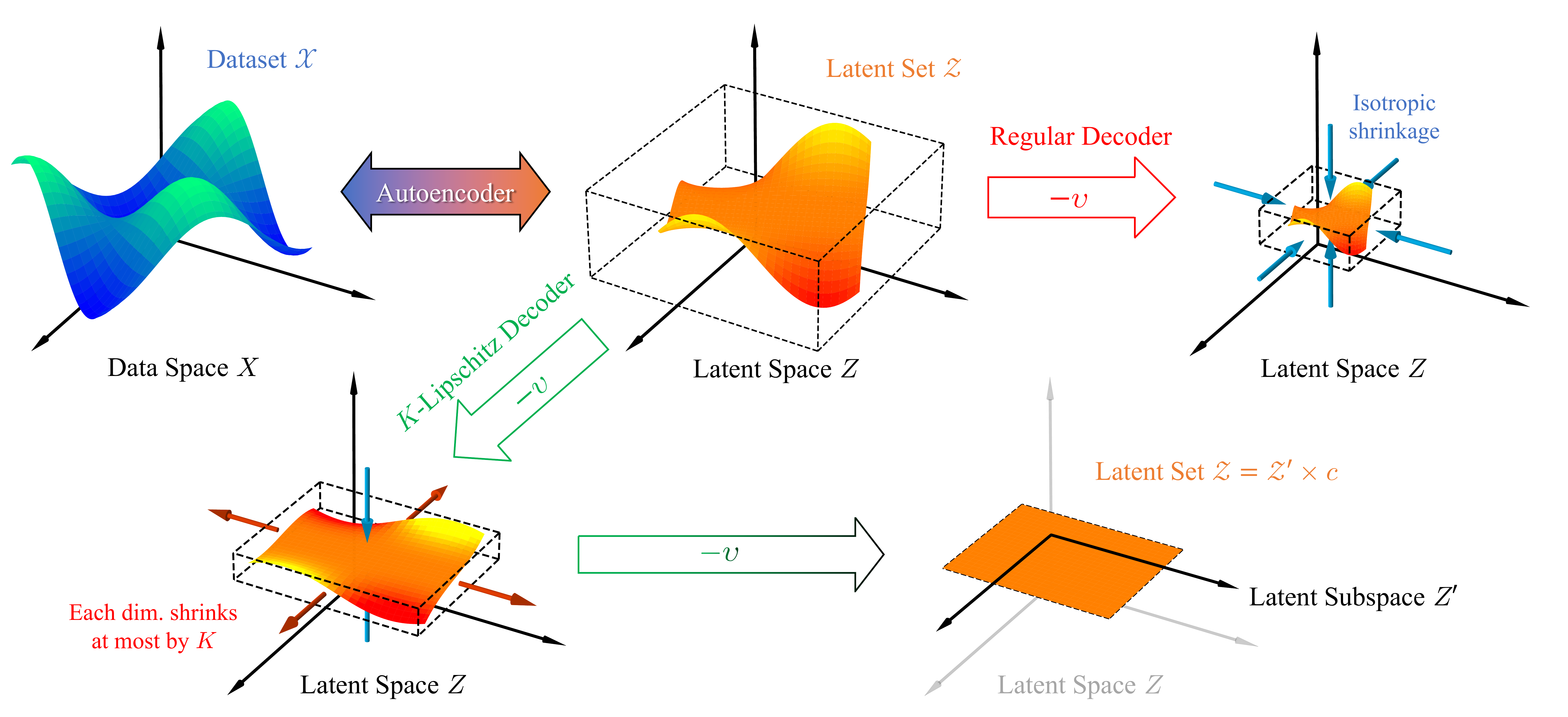}
    \caption{Flattening the latent set via Least Volume (``$-v$''  means reducing the cuboid's volume).}
    \label{fig3:least_volume}
\end{figure}

An alternative regularizer is $\prod_i\sigma_i$\textemdash the \emph{product of all elements of the latent code's STD vector $\pmb{\sigma}$}. We call this quantity the \emph{volume}. It is based on the intuition that a curved surface can only be enclosed by a cuboid of much larger volume than a cuboid that encloses its flattened counterpart. The cuboid has its sides parallel to the latent coordinate axes (as shown in Fig.~\ref{fig3:least_volume}) so that when its volume is minimized, the flattened latent set inside is also aligned with these axes. To evaluate the cuboid's volume, we can regard the STD of each latent dimension as the length of each side of this cuboid. 
The cuboid's volume reaches zero only when one of its sides has zero length (\ie, $\sigma_i=0$), indicating that the latent surface is compressed into a linear subspace. Conceptually, we can then extract this subspace as the new latent space, and continue this compression and extraction \emph{recursively} until the latent set cannot be compressed any more in the final latent space. 
In practice, we can realize this conceptual recursion with an algorithm named \emph{Dynamic Pruning} (see \S\ref{sec:dpa}).

\subsection{Lipschitz Continuity Regularization on Decoder}
\label{sec:lipdec}
Mechanically encouraging such latent sparsity, however, can lead to a trivial solution where we drive all latent STDs close to zero without learning anything useful. This can occur when the elastic latent surface shrinks itself \emph{isotropically} to na\"ively shorten all of the enclosing cuboid's sides, without further flattening the latent surface. To forestall this arbitrary shrinking that causes the trivial solution, we need to properly regularize the autoencoder's latent set's elasticity.
 
The appropriate regularization turns out to be the \emph{Lipschitz continuity of the decoder}. The intuition is as follows.
In the trivial solution, the encoder would na\"ively compresses all latent STDs to almost zero. The decoder would then fight the encoder's shrinking by scaling up its network weights such that the now small perturbations in the latent input still induce large enough output variations in data space to achieve low reconstruction error. To do this, however, the decoder must possess a large Lipschitz constant ($K$) by definition. If instead, the decoder had a small bounded Lipschitz constant, then any latent dimension with STD close to zero must also have close-to-zero variation in the data space due to the small $K$ and thus correspond to a dimension perpendicular to the data surface. Meanwhile, a principal data dimension must retain a large variation in the latent space, as otherwise, the decoder would need to violate its bounded Lipschitz constant to achieve low reconstruction error. This bounded Lipschitz constant on the decoder thereby prevents the encoder from arbitrarily collapsing the latent codes to the trivial solution. 
Figure~\ref{fig3:least_volume} illustrates this intuition. \cite{ghosh2019variational} imposed Lipschitz constraint on the decoder under a similar motivation. 
It can be enforced by \emph{spectral normalization} (see \S\ref{sec:spectral_normalization}). 

\subsection{Least Volume Problems over Euclidean Spaces}
\label{sec:lvf}

For the clarity of the following discussion, we introduce the \emph{volume preorder} over the non-negative orthant $\sR^n_+$ to formalize the concept of volume minimization:
\begin{definition}[Volume Preorder]
\label{def:volumeorder}
    Let the $i$th dimension of a vector $\rva$ be denoted by $a_i$. The \emph{volume preorder} $\lesssim$ is the binary relation on $\sR^n_+$ defined by
    \begin{equation}
        \forall\{\rva, \rvb\} \subset \sR^n_+,\quad
        \rva \lesssim \rvb
        \;\Leftrightarrow\;
        \|\rva\|_0 < \|\rvb\|_0 
        \;\vee\; 
        \left(\|\rva\|_0 = \|\rvb\|_0 
        \;\wedge\; 
        \prod_{a_i > 0} a_i \leq \prod_{b_i > 0} b_i\right).
    \end{equation}
\end{definition}
\noindent It can be verified that this is a \emph{total preorder}~\citep[Definition~1.8]{ehrgott2005multicriteria}. 

\emph{Least Volume} problem aims to train an autoencoder (AE) to achieve a latent set $\gZ$ with a standard deviation (STD) vector $\vsigma$ that is minimized in terms of this volume preorder. Consequently, the optimal $\vsigma$ should not only have the least number of positive components, but also the least amount of \emph{\textbf{volume}} $\prod_{\sigma_i > 0} \sigma_i$ over its positive components. The former leads to dimension reduction and the latter to PCA-like importance ordering (see \S\ref{sec:pca_relation}).
Specifically, the LV problem of a \emph{continuous} autoencoder $(e_\theta, g_\theta)$ with encoder $e_\theta: (X,\normltwo)\to (Z,\normltwo)$ and decoder $g_\theta: (Z,\normltwo)\to (X,\normltwo)$ on a dataset $\gX\subseteq X$ is:
\begin{align}
    \argmin_\gZ \quad& \vsigma(\gZ) \in \left(\sR^n_+,\,\lesssim\right) 
    \label{eq4:lvvol}
    \\
    \text{s.t.} \quad& \gZ \in \mathscr{Z}^\star 
    \coloneqq 
    \left\{\gZ = e_{\theta^\star}(\gX) \mid 
    \theta^\star = \argmin_\theta
    \E_{x\sim\gX}\|g_\theta\circ e_\theta(x)- x\|_2\right\} \label{eq4:lvrec}
    \\
    & \|g_\theta(z_1) - g_\theta(z_2)\|_2 \leq K\|z_1 - z_2\|_2,\quad \forall\{z_1, z_2\} \subseteq Z \label{eq4:lip}
\end{align}
\[
    \begin{tikzcd}
    (X, \normltwo) 
        \arrow[shift left]{r}{e_\theta} 
    & (Z, \normltwo) 
        \arrow[shift left]{l}{g_\theta}
    \end{tikzcd}
\]
where $\vsigma(\gZ)$ is the latent set $\gZ$'s STD to be minimized in terms of the volume preorder. Here the $L_2$ reconstruction loss in (\ref{eq4:lvrec}) can only be minimized to 0 when ${g_\theta\circ e_\theta(x)} = x$ for all $x \in\gX$\textemdash as $\|g_\theta\circ e_\theta(x)- x\|_2$ is continuous \wrt $x\in X$ and $\closure{\gX} = \supp\mdata$\textemdash in which case $\gZ$ is \emph{homeomorphic} to $\gX$. This reconstruction constraint (\ref{eq4:lvrec}), together with the Lipschitz constraint on $g_\theta$ (\ref{eq4:lip}), prevents $\vsigma$ from collapsing to $\vzero$ trivially, as will be discussed in detail in \S\ref{sec:ae_tpem} and \S\ref{sec:safety}. 

Observe that for any \emph{homeomorphism} $h$, the new latent set $h(\gZ) = (h\circ e_\theta)(\gX)$, which is homeomorphic to $\gZ$, is equivalent to $\gZ$ in the sense that $g_\theta\circ e_\theta(x) = (g_\theta\circ h^{-1})\circ(h\circ e_\theta)(x)$. Therefore, as long as $g_\theta$ and $e_\theta$ have enough complexity to respectively represent $g_\theta\circ h^{-1}$ and $h\circ e_\theta$ for a given set $\mathcal{H}$ of $h$, each homeomorphic latent set $h(\gZ)$ with $h\in\mathcal{H}$ must also be an optimal solution to the reconstruction problem, thus residing in $\mathscr{Z}^\star$. Hence, the more complexity $g_\theta$ and $e_\theta$ have, the larger the set $\mathcal{H}$ is, thus a more complicated homeomorphism $h$ we can obtain to flatten $\gZ$ more sufficiently. 
For convenience, we make the following additional assumption:
\begin{assumption}
    We assume that all the neural network encoders $e_\theta$ and decoders $g_\theta$ to be discussed hereafter are \emph{continuous} and \emph{theoretically} have arbitrarily high complexity to serve as universal approximators.
\end{assumption}

\section{Theoretical Analysis of Least Volume}
\label{sec:theory}
Given the least volume formulation, in this section we formalize its surface-flattening intuition with theoretical analysis, inspect the introduced variables' properties and their relationships, and demonstrate its equivalency to PCA in the linear special case. These theoretical results lay the foundation for \S\ref{sec:glv} and \S\ref{sec:implementation}, in which we shall generalize LV for the data spaces equipped with metrics other than the Euclidean distance, and implement the volume minimization process with reliability. 

\subsection{What Exactly Do We Mean by Volume?}
\label{sec:what_volume}
The volume $\prod_i\sigma_i$ of a latent set is the square root of the diagonal product of the covariance matrix $S$ of the latent codes. Apart from this trivial fact, below we show that its square is the tight upper bound of the latent determinant $\det S$, which is referred to as \emph{Generalized Variance} (GV) in some literature. 

\begin{theorem}
The square of volume\textemdash \ie, $\prod_i\sigma_i^2$ \textemdash is a tight upper bound of the latent determinant $\det S$. If $\det S$ is lower bounded by a positive value, then $\prod_i\sigma_i^2$ is minimized down to $\det S$ if and only if $S$ is diagonal.
\label{thm3:bound}
\end{theorem}
\begin{proof}
The diagonal entries of positive semi-definite matrices are always non-negative, and Chol\-esky decomposition states that any real positive definite matrix $S$ can be decomposed into $S = LL^\top$ where $L$ is a real lower triangular matrix with positive diagonal entries. So $\det S = (\det L)^2 = \prod_i [L]^2_{ii} \leq \prod_i \|\mathbf{l}_i\|^2_2 = \prod_i [S]_{ii} = \prod_i \sigma_i^2$, where $\mathbf{l}_i$ is the $i$th row of $L$.

Hence for a positive definite $S$, the inequality can only be reduced to equality when $L$ is diagonal, which in turn makes $S$ diagonal. 
\end{proof}

Theorem~\ref{thm3:bound} can lead to some interesting results that we will see in \S\ref{sec:pca_relation}. Here the \emph{positive lower bound} of $\det S$ is an unknown inherent constant determined by the dataset, and it originates from the intuition that when the latent space reduces into the least dimensional one in which $\gZ$ cannot be flattened anymore, then 
$\det S$ cannot be zero, as otherwise it suggests $\gZ$ resides in a linear latent subspace and creates contradiction. Nor should $\det S$ 
 be arbitrarily close to 0, as the $K$-Lipschitz decoder prevents degenerate shrinkage.

\subsubsection{Volume is Better Than \texorpdfstring{$L_1$}{L1} for Flattening}
\label{sec:pedago}

\begin{figure}[hbt!]
    \centering
    \includegraphics[width=0.5\textwidth]{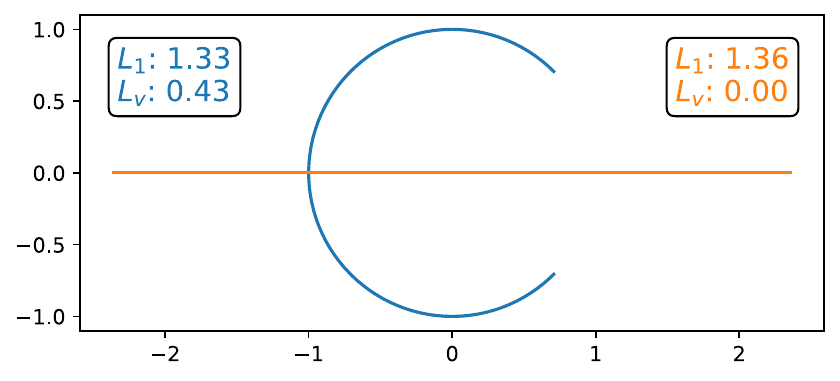}
    \caption{A pedagogical example where minimizing the $L_1$ regularizer produces less flattened representations than minimizing the volume. Here ``$L_1$'' refers to $\|\pmb{\sigma}\|_1$ and ``$L_v$'' refers to $\prod_i\sigma_i$.
    }
    \label{fig3:exm}
\end{figure}
We can use a simple example to show that the volume is better than $L_1$-based regularizers in terms of determining the sparsity of $\pmb{\sigma}$, \ie, how flat the latent set is.
Figure~\ref{fig3:exm} shows a straight orange line $\gZ_l$ of length $\frac{3}{2}\pi$ and a blue arc $\gZ_a$ of the same length. Clearly there exist an isometry $h: \gZ_l\to\gZ_a$. Suppose the arc is a latent set created through $\gZ_a = e(\gX)$, then $\gZ_l = h^{-1}\circ e(\gX)$ is an equivalently good one, provided that the encoder $e$ has enough complexity to learn $h^{-1}\circ e$. Moreover, because the isometry $h$ is 1-Lipschitz, then a $K$-Lipschitz decoder $g$ with enough complexity can also learn $g\circ h$ since this is also $K$-Lipschitz, so both $\gZ_l$ and $\gZ_a$ are latent sets that can be produced by $e$ and $g$ with enough complexity. 
Now we introduce uniform distribution over these two sets and evaluate $\|\pmb{\sigma}\|_1$ and $\prod_i\sigma_i$ on them. Figure~\ref{fig3:exm} shows that only the volume $\prod_i\sigma_i$ correctly tells the line $\gZ_l$ is flatter.

\subsection{An Errorless Continuous Autoencoder Learns a Topological Embedding}
\label{sec:ae_tpem}
It is pointless to only minimize the volume, given that we can always reduce it to 0 by trivially encoding every data sample to the same latent point. To make the ``flattening'' process in \S\ref{sec:vol_pen} meaningful, we need to ensure the autoencoder has good reconstruction performance over the dataset, such that the latent set is a low-dimensional replica of the dataset that preserves useful topological information. This is justified by the following propositions.

\begin{lemma}
If $\forall x\in \gX,\;g\circ e(x) = x$, then both $e$ and $g$ are bijective between $\gX$ and $\gZ = e(\gX)$.
\label{lemma3:bi}
\end{lemma}
\begin{proof}
$\restr{e}{\gX}$ is injective because for any pair $\{x,\,x'\}\subseteq\gX$ that satisfies $e(x) = e(x')$, we have $x = g\circ e(x) = g\circ e(x') = x'$. It is then bijective because $\restr{e}{\gX}$ is surjective onto $\gZ$ by definition. 
$\restr{g}{\gZ}$ is surjective onto $\gX$ because $g(\gZ) = g\circ e(\gX) = \gX$. It is then injective thus bijective because $\forall z = e(x)\in\gZ$, $e\circ g(z) = e\circ g\circ e(x) = e(x) = z$.
\end{proof}

The \emph{continuity} of the autoencoder then plays a crucial role in preserving the dataset's structure:

\begin{theorem}
A continuous autoencoder $(e, g)$ between the data space $X=\sR^n$ and the latent space $Z=\sR^m$ with the norm-based reconstruction error $\E_{x\sim \gX}\|g\circ e(x) - x\| = 0$ learns a topological embedding of the dataset $\gX$. In other words, the latent set $\gZ = e(\gX)\subseteq Z$ is a homeomorphic copy of the dataset.
\label{thm3:ae}
\end{theorem}
\begin{proof}
Due to the continuity of $\|g\circ e(x) - x\|$ \wrt $x$, the positive definiteness of norm and Lemma~\ref{lemma3:bi}, both the encoder and decoder restrictions $\restr{e}{\gX}$ and $\restr{g}{\gZ}$ are bijective functions between $\gX$ and $\gZ$. Because both the encoder $e:X\to Z$ and the decoder $g: Z\to X$ are continuous, their restrictions are also continuous (in terms of the subspace topology). Since $\restr{g}{\gZ}$ is then the continuous inverse function of $\restr{e}{\gX}$, by definition $\restr{e}{\gX}$ is a topological embedding (see Def.~\ref{def:topo_embedding}). 
\end{proof}

Because $\gZ$ is a homeomorphic copy of $\gX$, $\gX$'s important topological properties like connectedness and compactness\textemdash which are invariant under homeomorphism\textemdash are preserved on $\gZ$. This means if some of these invariant properties are the only information we rely on when analyzing $\gX$, then analyzing $\gZ$ is equivalent to directly analyzing $\gX$. For instance, if we want to know how many disconnected components $\gX$ has, or evaluate its local dimensionality, then theoretically we can derive the same results from $\gZ$. Of course, $\gZ$ becomes more efficient to analyze if it resides in a low-dimensional linear subspace that is easy to extract. 

\begin{corollary}
If the above errorless autoencoder's latent set has the form $\gZ = \gZ'\times c$ where $c\in\sR^p$ is constant and $\gZ'\subseteq Z' = \sR^{m-p}$, then $\pi \circ \restr{e}{\gX}$ is also a topological embedding of $\gX$, where $\pi: Z'\times c\ni z'\times c\mapsto z' \in Z'$ is the projection map. 
\label{corollary:prune}
\end{corollary}
\begin{proof}
$\pi$ is homeomorphic because it is bijective, while continuous and open in terms of the product topology. Thus $\pi \circ \restr{e}{\gX}$ is also homeomorphic onto its image $\gZ'$.
\end{proof}
\begin{remark}
Because $\gZ'$ is homeomorphic to $\gX$, the subspace dimension $\dim Z' = m-p$ cannot be lower than $\dim \gX$, as otherwise it violates the topological invariance of dimension~\citep[Theorem~13.22]{lee2010introduction}. So we need not to worry about extravagant scenarios like ``$\gZ'$ collapsing into a line while $\gX$ is a surface''. This is not guaranteed for a discontinuous autoencoder since it does not learn a topological embedding.
\end{remark}

Corollary~\ref{corollary:prune} suggests that if we can force the errorless autoencoder's latent set to assume such a form while increasing the vector $c$'s dimension $p$ as much as possible, we can obtain a latent space that is the lowest dimensional Euclidean space that can still correctly embed the dataset. If achieved, the resulting low-dimensional latent space is not only more efficient to analyze, but also provides useful information about the topology of $\gX$. For instance, not every smooth $m$ dimensional manifold can be embedded in $\sR^m$ (\eg, sphere and Klein bottle), but the \emph{Whitney Embedding Theorem}~\cite[Theorem~6.19]{lee2012smooth} states that it can be embedded in $\sR^{2m}$. Thus if $\gX$ is a smooth manifold, then obtaining the least dimensional $Z$ through a \emph{smooth} autoencoder can provide a lower bound of $\gX$'s intrinsic dimensionality. 

In practice, the autoencoder's ideal everywhere-errorless-reconstruction is enforced by minimizing the reconstruction error $\epsilon = \E\|g\circ e(x) -x\|$, yet due to inevitable numerical errors and data noise that should be ignored by the autoencoder reconstruction, $\epsilon$ cannot strictly reach zero after optimization. Likewise, during volume reduction, numerically the latent set cannot strictly take the form in Corollary~\ref{corollary:prune}, but rather only at best has marginal STDs in several latent dimensions. Intuitively, we may consider the latent set as flattened into a thin plane and discard these nearly-constant latent dimensions to obtain a latent set of lower dimensionality with $\pi$, in order to continue the recursive volume minimization as devised in \S\ref{sec:vol_pen}. But how do we properly trim off these dimensions? And more importantly, is it safe?

\subsection{Is it Safe to Prune Trivial Latent Dimensions?}
\label{sec:safety}
Suppose a latent set $\gZ$ has several STDs close to zero after numerical optimization. Although such $\gZ$ is not exactly in the form of $\gZ'\times c$ in Corollary~\ref{corollary:prune} for us to apply $\pi$, we can still regarded its latent dimensions of marginal STDs as trivial, in the sense that \emph{pruning} them\textemdash \ie, fixing their value at their mean\textemdash will not induce much difference to the decoder $g$'s reconstruction, provided that $g$'s Lipschitz constant is not large. This is justified by the following theorem.
\begin{theorem}
Suppose the STD of latent dimension $i$ is $\sigma_i$. If we fix $z_i$ at its mean $\bar{z}_i$ for each $i\in P$ where $P$ is the set of indices of the dimensions we decide to prune, then the $L_2$ reconstruction error of the autoencoder with a $K$-Lipschitz decoder $g$ increases by at most $K \sqrt{\sum_{i\in P} \sigma_i^2}$.
\label{thm3:safety}
\end{theorem}
\begin{proof}
Suppose the reconstruction error $\epsilon$ is measured by $L_2$ distance in the data space as $\epsilon = \E\|x - \hat{x}\| = \E\|x - g\circ e(x)\| = \E\|x - g(z)\|$. Let the new reconstruction error after latent pruning be $\tilde{\epsilon} = \E\|x - \tilde{x}_P\| = \E\|x - g\circ p_P \circ e(x)\| = \E\|x - g(\tilde{z}_P)\|$, where $p_P$ is the pruning operator defined by $[\tilde{z}_P]_i = [p_P(z)]_i = \begin{cases}z_i & i\not\in P \\ \bar{z}_i & i\in P\end{cases}$. Then due to the subadditivity of norm and the $K$-Lipschitz continuity of $g$ we have
\begin{align}
    \tilde{\epsilon} - \epsilon 
    &\leq \E\|\hat{x} - \tilde{x}_P\| 
    = \E\|g(\tilde{z}_P) - g(z)\| \nonumber\\
    &\leq  K\cdot\E\|\tilde{z}_P - z\|
    = K \sqrt{\E\left[\|\tilde{z}_P - z\|^2\right] - \var\left[\|\tilde{z}_P - z\|\right]} \label{eqn:ti_b} \\
    &\leq K \sqrt{\E\left[\|\tilde{z}_P - z\|^2\right]}
    = K \sqrt{\sum_{i\in P}\E\left[(z_i - \bar{z}_i)^2\right]}
    = K \sqrt{\sum_{i\in P} \sigma_i^2} \label{eqn:lo_b}
\end{align}
Of course, (\ref{eqn:ti_b}) is an upper bound tighter than (\ref{eqn:lo_b}), but the minimalistic (\ref{eqn:lo_b}) also conveys our idea clearly.
\end{proof}

The pruned $\tilde{\gZ}$ is then of the very form $\tilde{\gZ} = \gZ'\times c$ and can be fed to $\pi$ to extract $\gZ'$. The inverse $\pi^{-1}$ helps map $\gZ'$ back into the data space through $g\circ \pi^{-1}$ without inducing large reconstruction error.
This theorem also supports our intuition in \S\ref{sec:lipdec} about why having a Lipschitz continuous decoder is necessary for learning a compressed latent subspace\textemdash for small $K$, a latent dimension of near-zero STD cannot correspond to a principal dimension of the data manifold. In contrast, in the absence of this constraint, the decoder may learn to scale up $K$ to align near-zero variance latent dimensions with some of the principal dimensions.

\subsection{Equivalence to PCA and the Importance Ordering Effect}
\label{sec:pca_relation}
Surprisingly, one can prove in Proposition~\ref{prop3:pca} that PCA is a linear special case of least volume. This implies there could be more to volume minimization than just flattening the surface. 

\subsubsection{Equivalence to PCA}
\begin{lemma}
Suppose for a symmetric matrix $S$ we have $A^\top SA = \Sigma$ and $A^\top A = I$, where $\Sigma$ is a diagonal matrix whose diagonal elements are the largest eigenvalues of $S$. Then $A$'s columns consist of the orthonormal eigenvectors corresponding to the largest eigenvalues of $S$.
\label{lemma3:sim_diag}
\end{lemma}
\begin{proof}
Denote the eigenvalues of $S$ by $\sigma_1 \geq \sigma_2 \geq \cdots$, the corresponding orthonormal eigenvectors by $u_1, u_2, \cdots$, and the $i$th column vector of $A$ by $a_i$. Without any loss of generality, we also assume that the diagonal elements of $\Sigma$ are sorted in descending order so that $\Sigma_{ii} = \sigma_i$, as the following proof is essentially independent of the specific order.

We can prove this lemma by induction. To begin with, suppose for $a_1$ we have $a_1^\top S a_1 = \sigma_1$, then $a_1$ must fall inside the eigenspace $E_{\sigma_1}$ spanned by the eigenvector(s) of $\sigma_1$. Assume this is not true, then $a_1 = v + v_\perp$ where $v\in E_{\sigma_1}$, $v_\perp \perp E_{\sigma_1}$, $\|v_\perp\| \neq 0$ and $\|a_1\|^2 = \|v\|^2 + \|v_\perp\|^2 = 1$. It follows that $a_1^\top S a_1 = v^\top S v + v^\top_\perp S v_\perp < \sigma_1\|v\|^2 + \sigma_1\|v_\perp\|^2 = \sigma_1$, which violates the assumption. The inequality holds because we have the decomposition $v_\perp = \sum_{i\in\mathcal{I}} \alpha_i u_i$, $\mathcal{I} = {\{i\mid u_i\perp E_{\sigma_1}\}}$ where $u_i$'s corresponding $\sigma_i$ is smaller than $\sigma_1$, so $v^\top_\perp S v_\perp = \sum_{i\in\mathcal{I}} \sigma_i \alpha_i^2 < \sigma_1\sum_{i\in\mathcal{I}}  \alpha_i^2 = \sigma_1\|v_\perp\|^2$.

Now assume that for $a_k$, its predecessors $a_1\dots a_{k-1}$ respectively all fall inside the corresponding eigenspaces $E_{\sigma_1}\dots E_{\sigma_{k-1}}$. Since $a_k$ is orthogonal to $a_1\dots a_{k-1}$, we can only find $a_k$ in $\sum_{i\geq k} E_{\sigma_i}$. Then following the same rationale as in the $a_1$ case, $a_k$ must be inside $E_{\sigma_k}$.
\end{proof}

\begin{proposition}
An autoencoder recovers the principal components of a dataset $X$\footnote{For convenience, in this very proposition  we abuse the notation a bit by using $X$ to refer to the \emph{data matrix} in $\sR^{d\times n}$ rather than the data space where $\gX$ lives. In other words, the matrix $X$ consists of $n$ samples from $\gX$.} after minimizing the volume, if we: 
\begin{enumerate}
    \item Make the encoder $e(x) = Ax + a$ and the decoder $g(z) = Bz + b$ linear maps, 
    \label{cond:1}
    \item Force the reconstruction error $\|g\circ e(X) - X\|_F$ to be strictly minimized,
    \label{cond:2}
    \item Set the Lipschitz constant to 1 for the decoder's regularization,
    \label{cond:3}
    \item Assume $\rank(X)$ is not less than the latent space dimension $m$.
    \label{cond:4}
\end{enumerate}
Specifically, $B$'s columns consist of the principal components of $X$, while $A$ acts exactly the same as $B^\top$ on $X$. When $X$ has full row rank, $A = B^\top$.
\label{prop3:pca}
\end{proposition}
\begin{proof}
According to~\citep{bourlard1988auto}, minimizing the reconstruction error cancels out the biases $a$ and $b$ together, and simplifies the reconstruction loss into $\|BAX' - X'\|_F$ where $X' = X - \bar{X}$. Therefore, for simplicity and without any loss of generality, hereafter we assume $\bar X = 0$, $e(x)=Ax$ and $g(z)=Bz$. Given a dataset $X\in\sR^{d\times n}$ of $n$ samples, Condition~\ref{cond:1} means the corresponding latent codes are $Z = AX$, whose covariance matrix is $S = \frac{1}{n-1} Z Z^\top = \frac{1}{n-1} AX X^\top A^\top = A S_X A^\top$ where $S_X \coloneqq \frac{1}{n-1} X X^\top$ is the data covariance.

Condition~\ref{cond:2} necessitates that both the encoder $A$ and the decoder $B$ possess full-rank. This comes from the inequality $\rank(AB)\leq \min (\rank(A), \rank(B))$ and the \emph{Eckart–Young–Mirsky theorem}~\citep{eckart1936approximation}. Moreover, for a full-rank decoder $B=V\Sigma^{-1} U^\top$ (here for simplicity we use the SVD formulation where $\Sigma$ is diagonal), the encoder $A$ must satisfy $AX = B^\dagger X = U\Sigma V^\top X$ when we minimize the reconstruction loss $\|BAX-X\|_F$, so the reconstruction loss reduces into  $\|VV^\top X-X\|_F$. From the minimum-error formulation of PCA~\citep{Bishop07}, we know $V$ should be of the form $V = PQ$, where $Q^\top Q = QQ^\top = I$ and $P$’s column vectors are the orthonormal eigenvectors corresponding to the $m$ largest eigenvalues of the data covariance $S_X$, such that $P^\top S_X P = \Lambda$ where $\Lambda$ is a diagonal matrix of the $m$ largest eigenvalues. Therefore we can only minimize the volume by changing $U$, $\Sigma$ and $Q$. 

It follows that $S = A S_X A^\top = U\Sigma V^{\top} S_X V\Sigma^\top U^\top = U\Sigma Q^\top\Lambda Q\Sigma^\top U^\top$. Because according to Theorem~\ref{thm3:bound}, minimizing the volume $\prod_i^m \sqrt{[S]_{ii}}$ minimizes the determinant $\det S = (\det \Sigma)^2 \det \Lambda$, then under Condition~\ref{cond:3} and~\ref{cond:4}, this is minimized only when $[\Sigma]_{ii} = 1$ for all $i$, given that the decoder is required to be 1-Lipschitz so $[\Sigma]_{ii}\geq 1$ for all $i$, and $\det\Lambda>0$. Hence $\Sigma = I$, $B^\dagger = B^\top$ and $B$ has orthonormal columns because $B^\top B = U\Sigma^{-2} U^\top = I$.

Therefore $S^\star$\textemdash the volume-minimized $S$\textemdash satisfies $S^\star =UQ^\top\Lambda Q U^\top$ and thus is orthogonally similar to $\Lambda$. Since $S^\star$ is diagonal (Theorem~\ref{thm3:bound}), it must have the same set of diagonal entries as those of $\Lambda$, though not necessarily in the same order. Thus when the volume is minimized, we have $ AS_XA^\top = B^\dagger S_X (B^\dagger)^\top = B^\top S_X B = S^\star$. Because $B$'s columns are orthonormal, according to Lemma~\ref{lemma3:sim_diag}, this identity only holds when $B$'s column vectors are the eigenvectors corresponding to the $m$ largest eigenvalues of $S_X$, so the linear decoder $B$ consists of the principal components of $X$. 
\end{proof}
\begin{remark}
Although $A$ is not necessarily equal to $B^\top$, it projects $X$ into the latent space the same way as $B^\top$. It can also be verified that it must have the form $A = B^\top + W$ where each row of $W$ lies in $\ker X^\top$. This means the identity $A = B^\top$ holds when $X$ has full row rank, in which case $A$ also recovers the principal components of $X$.
It is also easy to check that setting the decoder’s Lipschitz constant to any value $K$ will just scale down all the eigenvalues by $K^2$, without hurting any of the principal dimension aligning of PCA or distorting the ratio between dimensions.
\end{remark}

This proof indicates that minimizing the volume is disentangled from reducing the reconstruction loss, at least for the linear case. Specifically, the reconstruction loss controls only $P$, while the least volume regularization controls the rotation and scaling in the latent space respectively through $U$, $Q$ and $\Sigma$. The product $U\Sigma Q^\top$ then models the family $\mathcal{H}$ of linear homeomorphisms that the linear autoencoder (without bias vectors) can additionally learn beyond preserving information through $P^\top$, as discussed in \S\ref{sec:lvf}. Indeed, the linear encoder is of the form $A = (U\Sigma Q^\top)\circ P^\top$ while the linear decoder is of the form $B = P \circ Q \Sigma^{-1} U^\top = P \circ (U\Sigma Q^\top)^{-1}$. 

Due to the minimization of volume, $\Sigma$ ends up being an identity matrix, making both $g$ and $e$ isometric over the dataset $X$. This means $e$ does not scale up or down any data dimension in the latent space. We argue that this is the primary reason why each PCA latent dimension's degree of importance scales with its variance, because one can easily check that as long as $\Sigma = I$, Proposition~\ref{prop3:exp_rec} below still holds for an ``imperfect'' PCA where $A = B^\top = UQ^\top P^\top$. We may expect a similar ordering effect for nonlinear autoencoders.

The rotations $U$ and $Q$, on the other hand, help cram as much data information into the most primary latent dimensions as possible, such that when we remove any number of the least informant latent dimensions, the remaining dimensions still store the most amount of information they can extract from $X$ in terms of minimizing the reconstruction loss. For nonlinear autoencoders, we may expect a cramming effect that is similar yet not identical, because after stripping off some least informant latent dimensions, the nonlinear autoencoder might further reduce the reconstruction loss by curling up the decoder's image\textemdash now  of lower dimension\textemdash in the dataset to traverse more data region.

\subsubsection{Importance Ordering Effect}
The proof of Proposition~\ref{prop3:pca} suggests that the ordering effect of PCA\textemdash \ie, the magnitude of each latent dimension's variance indicates each latent dimension's degree of importance\textemdash is at least partially a result of reducing the latent determinant until the singular values of $B$ reach their upper bound 1, such that $g$ becomes isometric and \emph{preserves distance}. In other words, this means $e$ is not allowed to scale up the dataset $X$ along any direction in the latent space if it is unnecessary, as it will increase the latent determinant, and thus the volume. Therefore, likewise, although Theorem~\ref{thm3:safety} does not prevent a latent dimension of large STD from being aligned to a trivial data dimension, we may still expect minimizing the volume or any other sparsity penalties to hinder this unnecessary scaling by the encoder $e$, given that it increases the penalty's value. This suggests that the latent dimension's importance in the data space should in general scale with its latent STD.

To investigate if least volume indeed induces a similar importance ordering effect for nonlinear autoencoders, we need to first quantify the importance of each latent dimension. This can be done by generalizing the \emph{Explained Variance} used for PCA, after noticing that it is essentially \emph{Explained Reconstruction}:

\begin{proposition}
Let $\lambda_i$ be the eigenvalues of the data covariance matrix. The explained variance $\mathcal{R}(\{i\}) = \frac{\lambda_i}{\sum_j^n \lambda_j}$ of a given latent dimension $i$ of a linear PCA model is the ratio between $\mathcal{E}_{\|\cdot\|_2^2}(\{i\})$\textemdash the MSE reconstruction error induced by pruning this dimension (\ie, setting its value to its mean $0$) and $\mathcal{E}_{\|\cdot\|_2^2}(\Omega)$\textemdash the one induced by pruning all latent dimensions, where $\Omega$ denotes the set of all latent dimension indices.
\label{prop3:exp_rec}
\end{proposition}
\begin{proof}
First we show that the $i$th eigenvalue $\lambda_i$ of the data covariance matrix equals the MSE introduced by pruning principal dimension $i$. Let $v_i$ denote the eigenvector corresponding to the $i$th principal dimension. Then pruning this dimension leads to an incomplete reconstruction $\tilde{x}_{\{i\}}$ that satisfies:
\begin{align}
    \mathcal{E}_{\|\cdot\|_2^2}(\{i\}) := \E[\|\tilde{x}_{\{i\}} - x\|^2_2] 
    = \E[\|\langle x, v_i\rangle \cdot v_i\|^2_2] 
    = \E[\langle x, v_i\rangle^2]
    = \lambda_i
    \label{eqn:lamb}
\end{align}
Next we show that the $\mathcal{E}$ of linear PCA has additivity, \ie, the joint MSE induced by pruning several dimensions in $P$ altogether equals the sum of their individual induced MSE:
\begin{equation}
\begin{aligned}
    \mathcal{E}_{\|\cdot\|_2^2}(P) :=\E[\|\tilde{x}_P - x\|^2_2] 
    &= \E[\|\textstyle\sum_{i\in P}\langle x, v_i\rangle \cdot v_i\|^2_2] \\
    &= \E[\langle \textstyle\sum_{i\in P}\langle x, v_i\rangle \cdot v_i,\;\textstyle\sum_{i\in P}\langle x, v_i\rangle \cdot v_i\rangle] \\
    &= \E[\textstyle\sum_{i\in P}\langle x, v_i\rangle^2]
    = \textstyle\sum_{i\in P}\E[\langle x, v_i\rangle^2] \\
    &= \textstyle\sum_{i\in P}\lambda_i
    =\textstyle\sum_{i\in P}\mathcal{E}_{\|\cdot\|_2^2}(\{i\})
\end{aligned}
\end{equation}
Hence the claim is proved.
\end{proof}

So for PCA, the explained variance actually measures the contribution of each latent dimension in minimizing the MSE reconstruction error $\E\|x-\hat x\|_2^2$ as a percentage. Since for a nonlinear model the identity $\mathcal{R}(\{i\}) + \mathcal{R}(\{j\}) = \mathcal{R}(\{i,j\})$ generally does not hold, there is no good reason to stick with the MSE. It is then natural to extrapolate $\mathcal{R}$ and $\mathcal{E}$ to our nonlinear case by generalizing $\mathcal{E}_{\|\cdot\|_2^2}(P)$ to $\mathcal{E}_D(P)$, \ie, the \emph{induced reconstruction error \wrt metric $D$ after pruning dimensions in $P$}:
\begin{equation}
    \mathcal{E}_D(P)=\E[D(\tilde{x}_P, x)] - \epsilon = \E[D(\tilde{x}_P, x)] - \E[D(\hat{x}, x)]
    \label{eq:ind_rec}
\end{equation} 
Then $\mathcal{R}_D(P)$\textemdash the \emph{explained reconstruction} of latent dimensions in $P$ \wrt metric $D$\textemdash can be defined as $\mathcal{R}_D(P) = \mathcal{E}_D(P) / \mathcal{E}_D(\Omega)$. 
In later experiments we shall use $L_2$ error as $D$ to verify the  importance ordering effect (see \S\ref{sec:img_exp}).
\begin{remark}
\label{rm:gpca}
The subtraction in Eqn.~\ref{eq:ind_rec} is a generalization of the case where the PCA model’s number of components is set smaller than the data covariance’s rank $r$. In this case, the PCA model—as a linear AE—will inevitably have a reconstruction error $\epsilon$ even if no latent dimension is pruned.
Yet we may regard this inherent error $\epsilon = \E[\|\tilde{x}_I - x\|_2^2]$ as the result of implicitly pruning a fixed collection $I$ of additional $r-n$ latent dimensions required for perfect reconstruction. Then $\lambda_i$ in (\ref{eqn:lamb}) can instead be explicitly expressed as $\lambda_i = \E[\|\tilde{x}_{\{i\}\cup I} - x\|_2^2] - \epsilon$ to accommodate this implicit pruning. Therefore the induced reconstruction error $\mathcal{E}_{\|\cdot\|_2^2}(\{i\})$ is actually the change in reconstruction error before and after pruning $i$ (with the inherent reconstruction error $\epsilon$ taken into account). The one in Eqn.~\ref{eq:ind_rec} is thus a generalization of this, using the change in reconstruction error to indicate each dimension's importance.
\end{remark}

\section{Generalization of Least Volume Problems}
\label{sec:glv}
The LV problem marks a positive start for retrieving a least dimensional representation $\gZ$ of a given dataset $\gX$. 
However, this formulation is derived based on the strong assumption that both the latent space $Z$ and the data space $X$ are equipped with \emph{Euclidean distances}. As reasoned in \S\ref{sec:pca_relation}, after solving such a problem, if a given latent dimension has a large $\sigma_i$, it only means the data variation it controls is important \wrt the Euclidean distance, but often we might be interested in a $X$ equipped with some more meaningful metrics (\eg, perceptual metric on images, performance-aware metric on a set of mechanical designs, such as the airfoil example in \S\ref{sec:introduction}). In addition, the current LV problem is designed only for an unlabeled dataset. For a labeled one, incorporating the label information in training may help derive a more useful representation for different downstream tasks (\eg, like what \emph{transfer learning}~\citep{weiss2016survey} is doing). Can we fulfill these two demands, and how might we do so? Can LV produce meaningful results in these cases?

In this section, we show that these two demands can actually be both solved with an elegant generalization of LV\textemdash dubbed \emph{Generalized Least Volume} (GLV). 
The general idea is that for applications in which the metrics $d_Z$ and $d_X$ of the latent space $Z$ and the data space $X$ are not $L_2$ distances, we can first transform them into $L_2$ metric spaces via \emph{isometries}\textemdash the distance-preserving maps\textemdash then apply LV equivalently between the $L_2$ spaces, and eventually transform the result back to the original metric spaces via isometries again. 
Moreover, this transformation naturally allows us to deal with labeled datasets. It will be shown that the labels can be readily incorporated into the GLV formulation if we make the data metric $d_X$ informed by the labels.

\paragraph{Nomenclature} For simplicity, hereafter we use $\cong$ and $\simeq$ between two given sets to respectively denote them being \emph{isometric} (Def.~\ref{def:isometry_metric}) and \emph{homeomorphic} (Def.~\ref{def:homeomorphism}) to each other. 
Also, we denote a metric space $X$ equipped with a metric $d_X$ by $(X, d_X)$, and unless there are multiple metrics constructed on the same set $X$, we sometimes omit $d_X$ and just denote the metric space by $X$ if it is already claimed or implied that $X$ has no metric other than $d_X$. 
Additionally, for a function $f:X\to Y$, we use $\Bar{f}$ to denote its \emph{corestriction} $\corestr{f}{f(X)}$.
Furthermore, we have the two equivalence relations $\rva\sim\rvb\Leftrightarrow \rva\lesssim\rvb\wedge\rvb\lesssim\rva$, and $\gZ_1\sim\gZ_2 \Leftrightarrow \vsigma(\gZ_1)\sim \vsigma(\gZ_2)$, which basically means $\gZ_1$ and $\gZ_2$ have the same dimensionality (\ie, $\|\vsigma(\cdot)\|_0$) and volume. We use $[\gZ]$ to denote the equivalence class of $\gZ$. When we minimize $\vsigma$, it is always done in $\left(\sR^n_+,\,\lesssim\right)$. 
At last, we use $(e, g)_{A\to B\to C}$ to denote an autoencoder with encoder $e:A\to B$ and decoder $g:B\to C$
\footnote{Although here $C$ is not necessarily equal to $A$, such that there may be nothing ``auto'' in $(e,g)_{A\to B\to C}$, we still refer to it as (generalized) autoencoder for convenience, as Lemma~\ref{lemma4:gae} shows that even if $C\neq A$, $(e, g)_{A\to B\to C}$ can still operate like a regular autoencoder $(e, g)_{A\leftrightarrow B}$ if we have the correct $C$.}. 
If $A=C$, then we use the notation $(e, g)_{A\leftrightarrow B}$ instead.

\subsection{Definitions and Lemmas}
We begin with a few important concepts and lemmas that will be used frequently later.

\begin{definition}[Isometry]
\label{def:isometry_metric}
    A map $\phi: (X, d_X)\to (Y, d_Y)$ between two metric spaces is called an \emph{isometry} if $d_X(x_1, x_2) = d_Y(\phi(x_1), \phi(x_2))$. If a \emph{bijective} isometry exists between $X$ and $Y$, we say $X$ and $Y$ are isometric to each other, denoted by $X\cong Y$.
\end{definition}

\begin{lemma}[Bijective Isometries are Homeomorphisms]
    $X\cong Y \Rightarrow X\simeq Y$.
\end{lemma}
\begin{proof}
    Let $V\subseteq Y$ be an open set and $\phi:X\to Y$ be a bijective isometry. For every point $x\in U\coloneqq \phi^{-1}(V)\subseteq X$, there must exists an open ball ${\{\phi^{-1}(y)\mid y\in V, d_Y(\phi(x), y)<\epsilon\}}\subseteq U$ centered at $x$, so $U$ is open, thus $\phi$ is continuous. Likewise, $\phi^{-1}$ is continuous. 
\end{proof}

\begin{lemma}[Equivalence of Lipschitz Continuity]
\label{lemma4:lip}
    $f:(Z, d_Z)\to (X, d_X)$ is $K$-Lipschitz \emph{iff} $\Tilde{f}: (\tdZ, d_{\tdZ})\to (\tdX, d_{\tdX}) =\phi \circ f \circ \psi^{-1}$ is $K$-Lipschitz, where $\psi: Z \to \tdZ$ and $\phi: X \to \tdX$ are isometries and $\psi$ is \emph{bijective}. 
\end{lemma}
\begin{proof}
$d_{\tdX}(\tdf(\tdz_1), \tdf(\tdz_2))
=d_{\tdX}(\phi\circ f(z_1), \phi\circ f(z_2))
=d_X(f(z_1), f(z_2))
\leq K\cdot d_Z(z_1, z_2)
= K\cdot d_{\tdZ}(\psi(z_1), \psi(z_2))
=K\cdot d_{\tdZ}(\tdz_1, \tdz_2)$.
\end{proof}

\begin{definition}[Pullback Metric]
\label{def:pullback}
    If $\phi:X\to (W, d_W)$ is injective, then $(X, \phi^*d_W) \cong (\phi(X), d_W)$, where the \emph{pullback metric} $\phi^*d_W(x_1, x_2) \coloneqq d_W(\phi(x_1), \phi(x_2))$ is the pullback of $d_W$ by $\phi$. 
\end{definition} 
\begin{remark}
    It is easy to verify that $\phi^*d_W$ is a metric. This provides us a metric on $X$ given another metric space $W$ if an injection $\phi: X\to W$ exists.
    This by definition makes $\phi$ isometric and thus homeomorphic onto its image $\phi(X)$ (in terms of the topologies induced by $\phi^*d_W$ and $d_W$). 
\end{remark}

\begin{lemma}[Pullback Metric from Embedding]
\label{lemma4:pulltopo}
    If $\phi:(X, d_X)\to (W, d_W)$ is a topological embedding (see Def.~\ref{def:topo_embedding}), then $(X, d_X)$ and $(X, \phi^*d_W)$ have the same topology\textemdash which means $\id_X: (X, d_X)\to (X, \phi^*d_W)$ is a homeomorphism.
\end{lemma}
\begin{proof}
    Denote the topologies of $(X, d_X)$ and $(X, \phi^*d_W)$ by $\gT_X$ and $\gT_{X^*}$. 
    By construction $(X, d_X)\simeq (\phi(X), d_W) \cong (X, \phi^*d_W)$, so $\gT_X \subseteq \gT_{X^*}$ and $\gT_{X^*} \subseteq \gT_X$. Hence $\gT_X = \gT_{X^*}$.
\end{proof}

\begin{lemma}[Generalized Autoencoders]
\label{lemma4:gae}
    Suppose there exists a homeomorphism $\phi:\gX\to\Tilde{\gX}\subseteq\tdX$. If ${g\circ e(x)} = \phi(x)$ for any $x\in\gX$ while the encoder $e:X\to Z$ and the decoder $g:Z \to \Tilde{X}$ are continuous, then $\gZ\coloneqq e(\gX)\simeq \gX$.  
\end{lemma}
\begin{proof}
    $g\circ e(x) = \phi(x)\Rightarrow (\phi^{-1}\circ g) \circ e(x)=x$. Then it follows from Theorem~\ref{thm3:ae}.
\end{proof}

\subsection{Generalized Least Volume}

Isometries allow us to generalize Least Volume problems to data spaces and latent spaces of metrics other than $L_2$ distance. We refer to any such LV problem as \emph{Generalized Least Volume} (GLV) problem when we want to emphasize its distinction from the regular LV problems between $L_2$ spaces.
The following theorem shows we can transform a GLV problem between two arbitrary metric spaces into an equivalent one between a different pair of metric spaces using isometries. 

\begin{theorem}[Equivalent Formulation of GLV Problem]
\label{thm4:equ_lv_form}
    For an AE $(e_\theta, g_\theta)_{(X, d_X)\leftrightarrow(Z, d_Z)}$, its \textbf{Generalized Least Volume problem} over the dataset $\gX\subseteq X$ is defined as:
    \begin{align}
        \argmin_\gZ \quad& \vsigma(\gZ)  \label{eq4:vol1} 
        \\
        \text{s.t.} \quad& \gZ \in \mathscr{Z}^\star 
        \coloneqq 
        \left\{\gZ = e_{\theta^\star}(\gX)\mid 
        \theta^\star = \argmin_\theta
        \E_{x\sim\gX}d_X(g_\theta\circ e_\theta(x), x)\right\} \label{eq4:rec1}
        \\
        & d_X(g_\theta(z_1), g_\theta(z_2)) \leq K d_Z(z_1, z_2),\quad \forall\{z_1, z_2\} \subseteq Z \label{eq4:lip1}
    \end{align}
    It has the \textbf{equivalent formulation} below on another autoencoder $(\tde_\theta, \tdg_\theta)_{\left(\tdX, d_\tdX\right)\leftrightarrow\left(\tdZ, d_\tdZ\right)}$ between a different pair of metric spaces $\tdX$ and $\tdZ$ on the transformed dataset $\Tilde{\gX}\coloneqq\phi(\gX)$, provided that we can find the isometries $\psi: Z \to \tdZ$ ($\psi$ is bijective) and $\phi: X\to \tdX$ that makes $Z\cong\tdZ$ and $X\cong\phi(X)$:
    \begin{align}
        \argmin_{\gZ=\psi^{-1}(\Tilde{\gZ})} \quad& \vsigma(\gZ) \label{eq4:vol2}
        \\
        \text{s.t.} \quad& \Tilde{\gZ} \in \Tilde{\mathscr{Z}}^\star 
        \coloneqq 
        \left\{\Tilde{\gZ} = \tde_{\theta^\star}(\Tilde{\gX})\mid 
        \theta^\star = \argmin_\theta
        \E_{\tdx\sim\Tilde{\gX}}d_\tdX(\tdg_\theta\circ \tde_\theta(\tdx), \tdx)\right\} \label{eq4:rec2}
        \\
        & d_\tdX(\tdg_\theta(\tdz_1), \tdg_\theta(\tdz_2)) \leq K d_\tdZ(\tdz_1, \tdz_2),\quad \forall\{\tdz_1, \tdz_2\} \subseteq \tdZ \label{eq4:lip2}
    \end{align}
    These two formulations are equivalent in the sense that their solution sets $[\gZ^\star]$ are equal. 
    \[
    \begin{tikzcd}
    (X, d_X) 
        \arrow[shift left]{r}{e_\theta} 
        \arrow[shift left]{d}{\phi} 
    & (Z, d_Z) \arrow[shift left]{d}{\psi} 
        \arrow[shift left]{l}{g_\theta}
    \\%
    (\tdX, d_\tdX) 
        \arrow[shift left]{r}{\tde_\theta} 
        \arrow[shift left,dashed]{u}{\Bar{\phi}^{-1}}
    & (\tdZ, d_\tdZ) 
        \arrow[shift left]{l}{\tdg_\theta} 
        \arrow[shift left]{u}{\psi^{-1}}
    \end{tikzcd}
    \]
\end{theorem}
\begin{proof}
    Let $\tdg_\theta$ and $\tde_\theta$ be parameterized by $\tdg_\theta=\phi\circ g_\theta \circ \psi^{-1}$ and $\tde_\theta=\psi \circ 
    e_\theta \circ \Bar{\phi}^{-1}$.
    The constraints (\ref{eq4:lip1}) and (\ref{eq4:lip2}) are equivalent because of Lemma~\ref{lemma4:lip}. 
    Consequently, (\ref{eq4:rec2}) is equivalent to (\ref{eq4:rec1}), because 
    \[
    \Tilde{\mathscr{Z}}^\star 
    = \left\{\Tilde{\gZ} = \psi\circ e_{\theta^\star}(\gX)
        \;\middle|\;
        \begin{aligned}
            \theta^\star 
            &= \argmin_\theta
            \E_{x\sim\gX} d_\tdX(\phi(g_\theta\circ e_\theta(x)), \phi(x)) \\
            &= \argmin_\theta \E_{x\sim\gX} d_X(g_\theta\circ e_\theta(x), x)
        \end{aligned}
        \right\}
    \]
    and thus $\Tilde{\mathscr{Z}}^\star = \psi(\mathscr{Z}^\star)$. Therefore, the volumes in (\ref{eq4:vol1}) and (\ref{eq4:vol2}) are identically minimized over the same collection of $\gZ$ satisfying $(\gZ, d_Z)\simeq (\gX, d_X)$.
\end{proof}

\begin{remark}
It is beneficial to have both $d_{\tdZ}$ and $d_{\tdX}$ be $L_2$ distance. 
Not only is $L_2$ distance common and intuitive, but it also allows us to enforce Lipschitz constraint on $\tdg_\theta$ through spectral normalization (\S\ref{sec:spectral_normalization}) easily. Also, the continuity of neural networks are determined in terms of the topologies induced by $L_2$ (or, equivalently, $L_p$) metrics, so we don’t need to worry about the continuity of $\tde_\theta$ and $\tdg_\theta$ in this equivalent LV formulation. Moreover, the latent space $Z$ is typically chosen as an $L_2$ metric space, which makes $\tdZ$ and $\psi$ unnecessary. So we can identify $(Z, \normltwo)$ with $(\tdZ, \normltwo)$ by setting $\psi=\id_Z$.
For these reasons, we will focus on $\tdX$ and $Z$ equipped with $L_2$ metrics, and use all $\tdX$ and $Z$ to denote $(\tdX, \normltwo)$ and $(Z, \normltwo)$ unless otherwise specified. 
\end{remark}

Though generic, Theorem~\ref{thm4:equ_lv_form} is too conceptual for real applications, because it is in general tricky to retrieve the isometries $\psi$ and $\phi$ given the predefined metrics $d_X$ and $d_Z$, not to mention that often the existence of $\phi$ and $\psi$ is also questionable, so we cannot readily obtain $(e_\theta, g_\theta)$ from $(\tde_\theta, \tdg_\theta)$.
However, suppose $Z$ is a $L_2$ metric space\textemdash so $\psi$ is not needed\textemdash while $d_X$ is \emph{not} provided but $\tdX$ is, for which we know there exists a \emph{topological embedding} $\phi: (X, \normltwo)\to (\tdX, \normltwo)$, even if the analytical form of $\phi$ is \emph{unknown} and we only have a corresponding dataset $\Tilde{\gX} = \phi(\gX)\subseteq\tdX$. In this case, we can induce a new metric space $(X, d_X)$ from $\tdX$ using the pullback $\phi^*$, as introduced in Def.~\ref{def:pullback} and Lemma~\ref{lemma4:pulltopo}. Consequently, $\phi:(X, d_X)\to\tdX$ becomes an \emph{isometry}, and because $\phi: (X, \normltwo)\to \tdX$ is \emph{homeomorphic} onto its image, we have $(\gX, \normltwo)\simeq \Tilde{\gX} \cong (\gX, d_X)$ by construction! 

Suppose for this pullback metric $d_X$ we want to solve the LV problem on $(e_\theta, g_\theta)_{(X, d_X)\leftrightarrow Z}$ over the dataset $\gX$. Instead of solving its equivalent on $(\tde_\theta, \tdg_\theta)_{\tdX\leftrightarrow Z}$ as per Theorem~\ref{thm4:equ_lv_form}, we can actually switch to the following alternative on $(e_\theta, \tdg_\theta)_{(X, \normltwo)\to Z\to \tdX}$, thanks to Lemma~\ref{lemma4:gae}.
\begin{corollary}[GLV Problem in Pullback Metric Spaces]
\label{corollary4:glv}
    For the GLV problem on the autoencoder $(e_\theta, g_\theta)_{(X, d_X)\leftrightarrow Z}$ over the dataset $\gX\subseteq X$ where $d_X\coloneqq \phi^* d_\tdX$ is the pullback metric induced by the topological embedding $\phi: (X, \normltwo)\to(\tdX, \normltwo)$, it has the equivalent formulation below on the (generalized) autoencoder $(e_\theta, \tdg_\theta)_{(X, \normltwo)\to Z\to \tdX}$:
    \begin{align}
        \argmin_\gZ \quad& \vsigma(\gZ) 
        \\
        \text{s.t.} \quad& \gZ \in \mathscr{Z}^\star 
        \coloneqq 
        \left\{\gZ = e_{\theta^\star}(\gX)\mid 
        \theta^\star = \argmin_\theta
        \E_{x\sim\gX}\|\tdg_\theta\circ e_\theta(x) - \phi(x)\|_2\right\} \label{eq4:grec}
        \\
        & \|\tdg_\theta(z_1) - \tdg_\theta(z_2)\|_2 \leq K \|z_1- z_2\|_2,
        \quad \forall\{z_1, z_2\} \subseteq Z 
    \end{align}
    \[
    \begin{tikzcd}
    (X, d_X)  
    & (X, \normltwo) 
        \arrow{d}{e_\theta} 
        \arrow[leftrightarrow,swap]{l}{\id_X}
    \\%
    (\tdX, \normltwo) 
        \arrow[harpoon,dashed]{u}{\phi^*}
    & (Z, \normltwo) 
        \arrow{l}{\tdg_\theta}
        \arrow[bend right=0]{ul}[pos=0.25,sloped]{g_\theta}
        \arrow[from=1-2, to=2-1, bend right=10, crossing over]{}[pos=0.6, sloped, swap]{\phi} 
    \end{tikzcd}
    \]
\end{corollary}

\begin{proof}
    Lemma~\ref{lemma4:gae} shows that (\ref{eq4:grec}) makes $\gZ \simeq (\gX,\normltwo) \simeq (\gX, d_X)$, so (\ref{eq4:grec}) is equivalent to (\ref{eq4:rec1}) in Theorem~\ref{thm4:equ_lv_form}. Lemma~\ref{lemma4:pulltopo} allows us to use the same $e_\theta$ for both $(X, d_X)$ and $(X, \normltwo)$, since it is continuous \wrt both topologies. 
    In other words, we identify $e_\theta$ with $e_\theta\circ\id_X$.
\end{proof}

After training, if needed, we can train another model $f_\theta$ to approximate $\Bar{\phi}^{-1}$ and thus obtain the $K$-Lipschitz $g_\theta: Z\to(X, d_X) = f_\theta\circ \tdg_\theta$, albeit often the encoder $e_\theta$ and the representation $\gZ$ are more interesting to us. 
Unlike in Def.~\ref{def:pullback}, here $\phi: (X, \normltwo)\to \tdX)$ must be \emph{homeomorphic} onto its image, as otherwise it may happen that $(\gX, d_X)\not\simeq (\gX,\normltwo)$, such that (\ref{eq4:grec}) in Corollary~\ref{corollary4:glv} does not guarantee $\gZ\simeq(\gX, d_X)$.

\subsection{Least Volume on Labeled Datasets}
\label{sec:lvlabel}
Corollary~\ref{corollary4:glv}, though more practical than Theorem~\ref{thm4:equ_lv_form}, may still appear too abstract to use in practice. \emph{How exactly do we obtain this topological embedding $\phi$ prior to inducing a pullback metric $d_X$?} 
Surprisingly, $\phi$ is quite ubiquitous, and Corollary~\ref{corollary4:glv} in fact lays the foundation for extrapolating GLV to the labeled datasets to supplement the data representations with additional label information. 
To see that, we begin with this important observation~\cite[Example~3.5]{lee2010introduction}:
\begin{lemma}[Graph of a Continuous Function]
\label{lemma4:graph}
    If $f:X\to Y$ is continuous, then its \emph{graph} $\gG(f)$ is homeomorphic to $X$, where $\gG(f)\coloneqq \{(x, y)\mid x\in X, y=f(x)\}$. 
\end{lemma}
\begin{proof}
    Let $\phi:X\to X\times Y = \id_X\times f$ and $\pi:X\times Y \ni x\times y \mapsto x \in X$. Both $\phi$ and $\pi$ are continuous, and $\pi\circ\phi = \id_X$, so $X\simeq\gG(f)\subseteq X\times Y$ (in terms of the product topology).
\end{proof}

Lemma~\ref{lemma4:graph} shows that for a labeled dataset whose \emph{label function} $f$ is continuous \footnote{More rigorously, here $f:X\to Y$ is the  \emph{continuous extension} of the true label function $\restr{f}{\gX}$ that should only be well-defined on the dataset $\gX\subseteq\supp\mdata\subseteq X$. 
We assume that $\restr{f}{\gX}$ is continuous over $\gX$ and adopts a continuous extension ${f}$ over the entire $X$.
}, this \emph{graph function} $\phi=\id_X\times f$ is exactly a topological embedding ready to use!
We can thus set $\Tilde{X} = X\times Y$ and directly apply Corollary~\ref{corollary4:glv} to a \textbf{\emph{labeled dataset}} $\Tilde{\gX}=\gG(\restr{f}{\gX}) \coloneqq \{(x, y)\mid x\in \gX, y=f(x)\}$, 
provided that both the \emph{original} data space $(X, d'_X)$ and the label space $(Y, d_Y)$ are \emph{$L_2$ metric spaces} while $\tdX = {X\times Y}$ is equipped with the \emph{$L_2$ product metric}: 
\[\ell_2((x_1, y_1), (x_2, y_2))
\coloneqq\|(d'_X(x_1, x_2), {d_Y(y_1 - y_2)})\|_2
=\|(x_1, y_1)- (x_2, y_2)\|_2.\] 
Note that the topology of $(X\times Y, \ell_2)$\textemdash \ie, $(\tdX, \normltwo)$\textemdash agrees with the product topology of $(X,\normltwo)\times(Y, \normltwo)$, so $\phi$ is still a topological embedding \wrt $(X\times Y, \ell_2)$. The corollary below formalizes this extrapolation:

\begin{corollary}[Labeled GLV in $L_2$ Product Metric Space]
\label{corollary4:labellv}
    Suppose $\Tilde{\gX} = \gG(\restr{f}{\gX})$ is a labeled dataset where the label function $f$ is continuous. For the autoencoder $(e_\theta, g_\theta)_{(X, d_X)\leftrightarrow Z}$ where $d_X\coloneqq \phi^*\ell_2$ is the label-informed pullback metric induced by $\phi: {(X, \normltwo)}\to {(X\times Y, \ell_2)} = {\id_X \times f}$, its GLV problem over the unlabeled dataset $\gX$ has the equivalent formulation below on the autoencoder $(e_\theta, \tdg_\theta)_{(X, \normltwo)\to Z\to (X\times Y, \ell_2)}$ over the labeled dataset $\Tilde{\gX}$:
    \begin{align}
        \argmin_\gZ \quad& \vsigma(\gZ) 
        \\
        \text{s.t.} \quad& \gZ \in \mathscr{Z}^\star 
        \coloneqq 
        \left\{\gZ = e_{\theta^\star}(\gX)\mid 
        \theta^\star = \argmin_\theta
        \E_{(x,y) \sim \Tilde{\gX}}\|\tdg_\theta\circ e_\theta(x) - (x,y)\|_2\right\}
        \\
        & \|\tdg_\theta(z_1) - \tdg_\theta(z_2)\|_2 \leq K \|z_1- z_2\|_2,
        \quad \forall\{z_1, z_2\} \subseteq Z \label{eq4:lip_label}
    \end{align}
    \[
    \begin{tikzcd}
    (X, d_X) 
        \arrow[leftrightarrow]{r}{\id_X}
    & (X, \normltwo) 
        \arrow[hook',sloped, "\phi" description]{ld}
        \arrow{r}{e_\theta}
    & (Z, \normltwo)
        \arrow[pos=0.25, bend left=0]{dll}[sloped,swap]{\tdg_\theta} 
        \arrow[ll, bend right=20, "g_\theta"]
    \\%
    \renewcommand\arraystretch{0.2}
    \begin{pmatrix}
        X\times Y, \\
        \ell_2
    \end{pmatrix} 
        \arrow[harpoon,dashed]{u}{\phi^*}
    & (Y, \normltwo) 
        \arrow[hook',"\phi" description]{l}
        \arrow[from=u, swap, crossing over]{}[fill=white]{f} 
    \end{tikzcd}
    \]
\end{corollary}
However, the constraint (\ref{eq4:lip_label}) based on $\ell_2$ may not be easy to implement in practice. This is because the data $x\in X$ and the labels $y\in Y$ are usually of different modalities, such that the decoder $\tdg_\theta: Z\to X\times Y$ is normally decomposed into $\tdg_\theta \coloneqq g_\theta^x \times g_\theta^y$ with $g^x_\theta: Z\to X$ and $g^y_\theta: Z\to Y$ of different architectures. The $L_2$ Lipschitz regularizations, such as spectral normalization, can be easily imposed on $g^x_\theta$ and $g^y_\theta$ separately, but it is non-trivial to regularize their product $g_\theta^x \times g_\theta^y$ jointly. 
What happens if we only force $g^x_\theta$ and $g^y_\theta$ to be $K$-Lipschitz \emph{individually}? 
The following shows it is equivalent to replacing $\ell_2$ in Corollary~\ref{corollary4:labellv} with the \emph{$L_\infty$ product metric}: 
\[\ell_\infty((x_1, y_1), (x_2, y_2))
\coloneqq \|(d'_X(x_1, x_2), d_Y(y_1, y_2))\|_\infty
=\max(\|x_1-x_2\|_2, \|y_1 - y_2\|_2).\]

\begin{corollary}[Labeled GLV in $L_\infty$ Product Metric Space]
\label{corollary4:labellvmax}
    Suppose $\Tilde{\gX} = \gG(\restr{f}{\gX})$ is a labeled dataset where the label function $f$ is continuous. For the autoencoder $(e_\theta, g_\theta)_{(X, d_X)\leftrightarrow Z}$ where $d_X \coloneqq \phi^*\ell_\infty$ is the label-informed pullback metric induced by $\phi: {(X, \normltwo)}\to {(X\times Y, \ell_\infty)} = {\id_X \times f}$, its GLV problem over the unlabeled dataset $\gX$ has the equivalent formulation below on the autoencoder $(e_\theta, \tdg_\theta)_{(X, \normltwo)\to Z\to (X\times Y, \ell_\infty)}$ over the labeled dataset $\Tilde{\gX}$, where $\tdg_\theta \coloneqq g^x_\theta \times g^y_\theta$ with $g^x_\theta: Z\to (X, \normltwo)$ and $g^y_\theta: Z\to (Y, \normltwo)$:
    \begin{align}
        \argmin_\gZ \quad& \vsigma(\gZ) 
        \label{eq4:volinfty}
        \\
        \text{s.t.} \quad& \gZ \in \mathscr{Z}^\star 
        \coloneqq 
        \left\{\gZ = e_{\theta^\star}(\gX)\mid 
        \theta^\star = \argmin_\theta 
        \E_{(x,y)\sim \Tilde{\gX}}
        \|\tdg_\theta\circ e_\theta(x) - (x,y)\|_2
        \right\} \label{eq4:recinfty}
        \\
        & 
        \forall\{z_1, z_2\} \subseteq Z, \quad 
        \begin{cases}
           \|g^x_\theta(z_1) - g^x_\theta(z_2)\|_2 \leq K \|z_1- z_2\|_2 \\
            \|g^y_\theta(z_1) - g^y_\theta(z_2)\|_2 \leq K \|z_1- z_2\|_2
        \end{cases} \label{eq4:lipinfty}
    \end{align}
    \[
    \begin{tikzcd}
    (X, d_X) 
        \arrow[leftrightarrow]{r}{\id_X}
    & (X, \normltwo) 
        \arrow[swap]{d}{f} 
        \arrow[hook', sloped, "\phi" description]{ld}
        \arrow[shift left]{r}{e_\theta}
    & (Z, \normltwo)
        \arrow[shift left]{l}{g^x_{\theta}}
        \arrow{dl}[sloped, swap]{g^y_\theta} 
        \arrow[ll, bend right=20, "g_\theta"]
    \\%
    \renewcommand\arraystretch{0.2}
    \begin{pmatrix}
        X\times Y, \\
        \ell_\infty
    \end{pmatrix} 
        \arrow[harpoon,dashed]{u}{\phi^*}
    & (Y, \normltwo) 
        \arrow[hook',"\phi" description]{l}
    \end{tikzcd}
\]
\end{corollary}
\begin{proof}
    (\ref{eq4:lipinfty}) is equivalent to
    $\max\left(\|g^x_\theta(z_1) - g^x_\theta(z_2)\|_2, \|g^y_\theta(z_1) - g^y_\theta(z_2)\|_2\right) \leq K \|z_1- z_2\|_2$.
    Moreover, because the $\ell_\infty$ topology equals the product topology,
    $\phi$ is still a topological embedding,
    while $g^x_\theta: Z\to (X, \normltwo)$ and $g^y_\theta: Z\to (Y, \normltwo)$ being continuous makes $\tdg_\theta: Z\to(X\times Y, \ell_\infty)$ continuous.
    At last, the $\ell_2$ loss in (\ref{eq4:recinfty}) can replace $\ell_\infty$, as they both make $\tdg_\theta\circ e_\theta(x) = (x,y)$ when reaching 0.
\end{proof}

\paragraph{Intuition} 
This practical labeled GLV formulation in Corollary~\ref{corollary4:labellvmax} should help us learn meaningful latent representations that an unlabeled LV problem cannot achieve. Inspired by the intuition in \S\ref{sec:methodology}, we may expect the label information to additionally deform the homeomorphic latent set $\gZ$ like this: it drives the data samples of different labels farther away from each other\textemdash as otherwise the $K$-Lipschitz constraint on $g^y_\theta$ will be violated\textemdash while forcing the samples of similar labels to stay as close to each other as possible\textemdash as otherwise it increases the volume unnecessarily. This label-induced deformation should become more prominent if we scale up $d_Y$ with some factor $c$ to make $c\cdot d_Y$ dominate $d'_X$. It may have distinct appearances on labeled datasets with different topological structures:
\begin{itemize}
    \item If $\gY\coloneqq f(\gX)$ is \emph{discrete} (\ie, categorical) while $f$ is continuous, then the subsets $f^{-1}(y)$ of different categories in $\gX$ must be disconnected from each other. GLV's regularization should make the corresponding latent subsets $e(f^{-1}(y))$ of different categories $y$ stay far away from each other in $Z$, while forcing the latent codes $z$ in each $e(f^{-1}(y))$ to come close to each other. This indicates that labeled GLV could be a form of supervised (or semi-supervised) contrastive learning~\citep{le2020contrastive, chen2020simple, khosla2020supervised}.
    
    \item If $\gY$ is \emph{connected} (\eg, it is the set of performance values of some designs), then in the latent space $Z$, GLV may help scale up the data dimensions to which $y$ is sensitive (\ie, $f$ has larger Lipschitz constant along these dimensions), while shrinking down those dimensions trivial to $f$. For example, in the study of a design dataset and its performance values (as seen in \S\ref{sec:experiments_slva_airfoil}), such an effect could help extract a few design parameters that are crucial to design optimization and accelerate the design process if we only focus on them. 
    
\end{itemize}

\section{Implementation}
\label{sec:implementation}

The two constrained optimization problems\textemdash the unlabeled LV problem (\ref{eq4:lvvol}, \ref{eq4:lvrec}, \ref{eq4:lip}) and the labeled GLV problem (\ref{eq4:volinfty}, \ref{eq4:recinfty}, \ref{eq4:lipinfty})\textemdash cannot be solved exactly. On one hand, the reconstruction constraints (\ref{eq4:lvrec}) and (\ref{eq4:recinfty}) are hard to enforce due to the complexity of  neural networks; on the other hand, minimizing the latent STD vector $\vsigma(\gZ)$ under the volume preorder $\lesssim$ is also challenging, given that in practice we need to numerically evaluate the volume $\prod_{\sigma_i > 0} \sigma_i$ over latent dimensions with $\sigma_i(\gZ) \gg 0$ and ignore the dimensions with $\sigma_i(\gZ) \approx 0$, but the unary relations $\gg 0 $ and $\approx 0 $ are not well-defined yet. 
In this section, we present two methods for obtaining approximate solutions of least volume problems\textemdash the simpler \emph{na\"ive volume penalty} and the more reliable \emph{volume minimization with dynamic pruning}. At the end of this section, we also provide details about enforcing the Lipschitz constraints.  

\subsection{Na\"ive Volume Reduction}
\label{sec:naive_volume_reduction}
The recursive process devised in \S\ref{sec:vol_pen} is non-trivial to implement. Na\"ively, we may circumvent this hurdle by resorting to the weighted sum
\begin{equation}
    L = J + \lambda\cdot L^\eta_\text{vol}
    = J + \lambda\cdot \sqrt[\uproot{1} n]{\prod_{i} (\sigma_i(\gZ) + \eta)}
    \label{eq4:naive_loss}
\end{equation}
to perform the constrained optimization approximately, where $J$ is the reconstruction loss in either (\ref{eq4:lvrec}) or (\ref{eq4:recinfty}), $L^\eta_\text{vol}$ is the \emph{na\"ive volume penalty} augmented with a hyperparameter $\eta > 0$ that mitigates vanishing gradient when $\sigma_i(\gZ) \approx 0$, $n$ denotes $\dim Z$, and $\lambda$ is the weight coefficient making the trade-off between the reconstruction quality and the degree of volume reduction. We can evaluate $L^\eta_\text{vol}$ using the ExpMeanLog trick to avoid numerical issues.

Minimizing this penalty not only equivalently minimizes an upper bound of the volume, but also naturally interpolates between $\sqrt[\leftroot{-3}\uproot{3}n]{\prod_i\sigma_i}$ and $\frac{1}{n}\|\pmb{\sigma}\|_1$ gradient-wise, given that as $\eta\rightarrow\infty$:
\begin{equation}
    \nabla_\theta \sqrt[\leftroot{-3}\uproot{3}n]{\prod_i (\sigma_i + \eta)}
    =
    \frac{1}{n}\sum_i\frac{
        \sqrt[\leftroot{-3}\uproot{3}n]{\prod_i (\sigma_i + \eta)}
    }{
        \sigma_i + \eta
    }\cdot \nabla_\theta \sigma_i
    \quad\longrightarrow\quad
    \frac{1}{n}\sum_i\nabla_\theta \sigma_i
    =
    \nabla_\theta \frac{1}{n}\|\pmb{\sigma}\|_1
    \label{eqn:interpolation}
\end{equation}

\noindent So we can seamlessly shift from volume penalty to $L_1$ penalty by increasing $\eta$. 
We can see that the lower the $\eta$, the more the regularizer's gradient prioritizes minimizing smaller $\sigma_i$, which intuitively should make $\pmb{\sigma}$ sparser than the $L_1$ penalty does that scales every $\nabla_\theta\sigma_i$ equally. We shall see in \S\ref{sec:experiments} that $\sqrt[\leftroot{-3}\uproot{3}n]{\prod_i (\sigma_i + \eta)}$ is indeed more efficient in practice.

\subsubsection{Limitations}
This na\"ive volume penalty $L^\eta_\text{vol} = \sqrt[\uproot{1}n]{\prod_i (\sigma_i(\gZ) + \eta)}$ is indeed straightforward to evaluate, but it incurs several drawbacks.
First, the hyperparameter $\eta \geq 0$ is introduced to overcome the vanishing gradient issue caused by the group of $\sigma_i\approx 0$ that emerge as $\sqrt[\uproot{1}n]{\prod_i \sigma_i(\gZ)}$ decreases, which stagnates $\|\vsigma\|_0$'s reduction.  
$\eta$ helps $L^\eta_\text{vol}$ interpolate between $\sqrt[\uproot{1}n]{\prod_i \sigma_i(\gZ)}$ and $\frac{1}{n}\|\vsigma(\gZ)\|_1$ gradient-wise, so by setting a large enough $\eta$, we can provide those latent dimensions of large $\sigma_i$ with enough gradient to keep compressing themselves to reduce $\|\vsigma\|_0$ further. 
However, it is hard to tell whether an $\eta$ is ``large enough'' without some heuristic determination. In addition, when $\eta > 0$, minimizing $L^\eta_\text{vol}$ is not equivalent to minimizing the actual volume $\prod_i \sigma_i(\gZ)$. As $\eta$ increases, $L^\eta_\text{vol}$ behaves more like $\|\vsigma(\gZ)\|_1$ and the $\|\vsigma\|_0$ reduction performance deteriorates . 
    
In addition, $L^\eta_\text{vol}$ minimization is a makeshift for the conceptual \emph{recursion of subspace compression and extraction} in \S\ref{sec:methodology}. Without the recursive removal of the dummy latent dimensions with $\sigma_i\approx 0$, we have to perform the volume reduction in the whole $n$-D latent space and keep compressing those already annihilated dummy dimensions, rather than only focusing on those dimensions with $\sigma_i \gg 0$. We will show in \S\ref{sec:ldembed} that this rigidity causes the dimension reduction result of this na\"ive volume minimization to depend on the choice of the latent space dimension $n$, and to stagnate especially when $n$ significantly exceeds the dataset's embedding dimension.

\subsection{Volume Reduction with Dynamic Pruning Algorithm}
\label{sec:dpa}

    
To overcome these issues and realize the recursive process, we propose the following \emph{Dynamic Pruning} (\emph{DP}) algorithm to reduce the actual \emph{volume penalty} $L_\text{vol}\coloneqq \sqrt[\uproot{1}|\gI|]{\prod_{i\in\gI} \sigma_i(\gZ)}$, which is evaluated on the set of non-trivial latent dimensions $\gI\coloneqq\{i\mid \sigma_i(\gZ) \gg 0\}$ to reduce $\vsigma$ \wrt the volume preorder $\lesssim$. Here in $\gI$ we use $\gg 0$ instead of the $>0$ in Def.~\ref{def:volumeorder} because we cannot make $\sigma_i$ strictly zero in practice. We will provide an accurate definition of $\gg 0$ and its negation $\approx 0$ soon in Def.~\ref{def:approx0}. $L_\text{vol}$ replaces $L^\eta_\text{vol}$ in (\ref{eq4:naive_loss}) and gives the new training loss 

\begin{equation}
    L = J + \lambda\cdot L_\text{vol}
    = J + \lambda\cdot \sqrt[\uproot{1}|\gI|]{\prod_{i\in\gI} \sigma_i(\gZ)}
    \label{eq4:loss}
\end{equation}

As its name suggests, DP is built for dynamically pruning out the annihilated latent dimensions\textemdash \ie, the ones with $\sigma_i\approx 0$\textemdash when we minimize the loss (\ref{eq4:loss}) during the training of autoencoders, so that the set of indices ${\gI\coloneqq \{i\mid \sigma_i(\gZ) \gg 0\}}$ stays updated for extracting the least dimensional latent space $Z_\gI\coloneqq \bigtimes_{i\in\gI} Z_i$ where the flattened $\gZ$ currently lies. Therefore, the kernel of DP is essentially a criterion against which we can judge if a given latent dimension satisfies $\sigma_i\approx 0$ rather than $\sigma_i \gg 0$ and thus should be removed from $\gI$. 

Setting a small threshold $\delta>0$ for $\sigma_i$ to determine if $\sigma_i \approx 0$ ($\Leftrightarrow \sigma_i < \delta$) seems like a promising strategy. However, it is easier said than done. This is because the length scale of $\gZ$ not only varies across different datasets and autoencoders, but also is consistently changing throughout the training process, so a constant threshold $\delta$ that reasonably indicates small $\sigma_i$ values in one setting often fails in another. It is thus crucial to make $\approx 0$ and $\gg 0$ independent of $\gZ$'s length scale. 

\begin{figure}[hbt!]
    \centering
    \includegraphics[width=\linewidth]{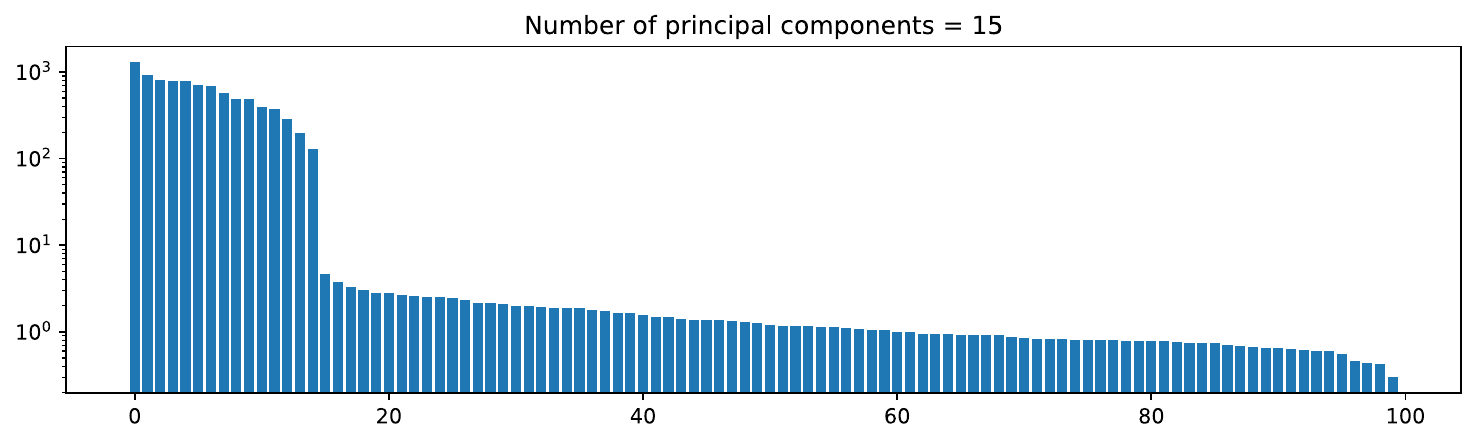}
    \caption{Typical plummet of $\sigma_i$ in LV problems.}
    \label{fig4:latent_plummet}
\end{figure}

Fortunately, we can build a length-scale-independent \emph{$\vrho$-criterion} based on the following phenomenon observed in~\citep{qiuyi2024compressing, chen2025bayesian}: For the LV problem on every dataset being tested, after minimizing the volume $\prod_{i=1}^n \sigma_i(\gZ)$ over the entire $n$-D latent space, if we \emph{reindex} all latent dimension \wrt $\sigma_i(\gZ)$ in descending order\textemdash such that $\sigma_i \geq \sigma_j$ for $i < j$\textemdash then plot them out in a bar plot and put the y-axis on log scale, there is always a noticeable plummet right after a given index $k$, as illustrated in Fig.~\ref{fig4:latent_plummet}. In other words, this $k$ is the index where $\sigma_i/\sigma_{i+1}$ is the greatest. 
Most latent dimensions after this dimension $k$ can be regarded as trivial, given their orders of magnitude smaller $\sigma_i$ relative to $\sigma_k$, and empirically also much lower influence on the reconstruction error when pruned, which is supported by Theorem~\ref{thm3:safety}. Therefore, after retrieving this $k$, we should use the relative magnitude $\sigma_i / \sigma_k$ to determine if $\sigma_i\approx 0$. Formally speaking:
\begin{definition}(Approximately Zero by $\vrho$-criterion)
\label{def:approx0}
    After specifying $\delta > 0$ and permuting $\vsigma\in\sR^n_+$ into $\bm{\varsigma}$ such that $\varsigma_i \geq \varsigma_j$ for $i < j$, we define the unary relations $\approx 0$ and $\gg 0$ on $\sigma_i$ as 
\begin{equation}
    \sigma_i \approx 0 
    \quad\Leftrightarrow\quad  
    \lnot\;(\sigma_i \gg 0)
    \quad\Leftrightarrow\quad  
    \rho_i\coloneqq \sigma_i / \varsigma_k < \delta
    \;\text{ for }\;
    k = \argmax_i \varsigma_i / \varsigma_{i+1}.
\end{equation}
\end{definition}

After $\gI$ is determined along with its complement $\gI^\complement$, we can define the pruning operator $p_P:Z\to Z_{\gI}\times \Bar{\rvz}_{\gI^\complement}$ by ${[p_P(z)]_i = \begin{cases}z_i & i\in \gI \\ \bar{z}_i & i\in \gI^\complement \end{cases}}$ together with the projection $\pi: Z_{\gI}\times \Bar{\rvz}_{\gI^\complement} \mapsto Z_{\gI}$ as per \S\ref{sec:safety}, where each element $\Bar{z}_i$ of $\Bar{\rvz}_{\gI^\complement}$ equals the \emph{mean} $\mu_i(\gZ)$ \emph{frozen at the moment when $i$ is included in $\gI^\complement$}. 
The autoencoder ${(\pi\circ p_P\circ e_\theta, g_\theta \circ \pi^{-1})}$ is then the desired one mapping between the data space $X$ and the least dimensional latent space $Z_\gI$. Here $p_P: Z\to Z_{\gI}\times \Bar{\rvz}_{\gI^\complement}$ and $\pi^{-1}:Z_\gI \to Z_{\gI}\times \Bar{\rvz}_{\gI^\complement}$ allow us to ignore the trivial dimensions $i$ in $\gI^\complement$ when training $e_\theta$, since all of them only outputs and intakes the constant $\bar{z}_i$ by construction once included in $\gI^\complement$. For simplicity, hereafter when discussing the \emph{least volume autoencoders trained with dynamic pruning} (LVAE-DPs), we use $e_\theta$ and $g_\theta$ to refer to $\pi\circ p_P\circ e_\theta$ and $g_\theta \circ \pi^{-1}$, and $Z$ to refer to $Z_\gI$, unless otherwise specified.

\begin{algorithm}[bt!]
\caption{Least Volume Autoencoder with Dynamic Pruning}\label{alg:dp}
\begin{algorithmic}
\Require $N > 0$, $\lambda > 0$, $\beta\in(0, 1]$
\Comment{epochs,  weight of volume, weight for moving average}

\Require $\delta > 0$, $N_p\geq 0$ 
\Comment{tolerance, starting epoch of pruning}

\Ensure $g_\theta$ is $K$-Lipschitz. 
\Comment{via spectral normalization}

\Procedure{Update}{$\Bar{\vmu}, \Bar{\vsigma}, \Bar{\vrho}; \gZ, \beta$}
\Comment{update moving average}
    \State $\Bar{\vmu} \gets \beta\Bar{\vmu} + (1-\beta)\vmu({\gZ})$
    \Comment{mean}
    \State $\Bar{\vsigma} \gets \beta\Bar{\vsigma} + (1-\beta)\vsigma({\gZ})$
    \Comment{standard deviation}
    \State $\Bar{\vrho} \gets \beta\Bar{\vrho} + (1-\beta)$\Call{$\vrho$}{$\bar{\vsigma}$}
    \Comment{relative magnitude}
\EndProcedure

\Procedure{DynamicPruning}{$\mathbf{p}, \Bar{\rvz}; \Bar{\vmu}, \Bar{\vrho}, \delta$}
\Comment{update $\gI$ and $\bar{\rvz}_{\gI}$ for pruning}
\For{$i \in \gI \coloneqq \{i\mid p_i = 0\}$}
    \If{$\Bar{\rho}_i < \delta$}
    \Comment{check if $\sigma_i \approx 0$ by $\vrho$-criterion}
        \State $p_i \gets 1$
        \Comment{turn on the switch}
        \State $\Bar{z}_i \gets \Bar{\mu}_i$
        \Comment{log and freeze the mean}
    \EndIf
\EndFor
\EndProcedure

\Function{$\vrho$}{$\vsigma$}
\Comment{return $\vrho$ to be compared to $\delta$}
  \State $\bm{\varsigma} \gets \text{sort}(\vsigma_\gI)$ where $\gI \coloneqq \{i\mid p_i = 0\}$
  \Comment{sort: in descending order}
  \State $k \gets \argmax_i \varsigma_i / \varsigma_{i+1}$
  \State\Return $\vsigma / \varsigma_k$
\EndFunction

\Function{$p_P$}{$\gZ; \mathbf{p}, \Bar{\rvz}$}
\Comment{prune latent codes}
  \For{$i \not\in \gI \coloneqq \{i\mid p_i = 0\}$}
    \State $\gZ_i \gets \Bar{z}_i$
    \Comment{replace all dimensions in $\gI^\complement$ with their means}
  \EndFor
    \State\Return $\gZ$ 
\EndFunction

\Statex
\Procedure{LVAE-DP}{$N, \lambda, \beta, \delta, N_p$; $e_\theta$, $g_\theta$, $\gX$}
\Comment{train Least Volume AE with DP}
    \State $\mathbf{p} \gets \mathbf{0}, \Bar{\rvz} \gets \mathbf{0}$
    \Comment{pruning switches, $\Bar{\rvz}_{\gI^\complement}$ storage}
    \State $\Bar{\vmu} \gets \mathbf{0}, \Bar{\vsigma} \gets \mathbf{0}, \Bar{\vrho} \gets \mathbf{0}$
    \Comment{moving average of $\vmu(\gZ)$, $\vsigma(\gZ)$, $\vrho(\gZ)$}
    \While{$j < N$}
    \Comment{epoch number}
      \State $\gX^{b} \sim \gX$
      \Comment{sample a batch of $b$ samples}
      \State $\gZ^b \gets$ \Call{$p_P$}{$e_\theta(\gX^b); \mathbf{p}, \Bar{\rvz}$} 
      \Comment{encoding followed by pruning}
      \State $\epsilon \gets L_\text{rec}(g_\theta(\gZ^b), \gX^b)$
      \Comment{reconstruction error}
      \State $v \gets \frac{|\gI|}{n} \cdot \sqrt[\uproot{1}|\gI|]{\prod_{i\in\gI} \sigma_i(\gZ^b)}, \quad\gI \coloneqq \{i\mid p_i = 0\}$
      \Comment{scaled volume penalty}
      \State $\theta \gets \Call{GradientDescent}{\theta; \nabla_\theta(\epsilon + \lambda \cdot v)}$ 
      
      \Statex
      \State $\Call{Update}{\Bar{\vmu}, \Bar{\vsigma}, \Bar{\vrho}; \gZ^b, \beta}$
      \Comment{update moving average}
      \If{$j \geq N_p$}
      \Comment{avoid premature pruning}
        \State $\Call{DynamicPruning}{\mathbf{p}, \Bar{\rvz}; \Bar{\vmu}, \Bar{\vrho}, \delta}$
        \Comment{decide which dimensions to prune}
      \EndIf
    \EndWhile
\EndProcedure
\end{algorithmic}
\end{algorithm}

Algorithm~\ref{alg:dp} presents the training process of LVAE-DPs based on Def.~\ref{def:approx0}. Typically, autoencoders are trained with mini-batch gradient descent, so we first introduce the \emph{Update} procedure to track the moving average of $\vmu$, $\vsigma$ and $\vrho$ of $\gZ$ over the mini-batch $\gZ^b$ to estimate their groundtruth values. The hyperparameter $\beta$ can be set to 0.9 to mitigate the disturbance caused by outliers. The \emph{DynamicPruning} algorithm updates $\gI$ based on the $\vrho$-criterion we set, and uses $\bar{\rvz}$ to record the frozen mean values $\mu_i$ of the to-be-pruned trivial latent dimensions. The pruning function $p_P$ uses the up-to-date $\gI$ and $\bar{\rvz}$ to prune the latent set. Here we scale the volume penalty $L_\text{vol}$ with the factor $\frac{|\gI|}{n}$ to unify the magnitude of $L_\text{vol}$'s gradient throughout pruning, given that $\nabla_\theta \sqrt[\leftroot{-3}\uproot{3}|\gI|]{\prod_i \sigma_i}
    =
    \frac{1}{|\gI|}\sum_i\frac{
        \sqrt[\leftroot{-3}\uproot{3}|\gI|]{\prod_i \sigma_i}
    }{
        \sigma_i
    }\cdot \nabla_\theta \sigma_i$. 
The threshold parameter $\delta$ is by default set to 2\%. Increasing it can lead to a more aggressive removal of the trivial latent dimensions. This could be useful if the dataset is noisy, as it should help to remove many latent dimensions that only correspond to meaningless noise. Due to Theorem~\ref{thm3:safety}, removing trivial latent dimensions does not noticeably affect the reconstruction error $J$. This allows the algorithm to proceed \emph{seamlessly}\textemdash \ie, without suspending volume reduction to counteract a surge in reconstruction loss.

\subsection{Enforcing Lipschitz Continuity via Spectral Normalization}
\label{sec:spectral_normalization}

The Lipschitz continuity in constraints (\ref{eq4:lip}) and (\ref{eq4:lipinfty}) can be conveniently enforced via \emph{spectral normalization}~\citep{miyato2018spectral}, namely normalizing the spectral norm of all linear layers to 1. For accurate regularization of any convolutional layers, we employ the power method in~\citep[Algorithm~1]{gouk2021regularisation}. The Lipschitz constant may also be enforced to some degree through gradient regularization based on the decoder's Jacobian-vector product\textemdash \eg, in \citep{gulrajani2017improved}, but this is both inefficient and inaccurate~\citep{miyato2018spectral}.
Combining these spectrally normalized linear layers with 1-Lipschitz activation functions such as LeakyReLU and Sigmoid, we can prevent the decoder's Lipschitz constant $K$ from exceeding 1~\citep[\S3]{gouk2021regularisation}. If for some reason we need to choose a $K \neq 1$\textemdash for instance, to facilitate the reconstruction error reduction on a dataset of large length scale\textemdash we can adjust it by simply attaching a pointwise scaling function $h(x)= K\cdot x$ to the 1-Lipschitz decoder $g$ and turn it into $g_\text{new} = g\circ h$.
\section{Related Works}
\label{sec:related}
The usual way of encouraging the latent set to be low dimensional is sparse coding~\citep{bengio2013representation}, which applies sparsity penalties like LASSO~\citep{tibshirani1996regression, ng2004feature, lee2006efficient}, Student's t-distribution~\citep{olshausen1996emergence, ranzato2007sparse}, KL divergence~\citep{le2011optimization, ng2011sparse}, \etc, over the latent code vector to induce as many zero-features as possible. However, as discussed in \S\ref{sec:lvf}, a latent representation transformed by a homeomorphism $h$ is equivalent to the original one in the sense that $g_\theta\circ e_\theta(x) = (g_\theta\circ h^{-1})\circ(h\circ e)(x)$, so translating the sparse-coding latent set arbitrarily\textemdash which makes the zero-features no longer zero\textemdash provides us an equally good representation. This equivalent latent set is then one that has zero-STDs along many latent dimensions, which is what this work tries to construct. Yet $h$ is not restricted to translation. For instance, rotation is also homeomorphic, so rotating that flat latent set also gives an equivalently good representation. IRMAE~\citep{jing2020implicit} can be regarded as a case of this, in which the latent set is implicitly compressed into a linear subspace not necessarily aligned with the latent coordinate axes. There are also stochastic methods like K-sparse AE~\citep{makhzani2013k} that compress the latent set by randomly dropping out inactive latent dimensions during training.

\begin{table}[hbt!]
  \caption{Comparison of Methods that Automatically Reduce Autoencoder Latent Dimensions}
  \centering
  \small
  \label{tab:comparision}
  \begin{tabular}{c|c|cccc}
    \toprule
    Method & Least Volume & Nested Dropout & PCA-AE & IRMAE & K-sparse AE\\
    \midrule
     Deterministic? & \cmark  & \xmark & \cmark & \cmark & \xmark\\
     \midrule
     \makecell[c]{Nonlinear AE?} & \cmark  & \cmark & \cmark & \cmark & \xmark\\
     \midrule
     \makecell[c]{Penalty Term?} & \cmark  & \xmark & \xmark & \xmark & \xmark\\
    \midrule
     \makecell[c]{Single-Stage Training?} & \cmark    & \cmark & \xmark & \cmark & \cmark\\
    \midrule
      \makecell[c]{Importance Ordering?}  & \cmark    & \cmark & \cmark & \xmark & \xmark\\
    \midrule
      \makecell[c]{Incorporating Label Info?}  & \cmark    & \xmark & \xmark & \xmark & \xmark\\
    \bottomrule
  \end{tabular}
\end{table}

People also care about the information content each latent dimension contains, and hope the latent dimensions can be ordered by their degrees of importance. 
Nested dropout \citep{rippel2014learning} is a probabilistic way that makes the information content in each latent dimension decrease as the latent dimension index increases. 
PCA-AE~\citep{pham2022pca} achieves a similar effect by gradually expanding the latent space while reducing the covariance loss.
Our work differs given that least volume is deterministic, requiring no multistage training, and we only require the information content to decrease with the STD of the latent dimension instead of the latent dimension's index number. Some researchers have also investigated the relationship between PCA and linear autoencoders~\citep{rippel2014learning, plaut2018principal, kramer1991nonlinear}.
In recent years, more researchers have started to look into preserving additional information on top of topological properties in the latent space~\citep{moor2020topological, trofimov2023learning, gropp2020isometric, chen2020learning, yonghyeon2021regularized, nazari2023geometric}, such as geometric information. A detailed methodological comparison between least volume and some aforementioned methods is given in Table~\ref{tab:comparision} to illustrate its distinction.


\section{Experiments: Na\"ive Least Volume Reduction}
\label{sec:experiments}
In this section, we employ the \emph{na\"ive} volume reduction in \S\ref{sec:naive_volume_reduction} to illustrate the characteristics of volume reduction before moving on to the more thorough applications and analyses of LVAE-DPs in \S\ref{sec:exp_ulva} and \S\ref{sec:exp_slva}. 
We examine LV's dimension reduction and importance ordering effect on toy problems and benchmark image datasets.
The technical details of dataset generations, model architectures, ablation study and hyperparameter tuning can be found in the published version~\citep{qiuyi2024compressing}.

\subsection{Toy Problems}
\begin{figure}[hbt!]
     \centering
     \begin{subfigure}{0.3\textwidth}
         \centering
             \begin{subfigure}[b]{\textwidth}
                 \centering
                \renewcommand\thesubfigure{a.\roman{subfigure}}
                 \includegraphics[width=\textwidth]{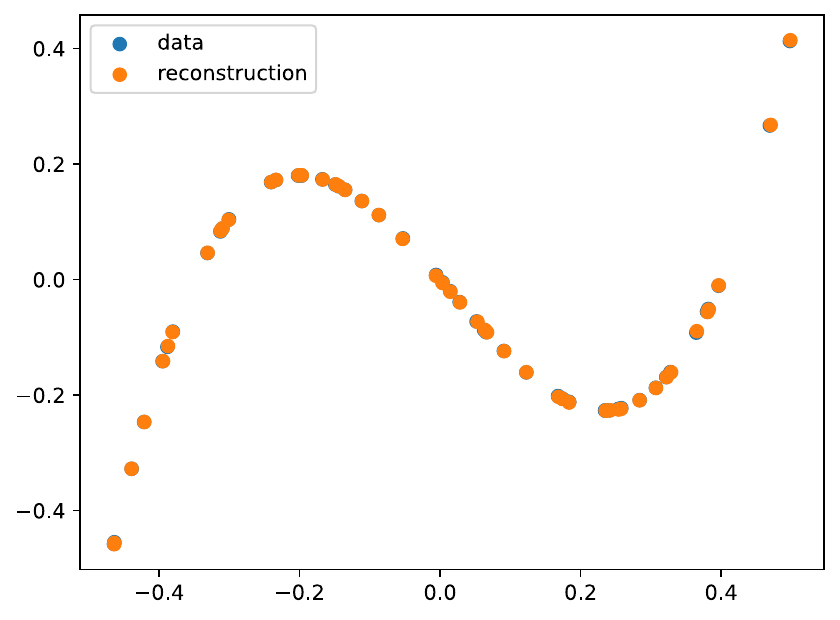}
                 \caption{Reconstruction in 2D}
                 \label{fig3:rec1}
            \end{subfigure}
            \begin{subfigure}{\textwidth}
                 \centering
                 \renewcommand\thesubfigure{a.\roman{subfigure}}
                 \includegraphics[width=\textwidth,  trim={0 0 0 -0.7in},clip]{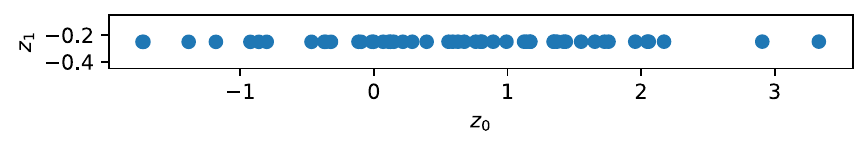}
                 \caption{Latent Space}
                 \label{fig3:lat1}
             \end{subfigure}
             
        \setcounter{subfigure}{0}
         \caption{1D Manifold in 2D Space}
         \label{fig3:mani1}
     \end{subfigure}
     \hfill
     \begin{subfigure}{0.68\textwidth}
         \centering
             \begin{subfigure}{0.45\textwidth}
                 \centering
                 \setcounter{subfigure}{0}
                \renewcommand\thesubfigure{b.\roman{subfigure}}
                 \includegraphics[width=\textwidth, trim={-0.5in 0 -0.5in 0},clip]{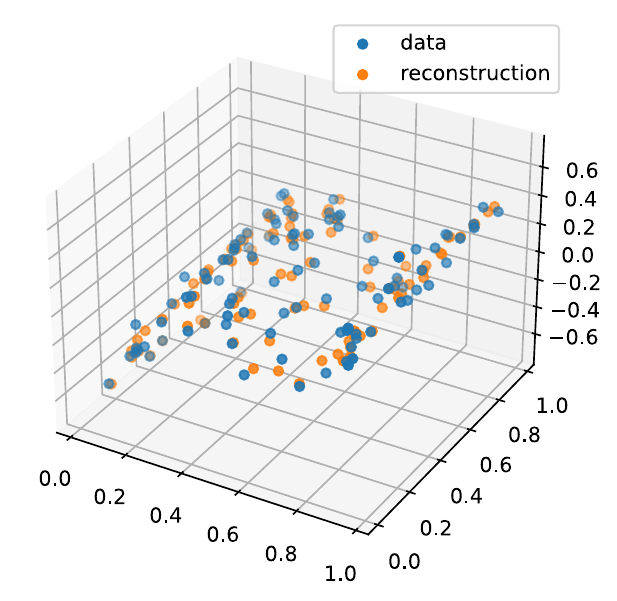}
                 \caption{Reconstruction in 3D}
                 \label{fig3:rec2}
             \end{subfigure}
            \hspace{0.1in}
             \centering
             \begin{subfigure}{0.45\textwidth}
                 \centering
                 \renewcommand\thesubfigure{b.\roman{subfigure}}
                 \includegraphics[width=\textwidth]{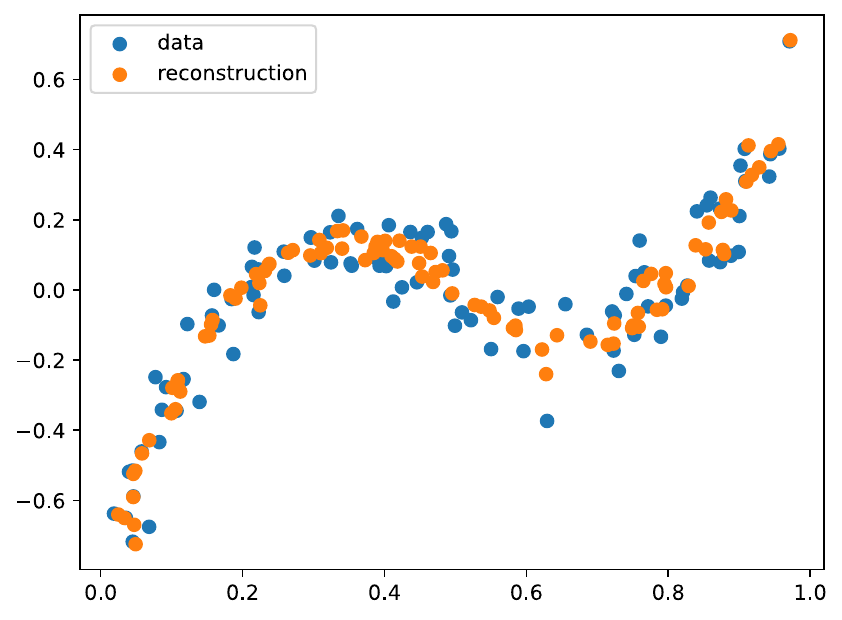}
                 \caption{Projection along Y Axis}
                 \label{fig3:prj2}
             \end{subfigure} \\
             \hfill
             \centering
             \begin{subfigure}{0.45\textwidth}
                 \centering
                 \renewcommand\thesubfigure{b.\roman{subfigure}}
                 \includegraphics[width=\textwidth]{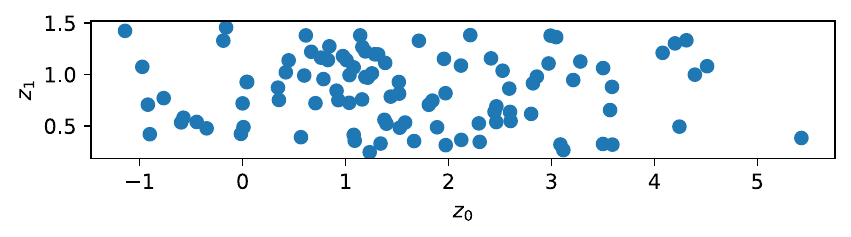}
                 \caption{Latent Subspace 0-1}
                 \label{fig3:lat0-1}
             \end{subfigure}
             \hfill
             \centering
             \begin{subfigure}{0.45\textwidth}
                 \centering
                 \renewcommand\thesubfigure{b.\roman{subfigure}}
                 \includegraphics[width=\textwidth]{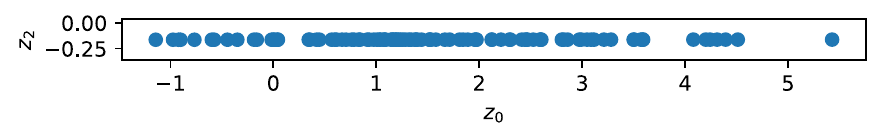}
                 \caption{Latent Subspace 0-2}
                 \label{fig3:lat0-2}
             \end{subfigure}
        \setcounter{subfigure}{1}
        \caption{Noisy 2D Manifold in 3D Space}
        \label{fig3:mani2}
        \end{subfigure}
        \caption{Least volume on low dimensional toy problems.}
        \label{fig3:exp1}
\end{figure}

We first apply least volume to low-dimensional toy datasets to pedagogically illustrate its effect. In each case, we make the latent space dimension equal to the data space dimension. Figure~\ref{fig3:exp1} shows that the autoencoders regularized by least volume successfully recover the low dimensionality of the 1D and 2D data manifold respectively in the latent spaces by compressing the latent sets into low dimensional latent subspaces, without sacrificing the reconstruction quality. Not only that, with a large enough weight $\lambda$ for the volume penalty, in the noisy 2D problem, the autoencoder also manages to remove the data noise perpendicular to the 2D manifold (Fig.~\ref{fig3:prj2}). 

\subsection{Image Datasets}
\label{sec:img_exp}

In this experiment, we compare the performance of different latent regularizers that have the potential of producing a compressed latent subspace. Specifically, these four regularizers are: $L_1$ norm of the latent code (denoted by ``lasso''), $L_1$ norm of the latent STD vector (denoted by ``l1''), na\"ive volume penalty $L^\eta_\text{vol}$ with $\eta = 1$ (denoted by ``vol'' or ``vol\_e1.0'') and the one in \cite{ranzato2007sparse} based on student's t-distribution (denoted by ``st''). We activate the spectral normalization on the decoder and apply these four regularizers to a 5D synthetic image dataset of moving circles~\citep{qiuyi2024compressing}, the MNIST dataset~\citep{deng2012mnist}, the CelebA dataset~\citep{liu2015deep} and the CIFAR-10 dataset~\citep{krizhevsky2014cifar}, then investigate how good they are at reducing the latent set's dimensionality without sacrificing reconstruction performance, and whether there is high correlation between the latent STD and the degree of importance. 
For these image datasets, we use latent spaces of 50, 128, and 2048 dimensions, respectively. For each case, we then conduct the experiment over a series of $\lambda$ ranging from $0.03$ to $0.0001$ with three cross-validations and use the result's STD as the error bars. 

\subsubsection{Dimension Reduction}

\begin{figure}[bt!]
     \centering
     \begin{subfigure}[b]{0.4\textwidth}
         \centering
         \includegraphics[width=\textwidth]{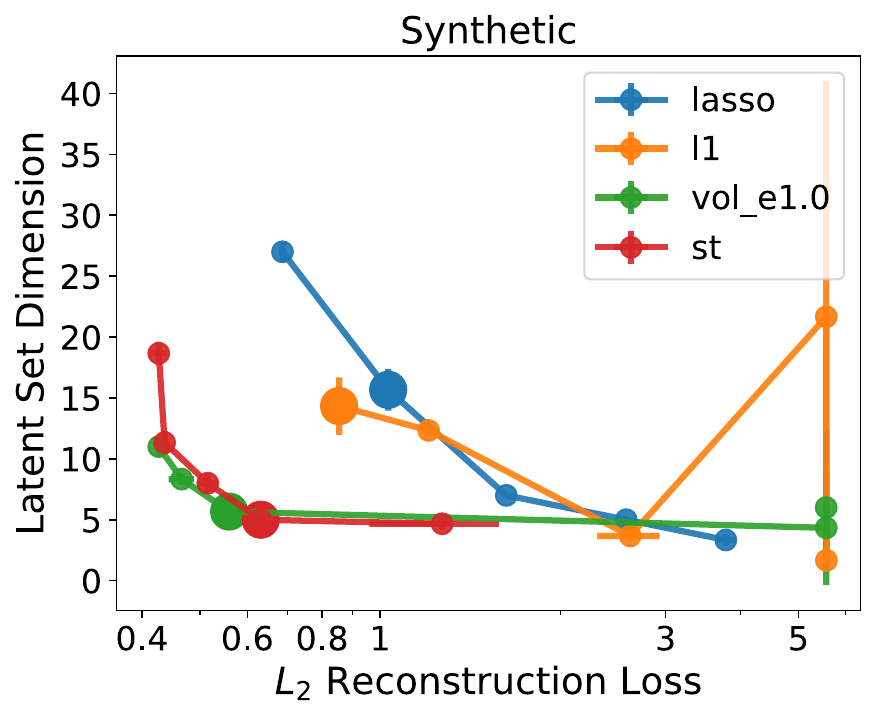}
         \caption{Synthetic}
         \label{fig3:syn_r}
     \end{subfigure}
     \hspace{2cm}
     \begin{subfigure}[b]{0.4\textwidth}
         \centering
         \includegraphics[width=\textwidth]{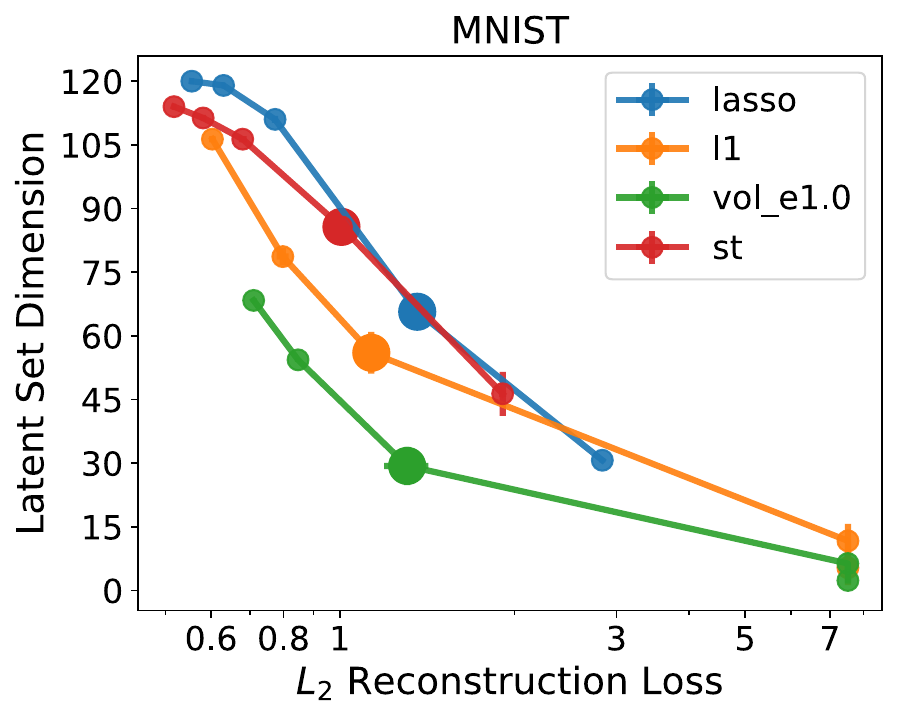}
         \caption{MNIST}
         \label{fig3:mni_r}
     \end{subfigure}
     \begin{subfigure}[b]{0.4\textwidth}
         \centering
         \includegraphics[width=\textwidth]{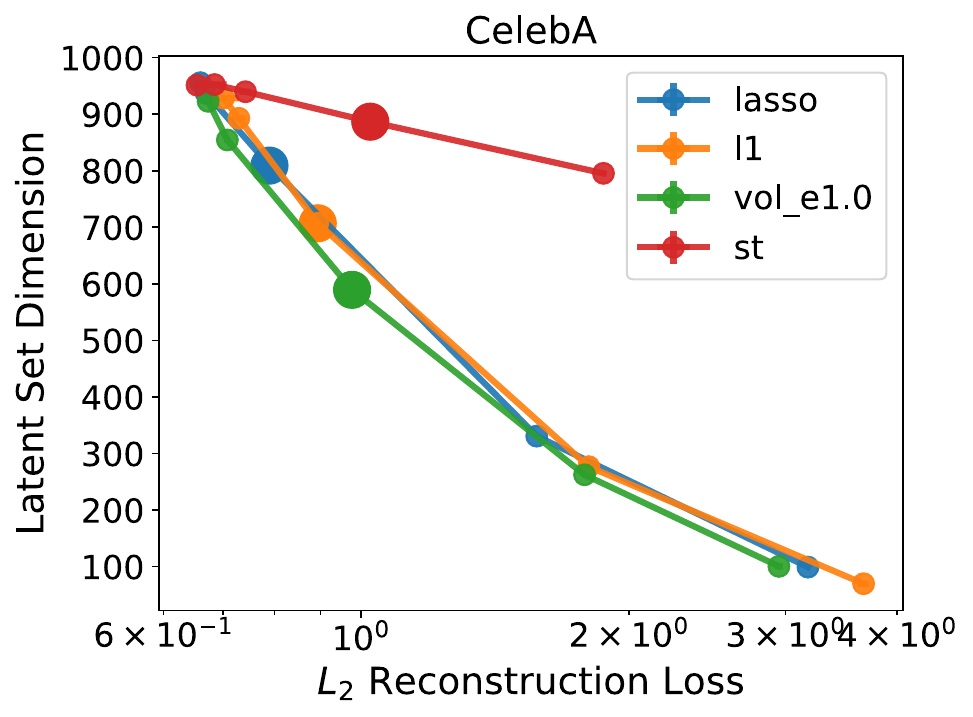}
         \caption{CelebA}
         \label{fig3:cel_r}
     \end{subfigure}
     \hspace{2cm}
     \begin{subfigure}[b]{0.4\textwidth}
         \centering
         \includegraphics[width=\textwidth]{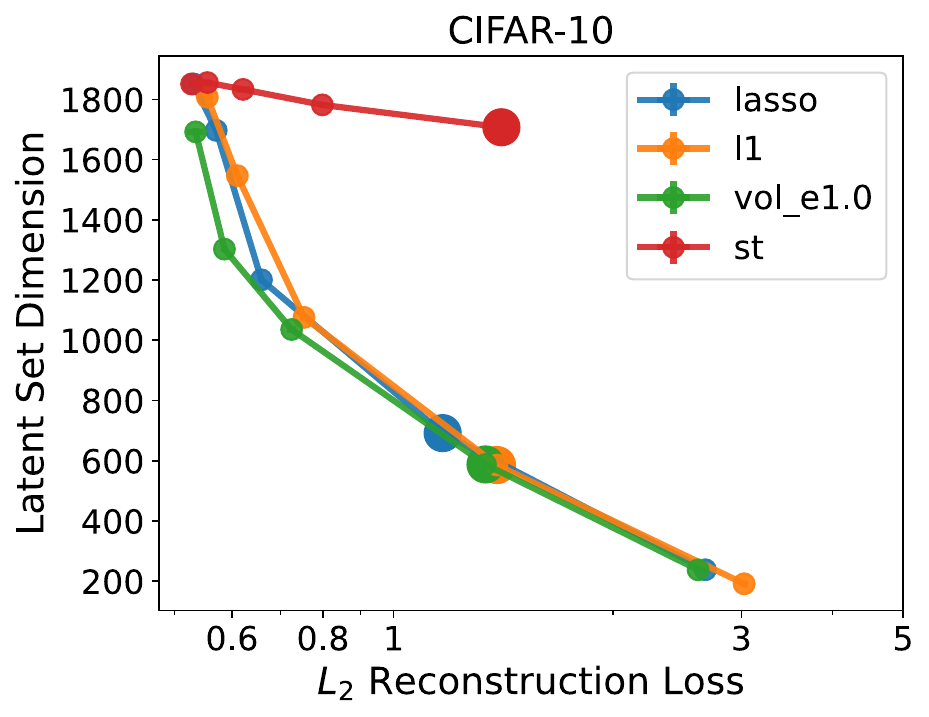}
         \caption{CIFAR-10}
         \label{fig3:cif_r}
     \end{subfigure}
        \caption{Latent Set Dimensionality vs $L_2$ Reconstruction Error. In general, the higher the $\lambda$, the lower the latent set dimension, but the reconstruction error also increases accordingly.}
        \label{fig3:rank}
\end{figure}

Comparing these regularizers' dimension reduction performance is not a straightforward task. This is because each latent regularizer $R$, when paired with the reconstruction loss $J$ via $L = J + \lambda\cdot R$, creates its own loss landscape, such that the same $\lambda$ leads to different $J$ for different $R$ after optimizing $L$. This means that under the same $\lambda$, some regularizers may choose to compress the latent set simply by sacrificing their reconstruction quality more, so we cannot compare these regularizers under the same $\lambda$. However, because adjusting $\lambda$ is essentially trading off between reconstruction and dimension reduction, we can plot \emph{the dimension reduction metric against the reconstruction metric} for comparison. An efficient $R$ should compress the latent set more for the same reconstruction quality.
Figure~\ref{fig3:rank} plots the latent set dimension against the autoencoder's $L_2$ reconstruction error, where the latent set dimension is the number of dimensions left after cumulatively pruning the latent dimensions with the smallest STDs until their joint explained reconstruction exceeds $1\%$. We can see that for all datasets, the na\"ive volume penalty $L^\eta_\text{vol}$ always achieves the highest compression when the $L_2$ reconstruction error is reasonable (empirically when the $L_2$ error is $< 2$, see~\citep{qiuyi2024compressing}). 

\subsubsection{Importance Ordering}

\begin{figure}[bt!]
     \centering
     \begin{subfigure}[b]{0.4\textwidth}
         \centering
         \includegraphics[width=\textwidth]{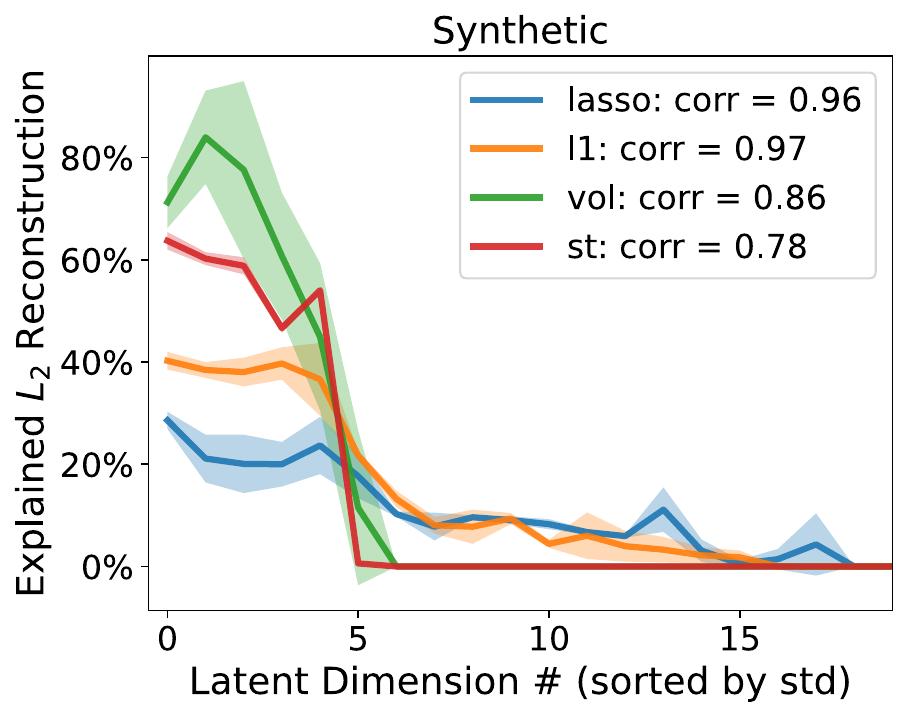}
         \caption{Synthetic}
         \label{fig3:syn_i}
     \end{subfigure}
     \hspace{2cm}
     \begin{subfigure}[b]{0.4\textwidth}
         \centering
         \includegraphics[width=\textwidth]{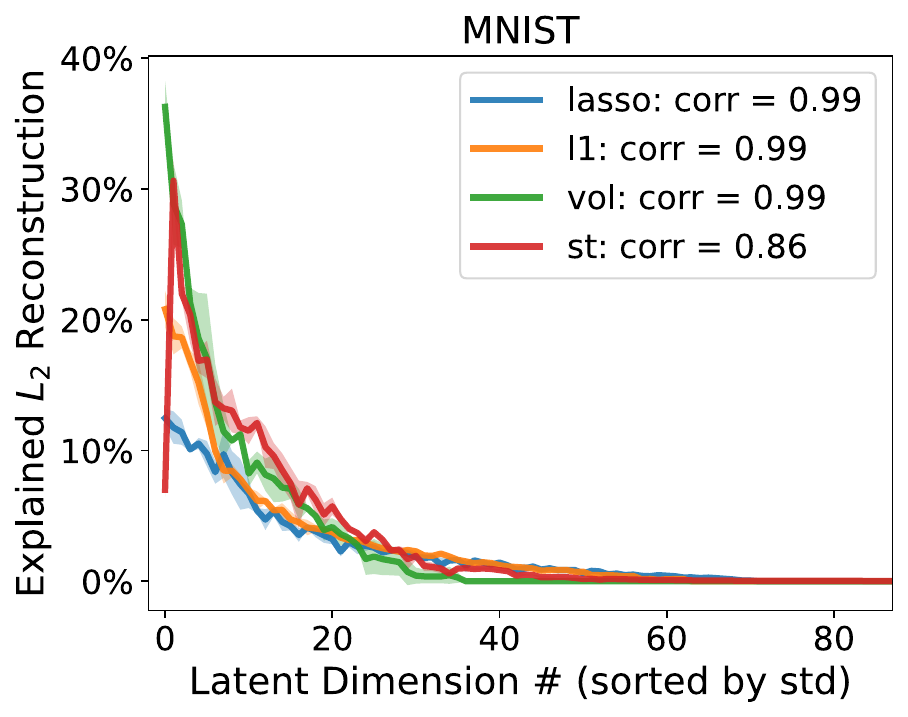}
         \caption{MNIST}
         \label{fig3:mni_i}
     \end{subfigure}
     \hfill
     \begin{subfigure}[b]{0.4\textwidth}
         \centering
         \includegraphics[width=\textwidth]{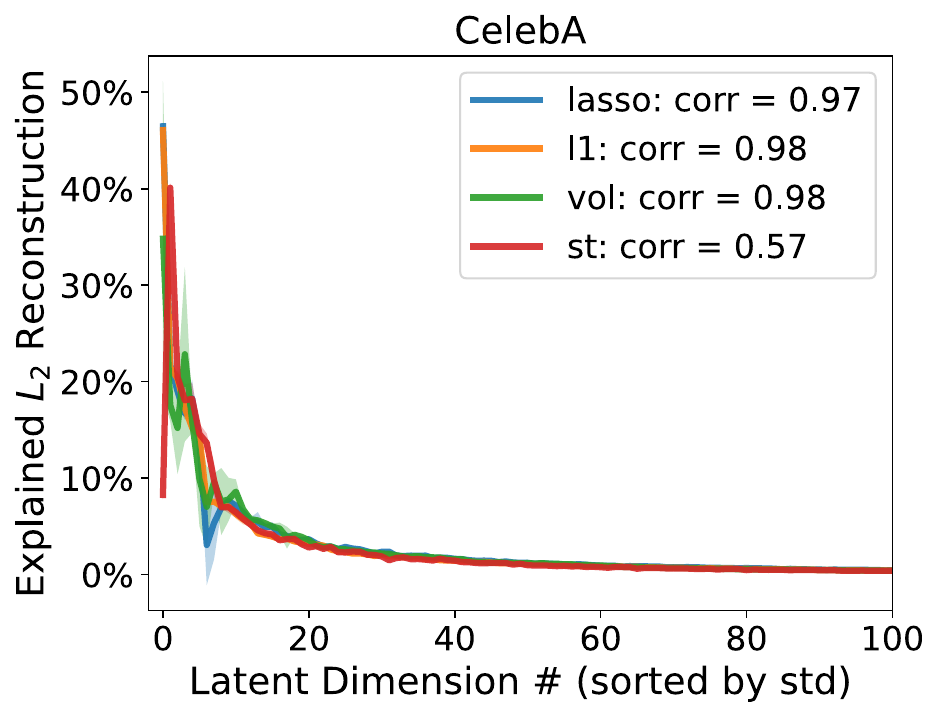}
         \caption{CelebA}
         \label{fig3:cel_i}
     \end{subfigure}
     \hspace{2cm}
     \begin{subfigure}[b]{0.4\textwidth}
         \centering
         \includegraphics[width=\textwidth]{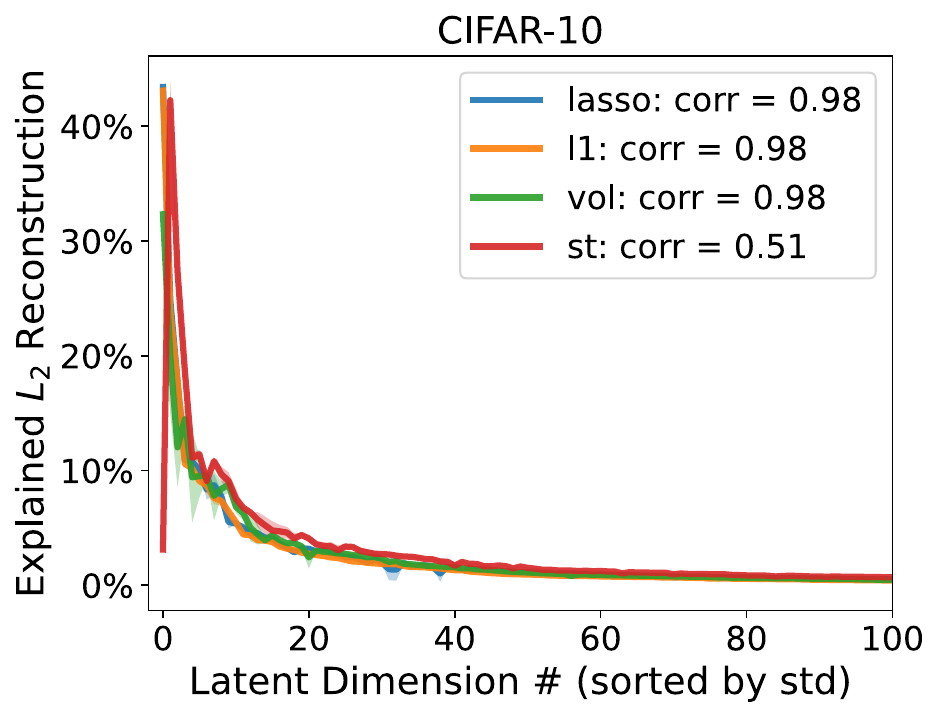}
         \caption{CIFAR-10}
         \label{fig3:cif_i}
     \end{subfigure}
        \caption{Explained Reconstruction vs Latent Dimension \# (sorted in descending order of STD)}
        \label{fig3:exr}
\end{figure}

For each dataset, we select three autoencoding cases with comparable reconstruction quality respectively for the four regularizers (marked by large dots in Fig.~\ref{fig3:rank}). Then in Fig.~\ref{fig3:exr}, for each case we sort the latent dimensions in descending order of STD value, and plot their individual explained reconstructions $\mathcal{R}_D(P)$ against their ordered indices. We choose $L_2$ distance as $D$, \ie, $L_2(\tilde{x}_P, x) = \|\tilde{x}_P - x\|_2$, and measure the Pearson correlation coefficient (PCC) between the latent STD and $\mathcal{R}_{L_2}(\{i\})$ to see if there is any similar ordering effect. If this coefficient is close to $1$, then the latent STD can empirically serve as an indicator of the importance of each latent dimension, just like in PCA.

For all regularizers, we can see that the explained $L_2$ reconstruction generally increases with the latent STD, although it is not perfectly monotonic. Nevertheless, we can conclude that the explained reconstruction is highly correlated with latent STD, since the PCCs are all close to 1 (except for ``st'' based on student's t-distribution, which has a sudden drop as the dimension index reach 0), as shown in Fig.~\ref{fig3:exr}.

\section{Experiments: Unsupervised Least Volume Analysis}
\label{sec:exp_ulva}
In this first primary experiment section, we demonstrate the DP algorithm's effectiveness on several unlabeled datasets, and compare the results to the original least volume autoencoder without DP to show its improvement. In addition, we investigate the least volume autoencoder's generative process from the perspective of topology, and demythify several common misconceptions regarding autoencoders (\eg, disentangled representation). Finally, by leveraging the LVAE-DP's ability to retrieve the least dimensional latent space that can still topologically embed the dataset, we analyze different datasets' topological complexities with LVAE-DP.

\subsection{Least Dimensional Embeddings}
\label{sec:ldembed}
\begin{figure}[bt!]
    \centering
    \begin{subfigure}[b]{\textwidth}
         \centering
         \includegraphics[width=\textwidth]{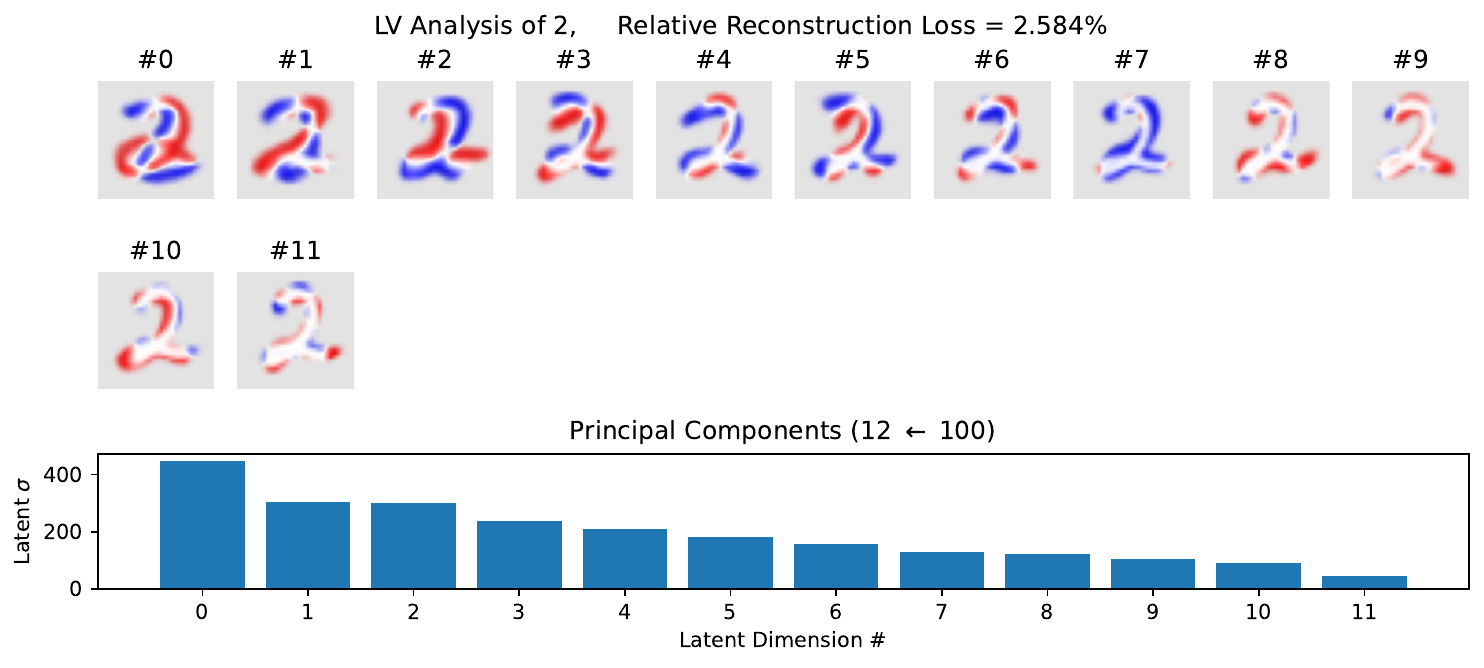}
         \caption{LV Analysis of MNIST Dataset - Digit 2}
         \label{fig4:latent_2}
     \end{subfigure}
     \begin{subfigure}[b]{\textwidth}
         \centering
         \includegraphics[width=\textwidth]{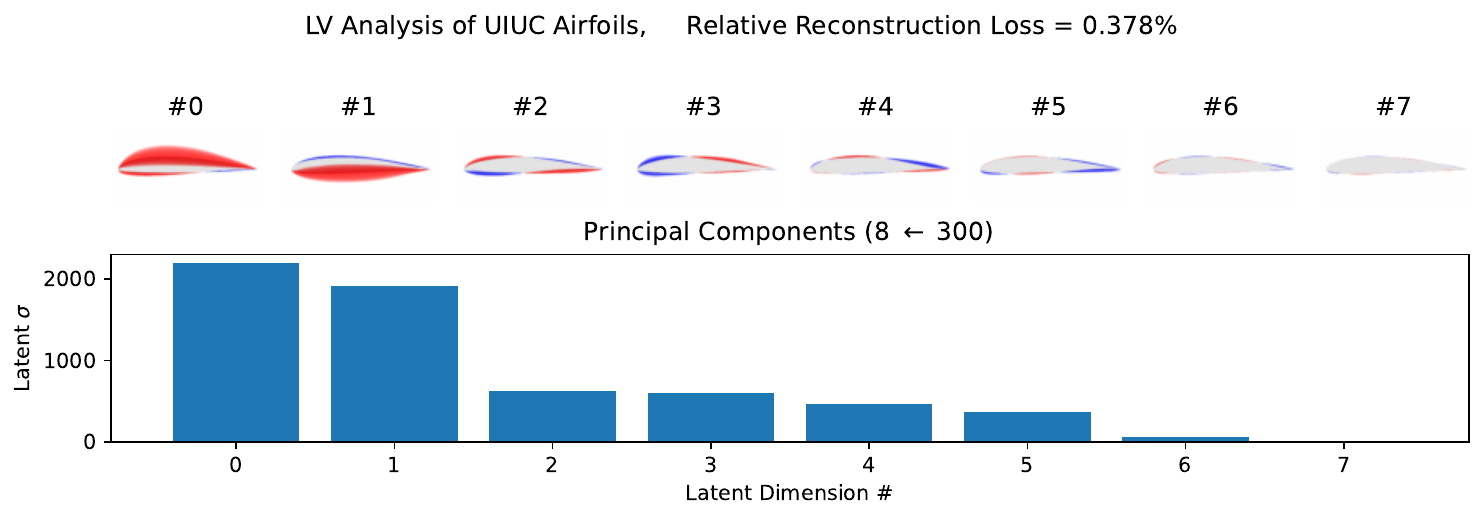}
         \caption{LV Analysis of UIUC Airfoil Dataset}
         \label{fig4:latent_uiuc}
     \end{subfigure}
    \caption{Least Volume analysis of MNIST-2 and UIUC airfoil dataset. Here the relative reconstruction error is defined by $\|\hat{x} - x\|_1 / (\ell \cdot \dim X)$, where $\ell$ is the length scale of pixel values (in our case $\ell = 1$ because the images are normalized) and $\dim X$ equals the number of pixels. The blue and red colors in Fig.~\ref{fig4:latent_2} and~\ref{fig4:latent_uiuc} represent the Pearson correlations between the pixels and the corresponding latent dimensions, giving us an understanding of which parts of the grayscale images the latent dimensions control to what degree.} 
    \label{fig4:lva}
\end{figure}

We first apply the new LVAE-DP on a variety of datasets to demonstrate its power and effect. Figure~\ref{fig4:lva} illustrates its results on the MNIST-2~\citep{deng2012mnist} dataset and the UIUC airfoil dataset~\citep{selig1996uiuc}. We can see that the LVAE-DP not only significantly reduces the dimensions of the latent spaces (100 to 12 for MNIST-2 and 300 to 8 for UIUC airfoils), but also possesses a PCA-like effect on the latent space that makes the importance of the latent space scale with its standard deviation, as discussed in \S\ref{sec:pca_relation}. 
The data variations that these latent dimensions induce are shown later in Fig.~\ref{fig4:lvint}.
Therefore, the LVAE-DP may be regarded as a nonlinear generalization of the traditional linear PCA.  We summarize the embedding dimensions retrieved by LVAE-DP in Table~\ref{tab:embedding_dim}.

\begin{table}[hbt!]
    \centering
    \caption{Embedding Dimensions of MNIST and UIUC Airfoils Retrieved by LVAE-DP}
    \label{tab:embedding_dim}
    \begin{tabular}{c|cccccccccccc}
    \toprule
         Dataset &  0&  1&  2&  3&  4&  5&  6&  7& 8 &9 & Airfoil\\ \midrule
         Dimension &  11&  8&  12&  13&  12&  12&  11&  10&  12& 11& $7\sim8$\\
    \bottomrule
    \end{tabular}
\end{table}

\begin{figure}[bt!]
    \centering
    \begin{subfigure}[b]{\textwidth}
         \centering
         \includegraphics[width=\textwidth]{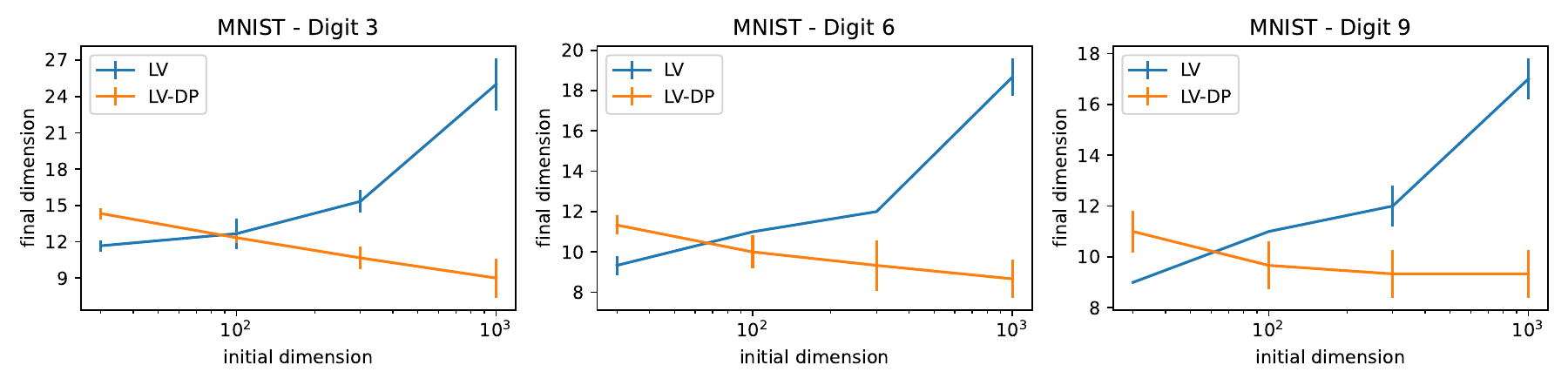}
         \caption{Final Latent Dimension vs Initial Latent Dimension}
         \label{fig4:lvvsdp_dim}
     \end{subfigure}
     \begin{subfigure}[b]{\textwidth}
         \centering
         \includegraphics[width=\textwidth]{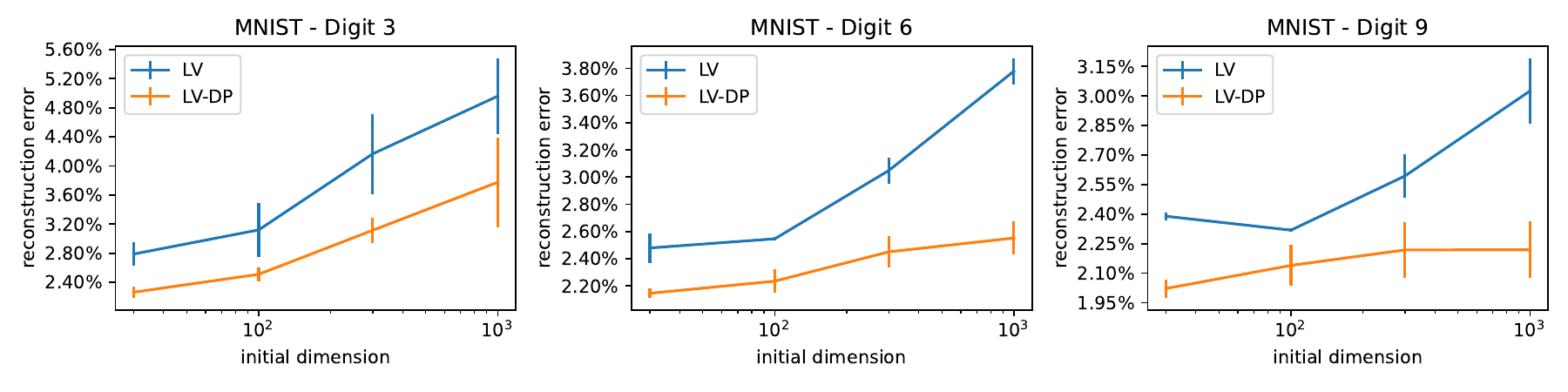}
         \caption{Relative Reconstruction Loss vs Initial Latent Dimension}
         \label{fig4:lvvsdp_rec}
     \end{subfigure}
    \caption{Dimension reduction results of LVAE with and without DP on MNIST datasets of digit 3, 6 and 9. All models are trained for the same number of epochs. The error bars indicate STD.} 
    \label{fig4:exp_lv_vs_dp}
\end{figure}

To highlight the importance and necessity of the DP algorithm, we study on a few datasets how the dimension reduction results of the new LVAE-DP and the old LVAE (based on $L'_\text{vol}$) change \wrt different initial latent space dimensions $\dim Z$. Figure~\ref{fig4:exp_lv_vs_dp} shows that the LVAE without DP tends to have worse dimension reduction performance as $\dim Z$ increases, given that not only its latent sets' final dimensions $\|\gZ\|_0$ but also its reconstruction errors increase with $\dim Z$. In contrast, the new LVAE with DP can consistently yield comparable $\|\gZ\|_0$ and reconstruction errors for different $\dim Z$. Apart from that, it also compresses the latent sets into latent subspaces of lower dimensions while achieving lower reconstruction loss. Therefore, the DP algorithm can significantly improve the efficiency of LVAEs. 

\subsection{Importance of Low Dimensional Latent Spaces}
\label{sec:ldspace}
Retrieving the least dimensional latent space not only can alleviate the curse of dimensionality for downstream tasks~\citep{bengio2013representation,van2009dimensionality,rombach2022high}, but also may improve the robustness of data sampling in the latent space. To demonstrate this effect, let us first recall the consensus that an ordinary autoencoder without any regularization cannot serve as a good data generator, based on the experience that its decoder usually generates invalid data samples when we feed it some latent codes randomly drawn from a simple latent distribution $\sP_z$ (\eg, uniform or Gaussian distribution). Regularized autoencoders such as VAEs~\citep{KingmaW13} were introduced in the past to overcome this deficiency. 
However, such deficiency should not be blamed totally on the (ordinary) autoencoder's simplistic formulation: If an autoencoder $(e, g)_{X\leftrightarrow Z}$ is trained to make $\E_{x\sim\gX}\|g\circ e(x) - x\| = 0$, then drawing the latent codes $z$ only from the latent set $\gZ \coloneqq e(\gX)\subseteq Z$ must allow us to produce realistic $x = g(z)$ in the vicinity of the dataset $\gX$, so every autoencoder's $g$ can potentially become a valid data generator\textemdash as long as we know how to draw $z$ from the latent set $\gZ = e(\gX)$ when we \emph{have no access to $e$ and $\gX$}.

\subsubsection{Implication of Dimension Mismatch}
But what makes sampling $\gZ$ tricky in general? We argue that this is primarily due to the \emph{mismatch between the intrinsic dimensionality of $\gX$ and the latent space $Z$'s dimension}. 

The detailed reasoning is as follows. Suppose after training an autoencoder $(e, g)$ for a dataset $\gX$ satisfying Assumption~\ref{asp:smooth_manifold}, we have $g\circ e(x) = x$ for all $x\in \gX$, such that $\gZ \simeq \gX$ and $\restr{e}{\gX}$ is a topological embedding. 
Then for any $\gX^i\subseteq \gX$ with ${\dim \gX^i < \dim Z}$, its latent set $e(\gX^i)\subseteq \gZ$ must have measure zero in $Z$, thanks to the following lemma based on \emph{Sard's theorem}:
\begin{lemma}[Sard's Theorem for Neural Networks]
\label{lemma4:manifold_measure}
    Suppose $\gX$ is a submanifold of $X$ and $e: X\to Z$ is a continuous neural network whose activations $h:\sR\to\sR$ are smooth with at most a countable number of non-differentiable points, while $e$ has a common neural network architecture that makes itself smooth if all its activation functions are smooth. 
    If $\dim \gX < \dim Z$, then $e(\gX)$ has measure zero in $Z$.  
\end{lemma}
\begin{proof}
    Assume that some of $e$'s activation functions $h^i:\sR\to\sR$ are only piecewisely smooth with a countable number of non-differentiable points. 
    For each $h^i$, its non-differentiable point(s) partition its domain into a countable number of intervals $I_j\subset \sR$, and on the closure of each $I_j$ a sub-function $h^i_j: \closure{I_j}\to \sR$ can be defined to match $h^i$ on $\closure{I_j}$. Then by the \emph{Whitney extension theorem}~\citep{whitney1992analytic}, each sub-function $h^i_j$ has a smooth extension ${\widetilde{h^i_j}: \sR \to \sR}$. If we replace every $h^i$ in $e$ with a certain $\widetilde{h^i_j}$, then by construction these possible combinations yield a \emph{countable} number of different smooth $e^k: X\to Z$. 
    \emph{Sard's theorem}~\cite[Corollary 6.11]{lee2012smooth} then shows that each $e^k(\gX)$ has measure zero in $Z$ when $\dim \gX < \dim Z$. It follows that ${e(\gX) \subseteq \bigcup_k e^k(\gX)}$ has measure zero in $Z$. 
\end{proof}

Therefore, if the latent probability measure $\sP_z$ is \emph{equivalent} to the Lebesgue measure\textemdash \ie, they agree on which sets have measure zero, \eg, Gaussian measure\textemdash while $\supp\sP_z \cap e(\gX^i) \neq \emptyset$, then $\sP_z(\supp\sP_z \cap e(\gX^i)) = 0$, which means it is impossible to sample any subset of $e(\gX^i)$ via $\sP_z$. 
Hence, for $\gX = \bigcup_i \gX^i$ that satisfies $\dim\gX < \dim Z$, it is impossible to sample $\gZ$ using such $\sP_z$, given that $\sP_z(\gZ) = \sP_z(\bigcup_i e(\gX^i)) \leq \sum_i \sP_z(e(\gX^i)) = 0$. In consequence, $z$ sampled from $\sP_z$ almost always fall outside $\gZ$, thus are very likely mapped to invalid $x = g(z)\not\in \gX$, especially when the decoder $g$ is injective over $\supp\sP_z$.

To enable sampling $z$ from $\gZ$ using such $\sP_z$, we need to make ${\sP_z(\supp\sP_z \cap \gZ) > 0}$. It can be achieved if $\supp\sP_z \cap \gZ$ has non-empty interior, such that $\supp\sP_z \cap \gZ$ has positive measure in $Z$. This becomes possible when \emph{$\dim Z=\dim \gX$}. Specifically, if a manifold $\gX^i\subseteq \gX$ satisfies $\dim \gX^i = \dim Z$, then due to the \emph{invariance of domain}~\citep[Corollary~19.9]{bredon2013topology}, for any open set $U \subseteq \gX^i \backslash \partial\gX^i$, its homeomorphic latent set $e(U)\subseteq \gZ$ must be \emph{open} and thus of positive measure in $Z$, because $\restr{e}{\gX}$ is injective and continuous. It follows that ${\supp\sP_z \cap e(U) \neq \emptyset} \Rightarrow {\interior{\supp\sP_z} \cap e(U) \neq \emptyset} \Rightarrow {\sP_z(\supp\sP_z \cap e(U)) > 0} \Rightarrow {\sP_z(\supp\sP_z \cap \gX^i) > 0}$. 
Therefore, if $\dim \gX^i = \dim Z$ for almost all $\gX^i\subseteq\gX$, we can sample $\gZ$ sufficiently with a latent distribution $\sP_z$ if $\supp\sP_z$ intersects most parts of $\gZ$, such that ${\sP_z(\supp\sP_z \cap \gZ) > 0}$. It becomes more ideal if $\sP_z(\supp\sP_z\cap\gZ) \gg \sP_z(\supp\sP_z \backslash \gZ)$, which makes most $z$ sampled from $\sP_z$ fall inside $\gZ$.

In brief, reducing the latent space $Z$'s dimension to the dataset $\gX$'s intrinsic dimension\textemdash ass\-uming that it matches $\gX$'s embedding dimension\textemdash can force the latent set $\gZ\simeq \gX$ to occupy a positive amount of space in $Z$, such that (at least some part of) it becomes easier to sample. 

\subsubsection{Experiment Results}
\begin{figure}[bt!]
    \centering
    \begin{subfigure}[b]{0.33\textwidth}
         \centering
         \includegraphics[width=\textwidth, trim={0 0 0 1cm},clip]{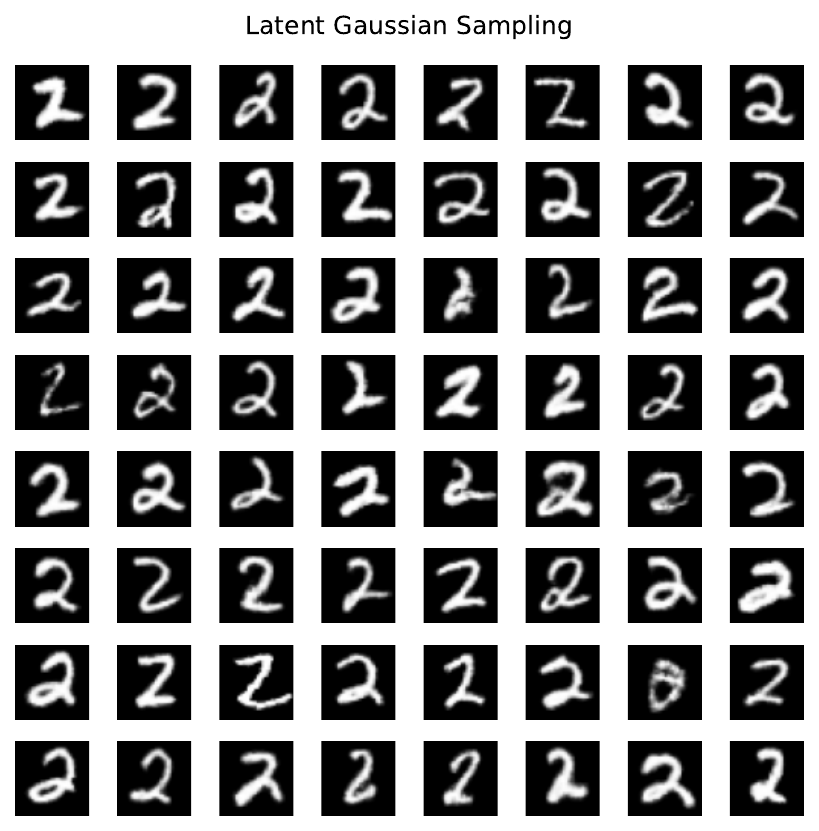}
         \caption{MNIST - Digit 2}
         \label{fig4:noreg_sample_2}
     \end{subfigure}
     \begin{subfigure}[b]{0.63\textwidth}
         \centering
         \includegraphics[width=\textwidth]{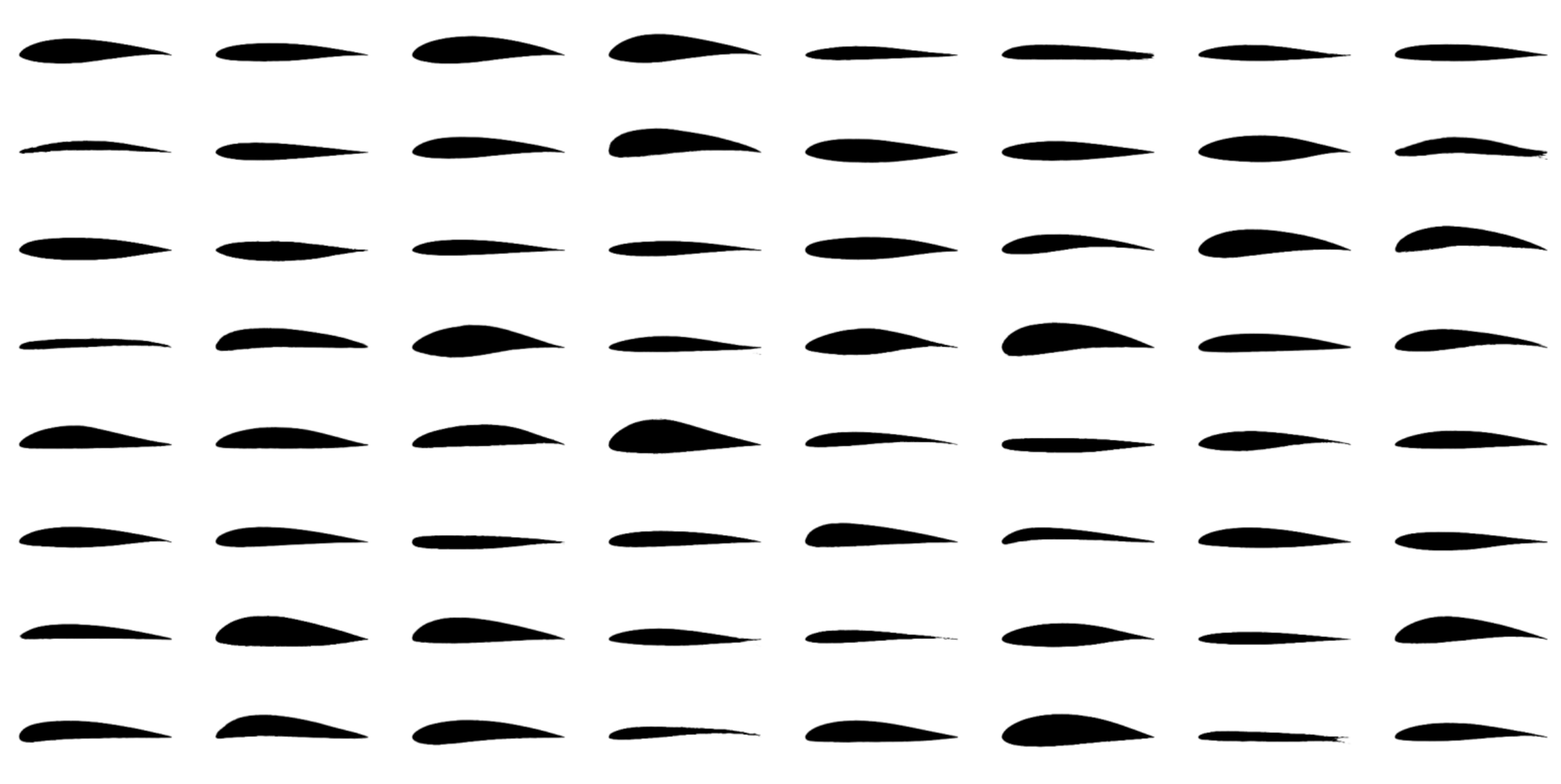}
         \caption{UIUC Airfoils}
         \label{fig4:noreg_sample_airfoil}
     \end{subfigure}
    \caption{Latent Gaussian samples produced by unregularized AEs with $\dim Z\approx \dim \gX$.} 
    \label{fig4:noregsample}
\end{figure}

To verify this hypothesis, we train a series of ordinary autoencoders without any regularization on the previous datasets in \S\ref{sec:ldembed}. However, we additionally set these autoencoders' latent dimensions to the embedding dimensions in Table~\ref{tab:embedding_dim} that we retrieved using LVAE-DP. After training, in each case we obtain a Gaussian approximation $\sP_z$ of the distribution of $e(x)$ (by evaluating $e(x)$'s mean and covariance), and sample points $z$ from this $\sP_z$ to produce the latent Gaussian samples $g(z)\in X$ through the decoder $g$. 
The realistic samples in Fig.~\ref{fig4:noregsample} shows that even for regular autoencoders without regularization, when the latent space dimensions equal the datasets' embedding dimensions (or intrinsic dimension, since these datasets are probably balls), $g$ can become valid generative models when we use the Gaussian approximations as $\sP_z$.

\begin{figure}[bt!]
    \centering
    \begin{subfigure}[b]{\textwidth}
         \centering
         \includegraphics[width=\textwidth]{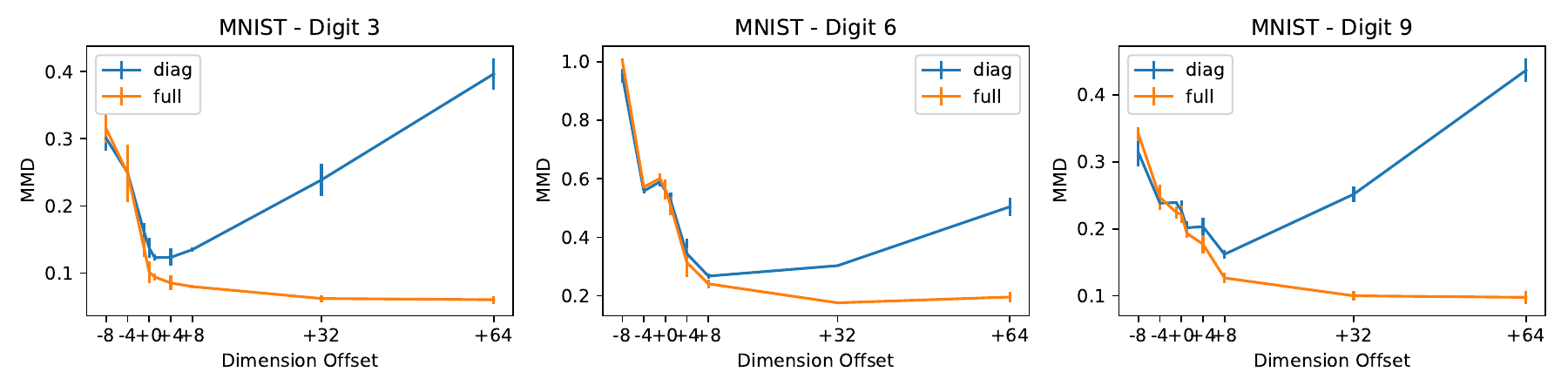}
         \caption{MMD between generated samples and true data samples. Here `full' means the latent Gaussian approximation has full covariance matrix, and `diag' means the Gaussian's covariance is diagonal.}
         \label{fig4:exp_mmd}
     \end{subfigure}
     \begin{subfigure}[b]{0.24\textwidth}
         \centering
         \includegraphics[width=\textwidth]{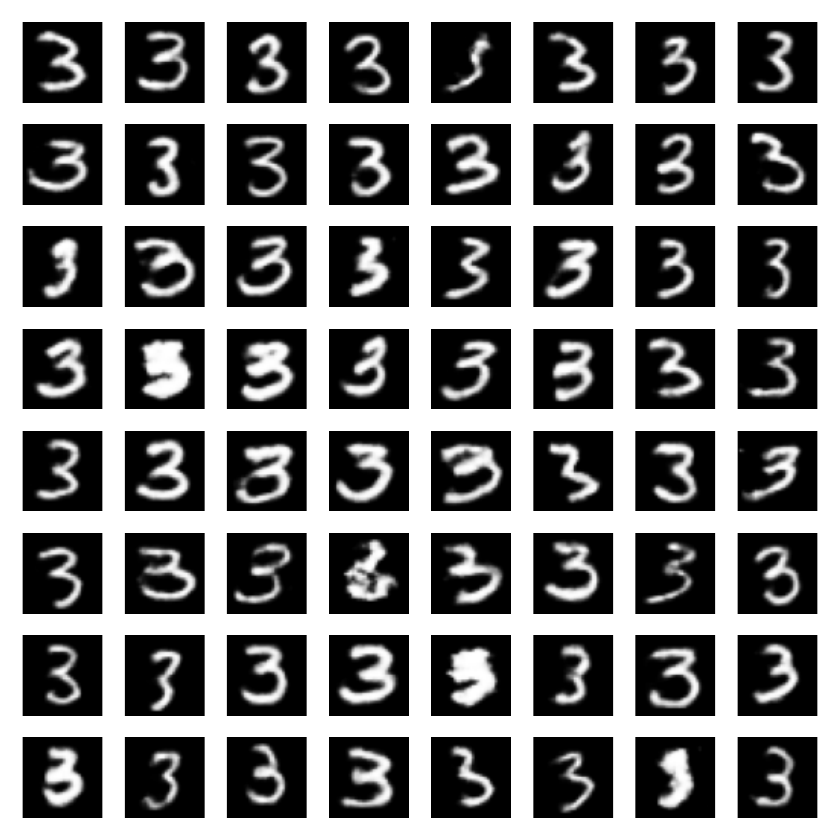}
         \caption{Diag, -8}
     \end{subfigure}
     \begin{subfigure}[b]{0.24\textwidth}
         \centering
         \includegraphics[width=\textwidth]{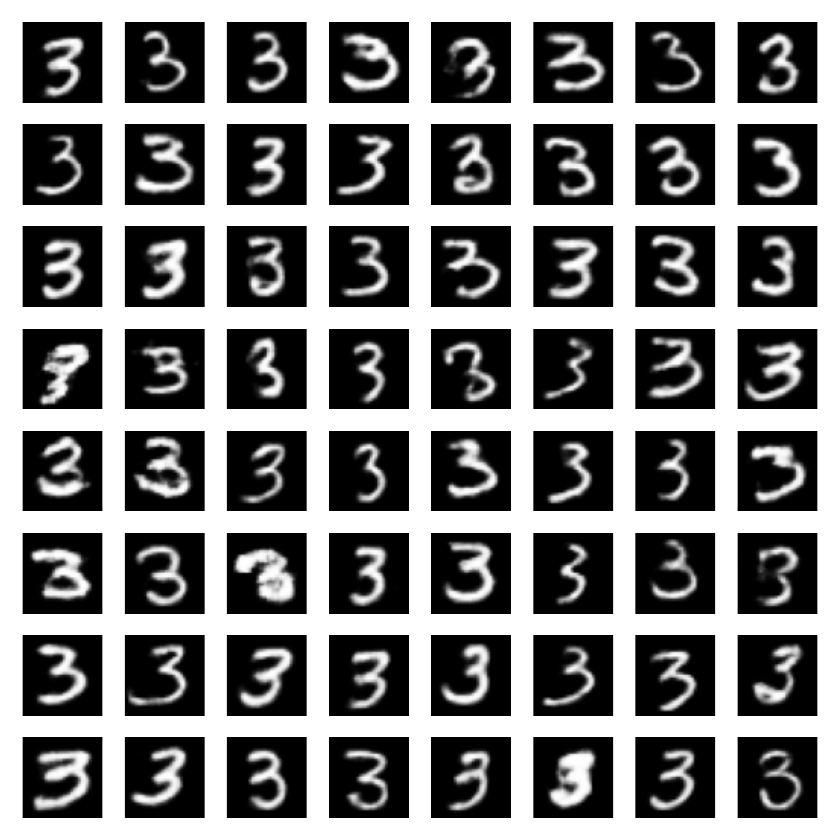}
         \caption{Full, -8}
     \end{subfigure}
     \begin{subfigure}[b]{0.24\textwidth}
         \centering
         \includegraphics[width=\textwidth]{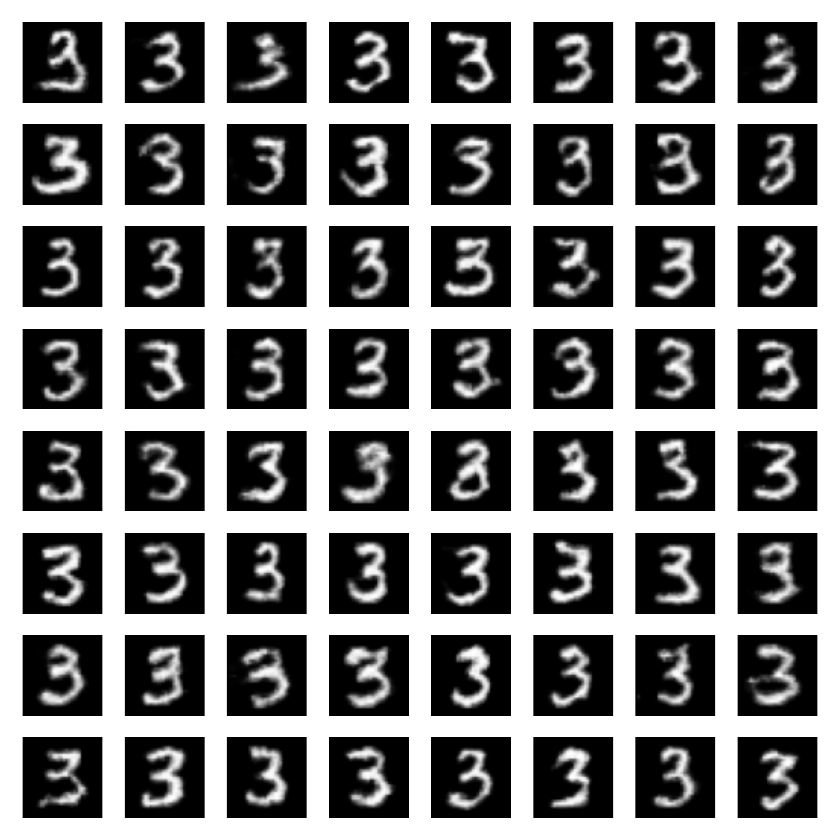}
         \caption{Diag, +64}
     \end{subfigure}
     \begin{subfigure}[b]{0.24\textwidth}
         \centering
         \includegraphics[width=\textwidth]{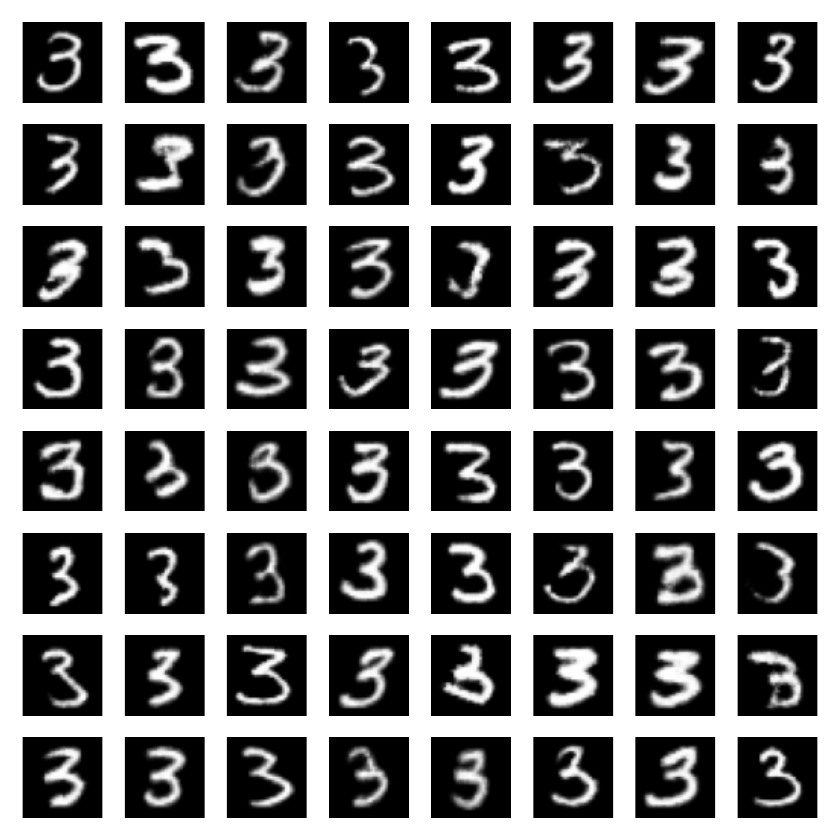}
         \caption{Full, +64}
     \end{subfigure}
    \caption{Relationship between latent Gaussian samples' quality and latent space dimension.} 
    \label{fig4:mmd}
\end{figure}

What happens if the latent space dimension does not equal the embedding dimension? We further investigate how the sampling quality of $g$ changes as we shift the autoencoders' latent dimensions away from the estimated embedding dimensions with a certain offset. We use the \emph{maximum mean discrepancy} (MMD) to evaluate the discrepancy between the distribution of the generated samples $\sP_g \coloneqq g_*\sP_z$ and the data distribution $\mdata$. The result is surprising: In each case, if we only use a Gaussian approximation with diagonal covariance, then $\sP_g$ deviates from $\mdata$ when the latent space dimension moves away from the estimated dimension in either direction. However, if the Gaussian approximation has full covariance matrix, then the discrepancy between $\sP_g$ and $\mdata$ in general only increases when the latent dimension decreases! 

\begin{figure}[bt!]
    \centering
    \begin{subfigure}[b]{0.38\textwidth}
         \centering
         \includegraphics[width=\textwidth]{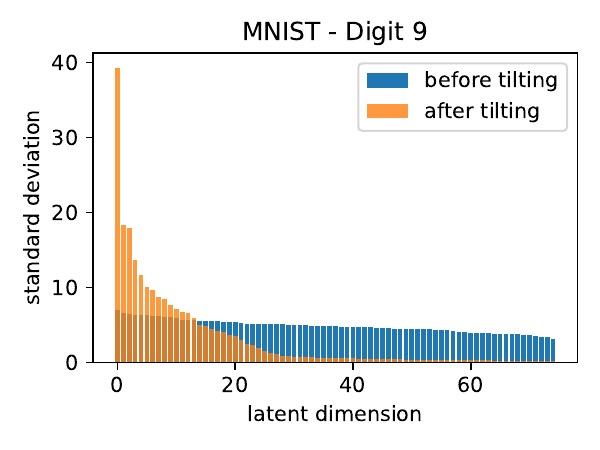}
         \caption{$\gZ$'s STD before and after tilting}
         \label{fig4:zsvd}
     \end{subfigure}
     \begin{subfigure}[b]{0.6\textwidth}
         \centering
         \includegraphics[width=\textwidth]{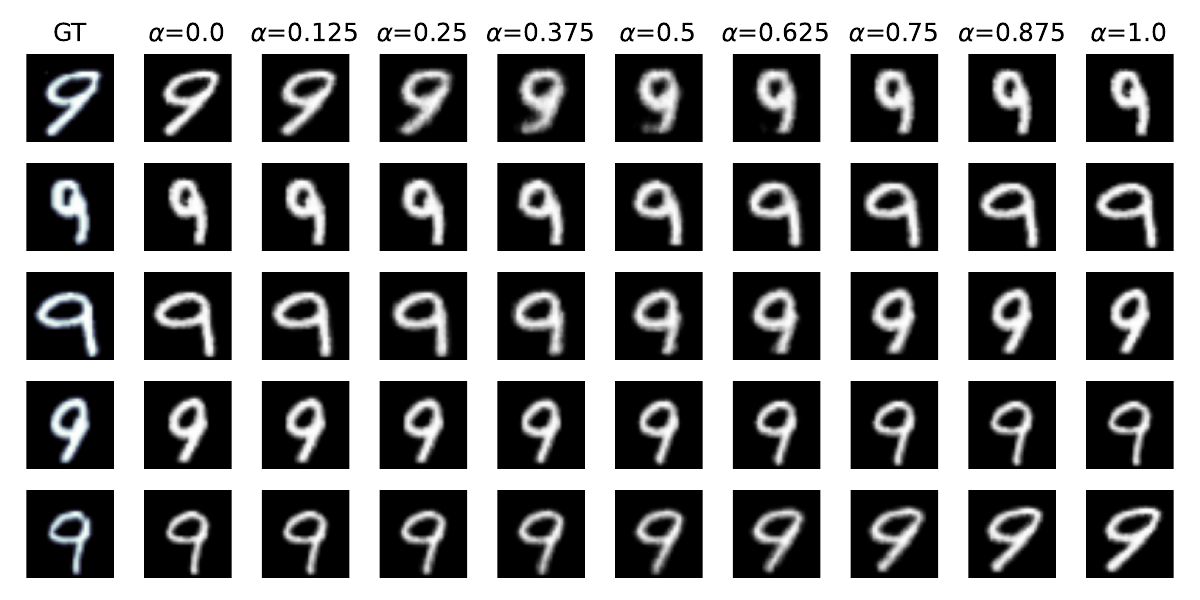}
         \caption{Latent interpolation}
         \label{fig4:zinter}
     \end{subfigure}
    \caption{Inspection of digit 9 latent set when $\dim Z = 64 \gg \dim \gX$. } 
    \label{fig4:zinspection}
\end{figure}

These results has several explanations and implications:
\begin{enumerate}
    \item The datasets we have processed so far are very likely balls (\ie, sets homeomorphic to Euclidean balls), because if this holds, then for each dataset, its latent set is also a ball when the latent space dimension is no less than the dataset's embedding dimension\textemdash or intrinsic dimension since these two dimensions of ball are equal. The latent set thus has non-empty interior when the latent space dimension equals the dataset's embedding dimension. This can explain why the Gaussian approximations $\sP_z$\textemdash whose supports are practically balls\textemdash can cover the latent sets so well and produce valid data samples $g$.
    
    \item When the latent space dimension is lower than the dataset's intrinsic dimension, the Gaussian approximation's support $\supp\sP_z$ is a manifold of dimension $\dim Z < \dim\gX$. Thus, the decoder's image $g(\supp\sP_z)$ can only live on a union of manifolds of dimension lower than $\dim \gX$, according to~\cite[Lemma 1]{arjovsky2017principled}. This means the samples produced by $\sP_g$ can at best only intersect and cover a tiny subset of $\gX$, and explains why the discrepancy between $\sP_g$ and $\mdata$ increases as the latent space dimension goes below $\dim \gX$.  

    \item When $\dim Z$ goes beyond $\dim \gX$, the diagonal Gaussian approximation $\sP_z$ cannot cover $\gZ$ well, but the one with full covariance can. This suggests that $\gZ$ very likely lives on a \emph{tilted flat plane} in the latent space, such that unlike the Gaussian with full covariance, the diagonal Gaussian's support $\supp\sP_z$ is not well aligned with the tilted $\gZ$, which makes it more likely to sample $z$ far away from $\gZ$. When the latent space dimension lowers to the dataset's embedding dimension, the flat latent set has less freedom to tilt itself, thus the diagonal Gaussian approximation $\sP_z$ works better at sampling $\gZ$.
    
    There are two evidences supporting this flat $\gZ$ conjecture. First, for each dataset, if we tilt the latent set to align its principal dimensions with the latent space dimensions, and evaluate the standard deviations of all the latent dimensions before and after the tilting, then the latent set always has only a few dimensions of large STD (an example is shown in Fig.~\ref{fig4:zsvd}). Second, when we interpolate between data samples in the latent space, the interpolation always yields realistic samples, which suggests the latent set is convex and thus cannot be curved (an example is shown in Fig.~\ref{fig4:zsvd}). 
\end{enumerate}

In summary, a regular autoencoder without any regularization can also become a good data generator when we can sample the latent set effectively. This is particularly achievable when the dataset is homeomorphic to a Euclidean ball. In this case, as long as the latent space dimension is not lower than the dataset's embedding dimension, the latent set is homeomorphic to the dataset and might also be flattened in the latent space by the autoencoder, even if the latent space dimension is much higher than the embedding dimension. 
The latent space flattening observed in this unregularized autoencoder is somewhat unexpected and warrants further investigation, but it may be attributed to the implicit regularization phenomenon discussed in~\citep{jing2020implicit}.
Nevertheless, when the flattening occurs, this flat but tilted latent set can be easily retrieved by PCA, and can be sampled efficiently by its Gaussian approximation to produce valid data samples. 


\subsection{Demystification of Disentangled Representations}
\label{sec:disentangle}
\begin{figure}[bt!]
    \centering
    \begin{subfigure}[b]{0.33\textwidth}
         \centering
         \includegraphics[width=\textwidth]{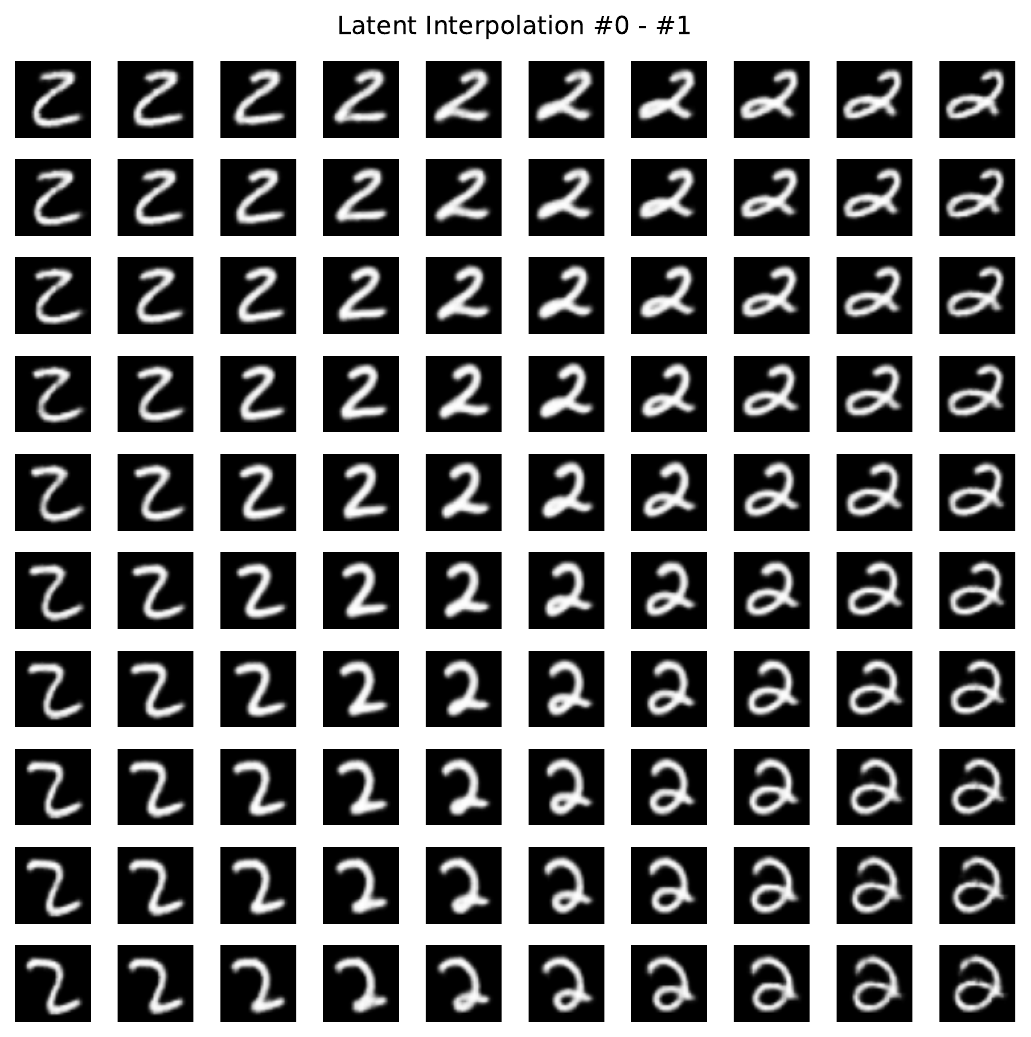}
         \caption{LVAE-DP - Digit 2}
         \label{fig4:lvint_2}
     \end{subfigure}
     \begin{subfigure}[b]{0.66\textwidth}
         \centering
         \includegraphics[width=\textwidth]{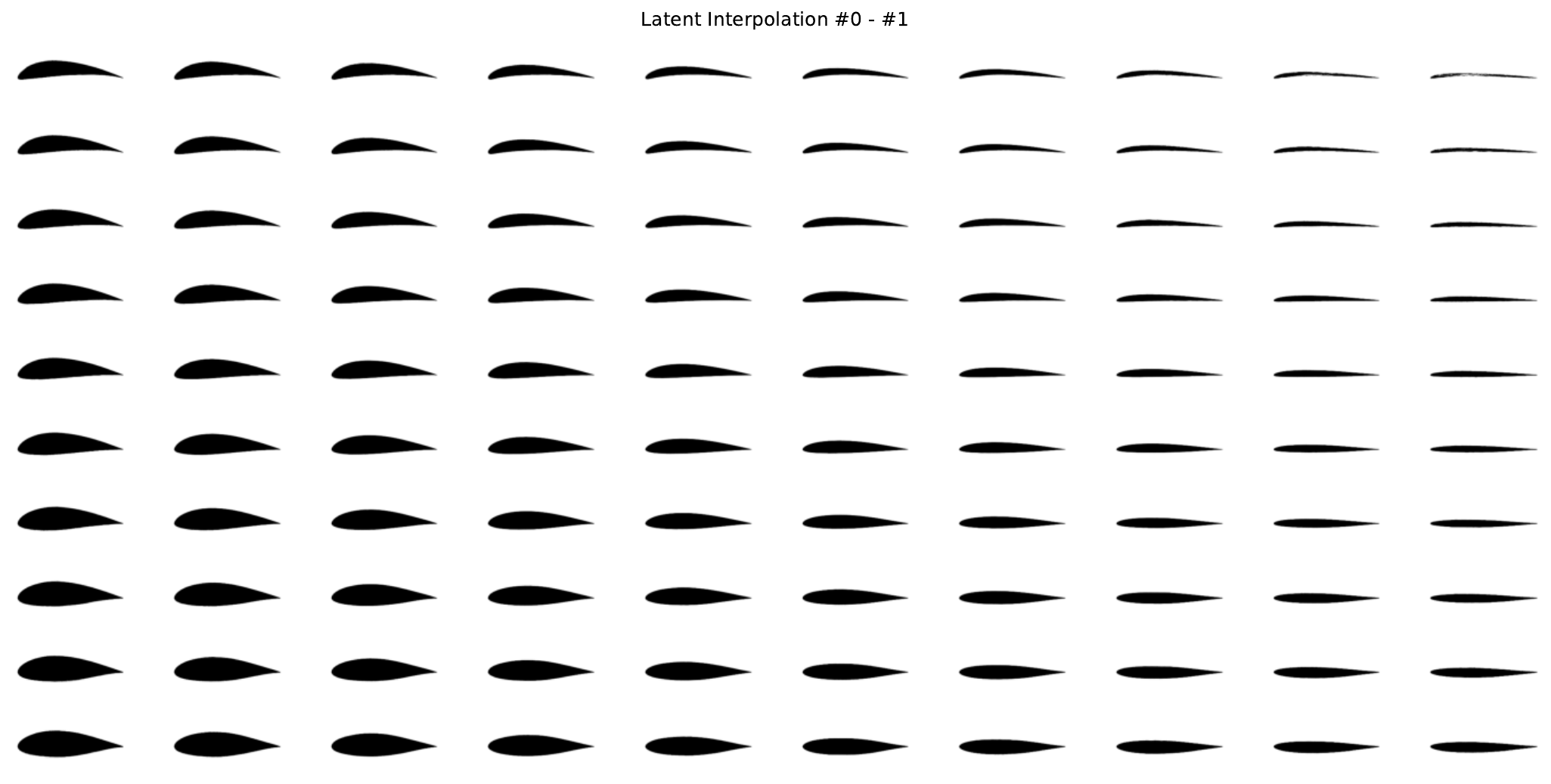}
         \caption{LVAE-DP - Airfoils}
         \label{fig4:lvint_airfoil}
     \end{subfigure}
     \\
     \begin{subfigure}[b]{0.33\textwidth}
         \centering
         \includegraphics[width=\textwidth]{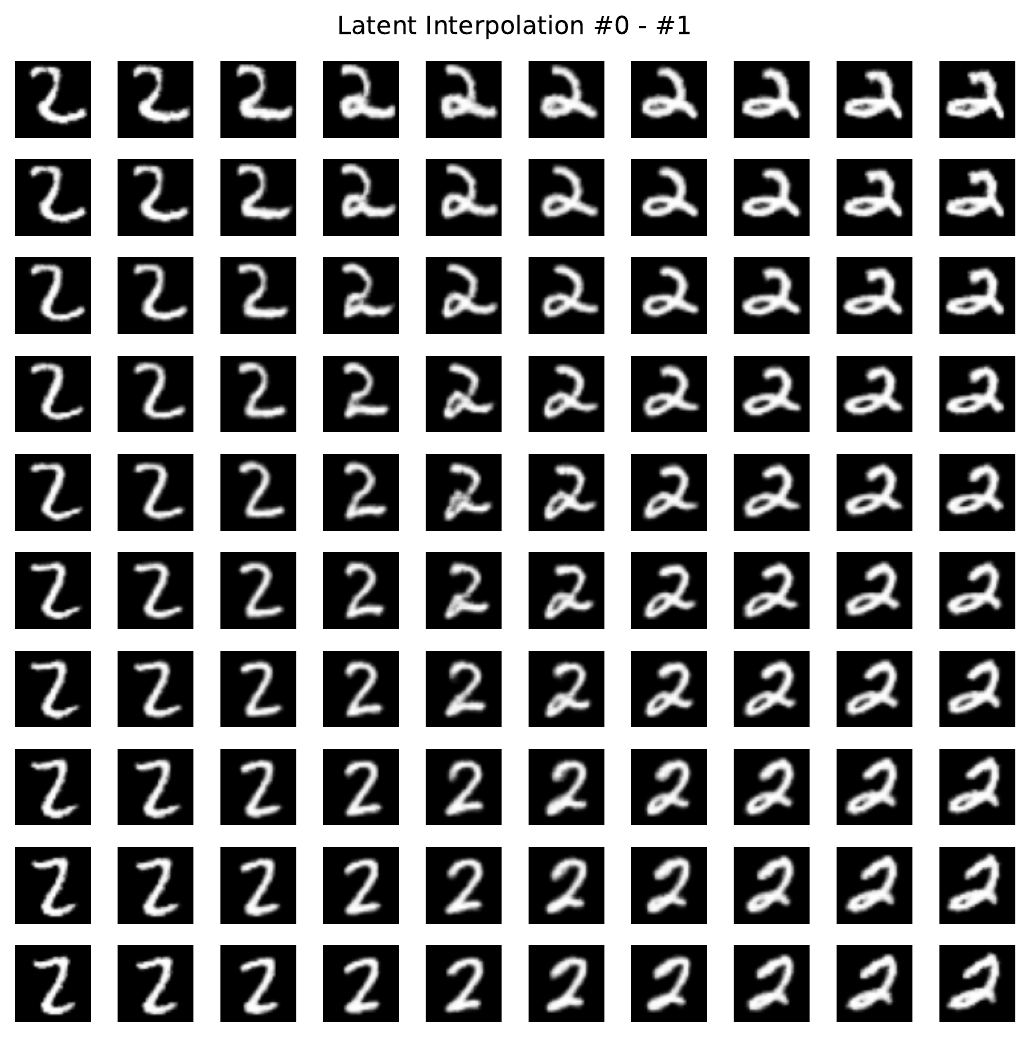}
         \caption{Unregularized - Digit 2}
         \label{fig4:unregint_2}
     \end{subfigure}
     \begin{subfigure}[b]{0.66\textwidth}
         \centering
         \includegraphics[width=\textwidth]{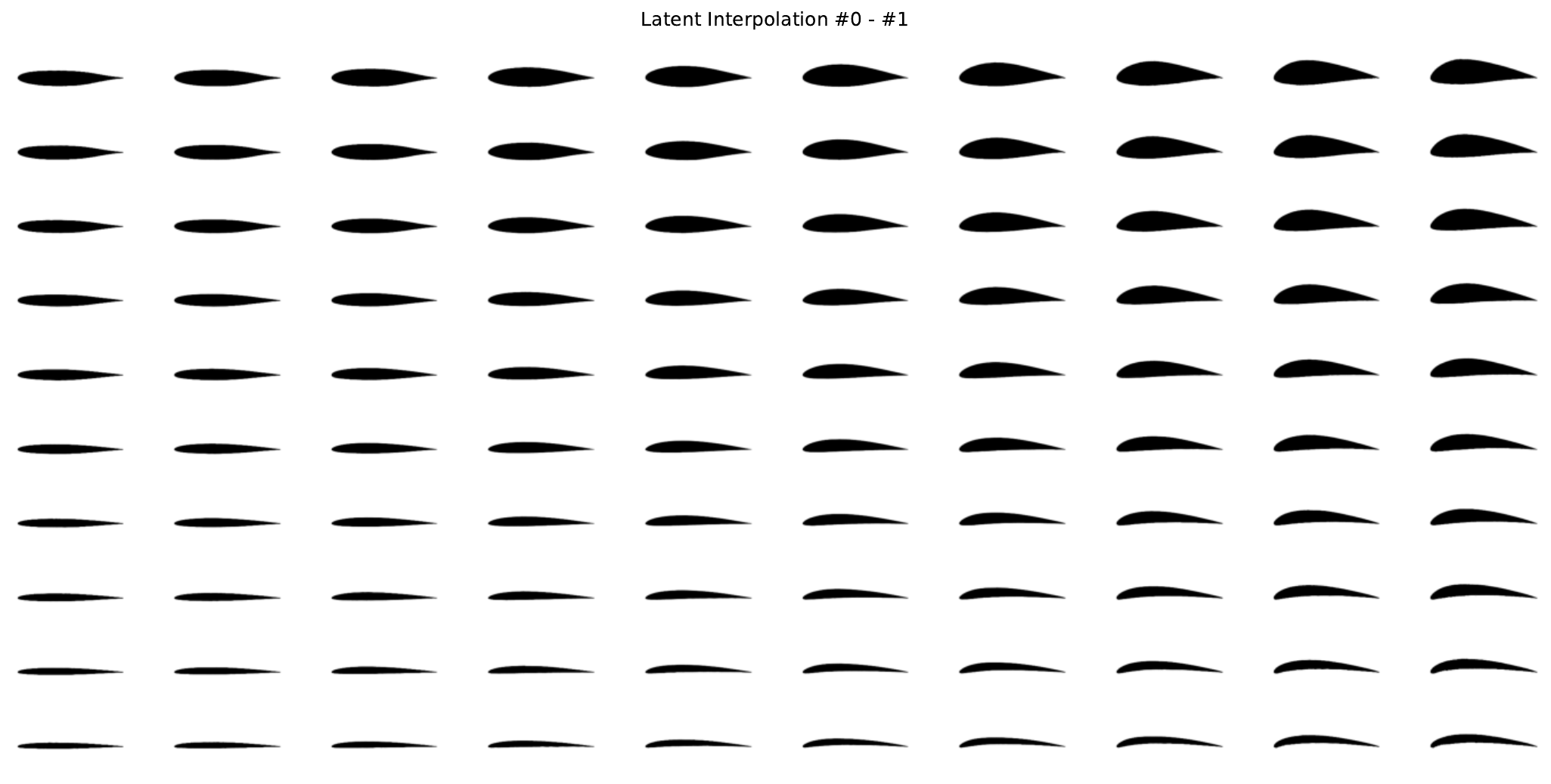}
         \caption{Unregularized - Airfoils}
         \label{fig4:unregint_airfoil}
     \end{subfigure}
    \caption{Latent interpolation of LVAE-DP and unregularized AE in 2-D latent subspaces, revealing disentangled representation. The unregularized AEs' latent dimensions are set to the embedding dimensions retrieved by LVAE-DPs. Here for the unregularized AEs, we perform latent interpolation in the 2-D linear subspaces spanned by the principal components of their latent sets, since unlike LVAE-DP, the unregularized AEs' latent sets do not have very diagonal covariance matrices, \ie, their latent dimensions are correlated.} 
    \label{fig4:lvint}
\end{figure}

Apart from the sampling ability in \S\ref{sec:ldspace}, both the regularized and unregularized autoencoders also appear to have learned \emph{disentangled representations}~\citep{bengio2013representation,schmidhuber1992learning,higgins2017beta,kumar2017variational,kim2019disentangling,van2019disentangled,higgins2018towards,chen2018isolating,eastwood2018framework, ridgeway2018learning, locatello2019challenging}, as we can see in their latent interpolation plots in Fig.~\ref{fig4:lvint}, where each latent dimension ``controls only a single factor of data variation distinctively''\textemdash a description widely used in other literatures for this phenomenon. This contradicts the general belief that disentangled representation can only be achieved through some special types of regularization.

Nevertheless, researchers still have not agreed on the definition of disentanglement. Very frequently, a disentangled representation is required to possess both \emph{modularity} and \emph{compactness}~\citep{eastwood2018framework, ridgeway2018learning}, where modularity implies that each latent dimension controls only one \emph{factor of data variation}, while compactness requires a single factor to be controlled by only one or few latent dimensions. However, this common definition is challenged and dismantled by~\citep{locatello2019challenging}, because it has a flaw rooted in the ambiguous concept of \emph{factor of data variation}. \citet{locatello2019challenging} showed that for any dataset, there are an infinite number of ways to construct the so-called factors of variation, such that it is impossible to tell which combination is the correct one without introducing some inductive biases. 
Indeed, consider a toy image dataset of a 2D ball moving across the 2D plane. It is of course natural to decompose the 2D movement into the horizontal and vertical translations and call them the two factors of variation, but it is not justified why we cannot decompose the 2D movement into the two diagonal translations instead. 
They also showed that the disentanglement scores assuming the existence of some groundtruth factors of variation are all sensitive to random seed, which suggests the occasional good alignment of the latent dimensions with the groundtruth factors might be a fluke. 

\subsubsection{Three Necessary Conditions for Disentangled Representations}

In an attempt to demystify the phenomenon of disentangled representations, we propose the following three \emph{necessary} conditions for the disentanglement effect to emerge. We prefer not to call this a sufficient definition of disentanglement because people, such as \citet{higgins2018towards}, may have additional conditions for their own definitions. 

\begin{proposition}[Disentanglement Conditions]
\label{prop4:disentanglement}
    The disentangled representation achieved by a decoder (or generator) $g: Z\to X$ \wrt a latent set $\gC\subseteq Z$ and an image dataset $\gX \in X$ is a reflection of: 
    \begin{enumerate}
        \item $\restr{g}{\gC}:\gC\to X$ is continuous. 
        \item $\restr{g}{\gC}:\gC\to X$ is injective. ($\star$)
        \item $\sup_{z\in\gC}\inf_{x\in\gX} d_v(g(z), x) < \epsilon$. In other words, $g(\gC)$ is close to $\gX$ everywhere.
    \end{enumerate}
    Here $d_v$ is the ``perceptual distance'' human vision uses to visually differentiate between images, $\epsilon>0$ is a threshold below which images are indistinguishable from each other \wrt $d_v$, and $\gC$ is a latent set (often convex) in which we perform the latent interpolation\textemdash \eg, the commonly used bounded cube in a 2-D hyperplane spanned by two latent dimensions of $Z$. 
\end{proposition}
This proposition is inspired by the 1-D fact that \emph{every continuous injective function $g:\sR \to \sR$ is strictly monotone}. 
Specifically, due to the continuity of $g$, if we move any $z$ only by a small enough amount, $g(z)$ also changes only by a small amount, so $g(z)$ varies \emph{continuously} as we move it from any $z_1$ to any $z_2$ in the domain $\sR$. 
In addition, the injectivity of $g$ implies that if we move $z$ from $z_1$ to $z_2$, the image $g(z)$ will vary from $g(z_1)$ to $g(z_2)$ \emph{consistently}\textemdash \ie, it changes without producing any duplicate $g(z) = g(z_1)$ or $g(z) = g(z_2)$ for all $z$ between $z_1$ and $z_2$, such that $g(z)$ appears to \emph{never stop varying} as $z$ moves from $z_1$ to $z_2$. 
This is similar in the high dimensional cases, for which we believe the disentanglement learned by most unsupervised models is nothing but a \emph{high dimensional version of strict monotonicity}:
\begin{enumerate}
    \item The continuity of $\restr{g}{\gC}$ prevents abrupt change in $g(z)$ as we linearly interpolate $z$ between any $z_1\in\gC$ and $z_2\in\gC$. The absence of continuity leads to shattered and fragmented representations. We may further require $\restr{g}{\gC}$ to be bi-Lipschitz \wrt the $L_2$ distance in $\gC$ and the perceptual distance $d_v$ in $X$ for a more uniform change of $g(z)$ across $\gC$. 
    
    \item The injectivity of $\restr{g}{\gC}$ is perhaps the most important condition here. It not only means $g(z)$ changes \emph{consistently} if we move any $z\in\gC$ along a single latent dimension~$i$\textemdash \ie, $g(z+\alpha\cdot e_i) = g(z+\beta\cdot e_i) \Leftrightarrow \alpha = \beta$ where $e_i$ is the standard basis vector, but also implies that $g(z)$ must change \emph{differently} if we move $z$ along a different latent dimension~$j$\textemdash \ie, $g(z+\alpha\cdot e_i) = g(z+\beta\cdot e_j) \Leftrightarrow \alpha=\beta=0$. Thus, there comes the modularity and compactness that ``each latent dimension only controls a single factor of data variation distinctively.'' 

    \item The last perceptual metric condition ensures that every $z\in\gC$ is mapped to a realistic image in the vicinity of the dataset $\gX$, so that the latent interpolation on $\gC$ generates valid samples. The low dimensionality of latent space could be helpful for finding $\gC$ satisfying this condition, as will be discussed next. 
\end{enumerate}

\subsubsection{Examples of Disentanglement Conditions}

We argue that both our autoencoders and the generative models in some other works that manifest this disentanglement effect are trained in ways that enforce these three conditions. For instance:
\begin{itemize}
    \item The autoencoders we use have continuous decoders $g$ by construction, and the reconstruction loss $\E_{x\sim\gX}\|x-g\circ e(x)\|$ forces $g$ to be injective over $\gZ=e(\gX)$ and satisfy $g(\gZ) = \gX$. Moreover, as discussed in \S\ref{sec:ldspace}, setting the latent dimension to the intrinsic dimension of $\gX$ gives $\gZ$ non-empty interior $\interior{\gZ} \subseteq Z$, so when we perform the latent interpolation within a 2-D latent linear subspace $Z'\subseteq Z$ that has intersection with $\interior{\gZ}$, the intersection $\interior{\gZ}\cap Z'$ is \emph{open} in $Z'$. Due to this openness, every $z\in \interior{\gZ}\cap Z'$ can have open balls as its neighborhoods\textemdash which are convex\textemdash so we must be able to find some convex $\gC \subseteq \interior{\gZ}\cap Z'\subseteq Z'$ as the latent subset for interpolation. In consequence, $\restr{g}{\gC}$ is a continuous injection, and $g(\gC)\subseteq g(\gZ)=\gX$. It would be ideal if $\gZ$ (or a great part of it) is convex, given that a large $\gC$ can then be found to produce $g(z)$ of large variation.

    \item VAE and its variations, such as $\beta$-VAE and FactorVAE, have loss functions comprising a reconstruction loss and a regularization loss. Unlike deterministic AEs, their encoders are usually modelled by a parametric Gaussian conditional distribution ${q(z\mid x)} = \gN(z\mid \mu(x), \mathrm{diag}(\sigma^2(x)))$. However, their reconstruction loss similarly encourages $x = g\circ\mu(x)$ for $x\in\gX$, which, if achieved, makes the continuous $g$ injective over $\gZ = \mu(\gX)$. 
    Moreover, their regularizers commonly contain the penalty 
    \begin{equation*}
        \E_{x} {D_{KL}(q(z|x)\mid p(z))} = \frac{1}{2} \sum_i \E_{x} \mu_i^2(x) + \frac{1}{2} \sum_i \E_{x}(\sigma_i^2(x) - \log\sigma_i^2(x) - 1)
    \end{equation*}
    where the first term $\frac{1}{2} \sum_i \E_{x} \mu_i^2(x)$ makes $\mu(x)$ sparse over $\gX$, and the second term encourages $\sigma_i(x)$ to converge to 1 if possible. This should induce a similar compression effect that makes $\gZ = \mu(\gX)$ lie in a low dimensional subspace in $Z$. The results in~\citep{higgins2017beta} support our claim, in which the VAEs learned many uninformative latent dimensions with $\mu_i(x)\approx 0$ and $\sigma_i(x)\approx 1$. Therefore, if this compression is effective enough to reduce that latent subspace's dimension to $\gX$'s intrinsic dimension, it will induce a similar disentanglement effect when we interpolate in some 2-D subspaces in it, as inferred in the previous case.  
    
    \item InfoGAN is a GAN with mutual information maximization (MIM). GANs\textemdash typically with continuous generators $g$\textemdash are trained to make $g_*(\sP_z) = \mdata$, where $g_*$ is the pushforward operator of $g$ and $\sP_z$ is a specified latent probability measure. $\sP_z$ is commonly set to a Gaussian or a uniform distribution, whose support $\supp\sP_z$ is a convex set with non-empty interior in $Z$.   
    If we select a $Z'$ intersecting $\interior{\supp\sP_z}$ and choose $\gC \subseteq \interior{\supp\sP_z}\cap Z'$, then a GAN's training process makes $g(\gC)\subseteq \gX \subseteq \supp\mdata$. 
    Furthermore, the MIM of InfoGAN enforces $\E_{z\sim\gC} \|z - e \circ g(z)\| = 0$, where $e$ is an continuous encoder modelling the auxiliary Gaussian distribution, as described in~\citep{chen2016infogan}. This makes $z = e\circ g(z)$ for all $z\in\gC$, thus $\restr{g}{\gC}$ is injective. In contrast, the GANs without MIM do not have the disentanglement effect, as its $\restr{g}{\gC}$ is not necessarily injective and thus $g(\gC)$ can have self-intersections.
\end{itemize}

Therefore, debating the importance of disentangled representations for downstream tasks is essentially discussing the utility of \emph{low dimensionality, continuity, and injectivity} of the representation, at least when we do not define disentanglement out of the scope of Proposition~\ref{prop4:disentanglement}.
Low dimensionality is crucial for efficiency and robustness, as discussed in \S\ref{sec:ldspace}. Continuity is important if the inputs or outputs of the downstream tasks must be continuous \wrt $\gX$. The importance of injectivity, which is the cornerstone of disentanglement, is not clear yet. It might simplify optimization tasks in which the $z = g^{-1}(x)$ for a desired $x\in\gX$ need to be found, since $z$ is then unique. It might also improve our analysis and interpretation of the datasets. Nevertheless, for now injectivity is more like a pleasing byproduct of seeking low dimensional representations using autoencoders, as we need it to make the latent sets homeomorphic to the datasets to avoid trivial solutions. Its usefulness needs further assessment.

\subsection{On the Topological Complexity of Datasets}

So far we have only trained LVAE-DP on a few basic datasets, which are simple in the sense that they are probably simply connected balls. It is interesting to investigate LVAE-DP's efficacy on some datasets of larger scales and potentially of much more complicated topological structures. This includes the whole MNIST dataset and the CelebA dataset.

\subsubsection{MNIST is A Union of Balls}
\label{sec:mnist_lv}

\begin{figure}
    \centering
    \includegraphics[width=\linewidth]{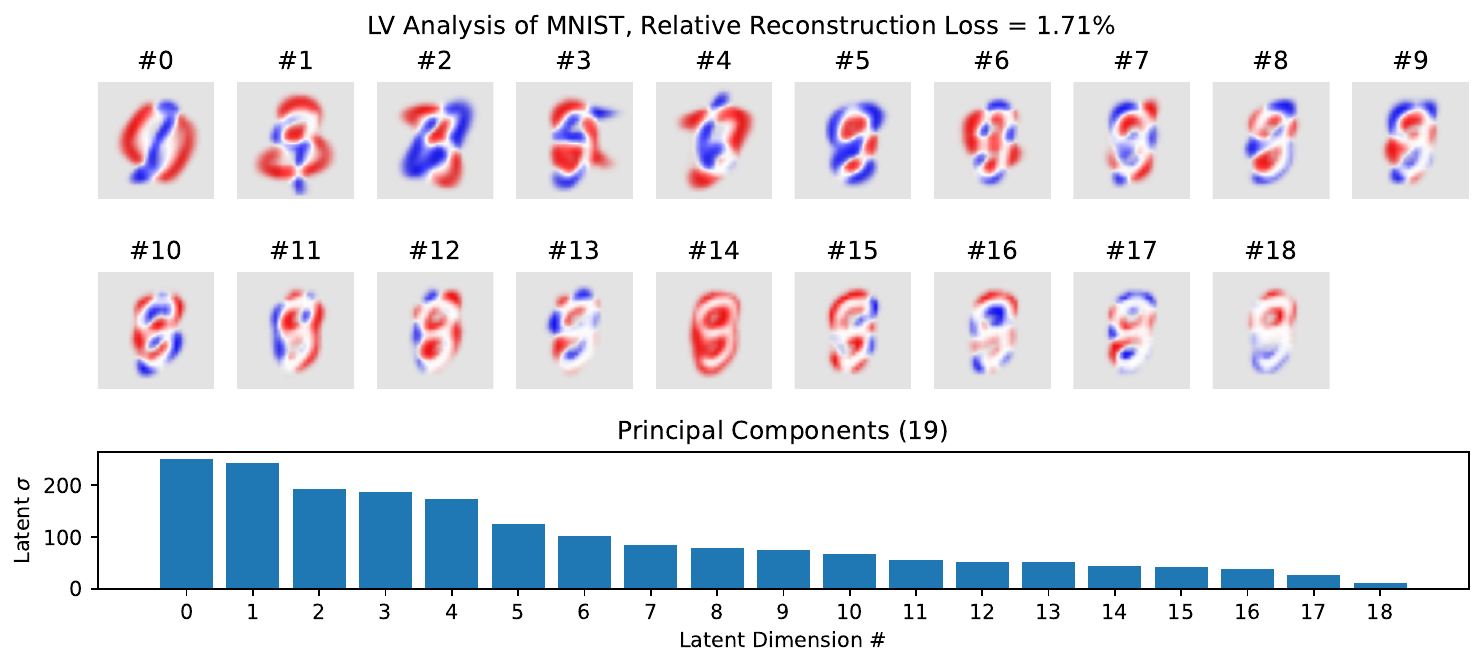}
    \caption{LV Analysis of MNIST}
    \label{fig4:mnist_lv}
\end{figure}

\begin{figure}[bt!]
    \centering
    \begin{subfigure}[b]{0.4\textwidth}
         \centering
         \includegraphics[width=\textwidth]{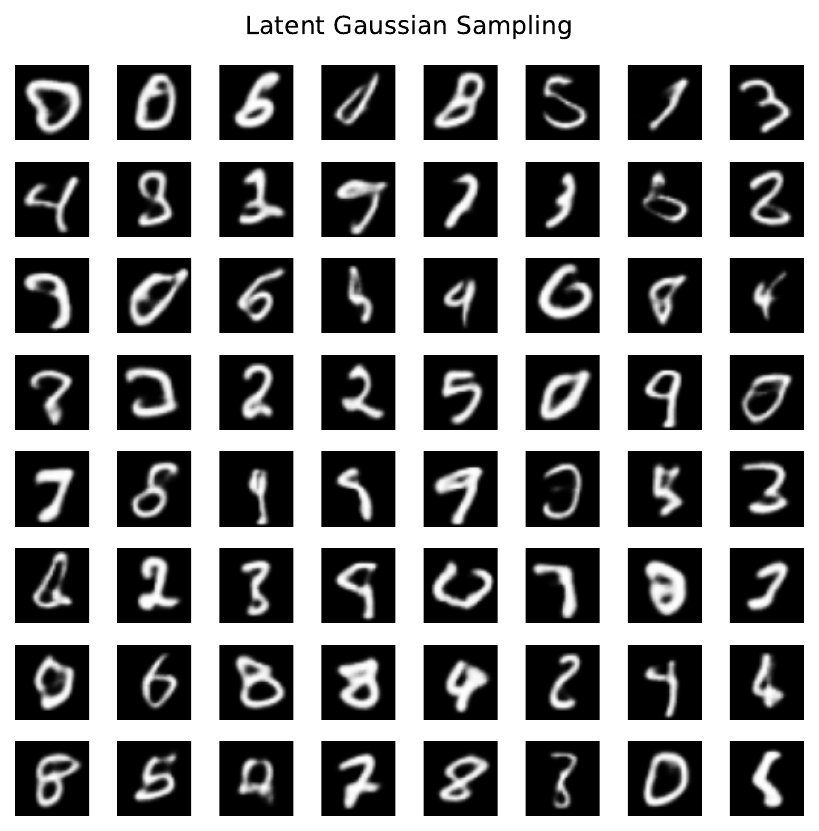}
         \caption{Latent Gaussian Samples}
         \label{fig4:lvint_mnist}
     \end{subfigure}
     \begin{subfigure}[b]{0.5\textwidth}
         \centering
         \includegraphics[width=\textwidth]{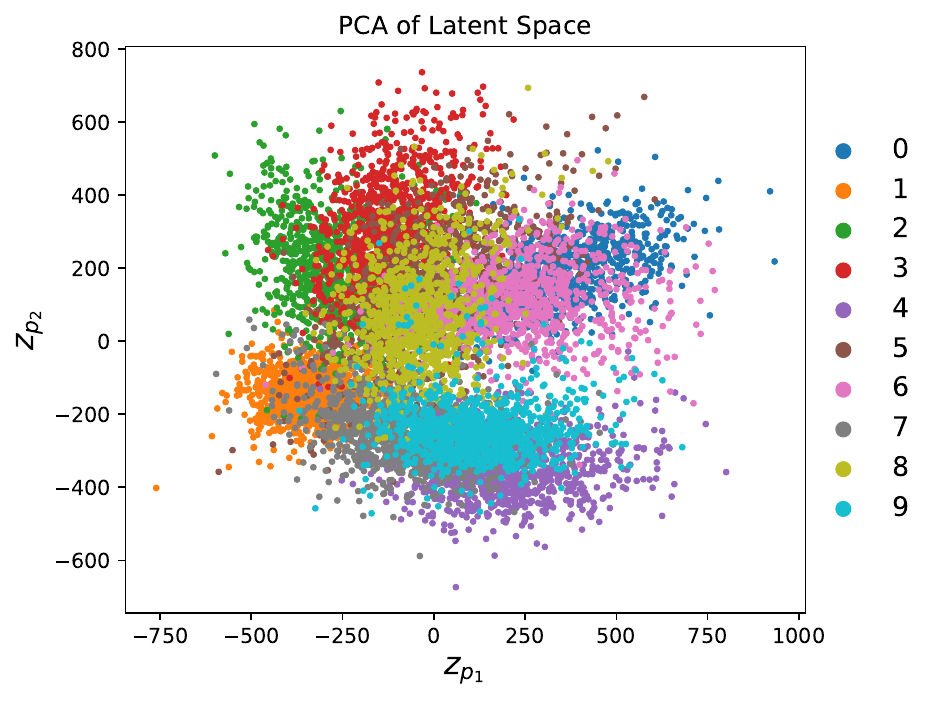}
         \caption{Distribution of Latent Code}
         \label{fig4:mnist_code}
     \end{subfigure}
    \caption{Latent Gaussian samples and the latent code's distribution (projected into the 2-D principal subspace) of the LVAE-DP trained on the entire MNIST dataset.} 
    \label{fig4:mnist_full_sample}
\end{figure}

\begin{figure}[bt!]
     \centering
     \includegraphics[width=\textwidth]{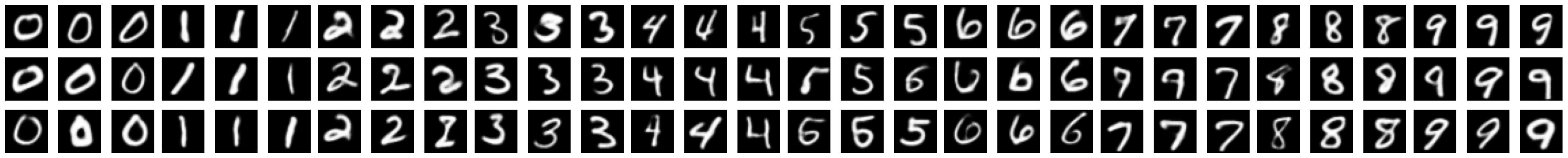}
    \caption{Individual latent Gaussian samples of the 10 digits.} 
    \label{fig4:mnist_individual}
\end{figure}

Let us begin with the MNIST dataset. We have already seen in \S\ref{sec:ldembed} and \S\ref{sec:ldspace} that MNIST is a union of 10 balls of different dimensions, so the embedding dimension of the entire dataset must be no less than the largest dimension of these balls\textemdash 13 (for digit 3). If these 10 balls are disjoint, we may expect the entire MNIST's embedding dimension to agree with this dimension. However, LVAE-DP shows that the embedding dimension of MNIST is about 19 (as shown in Fig.~\ref{fig4:mnist_lv}), which is much larger than 13. This result has several interpretations:
\begin{itemize}
    \item Some digit sets probably have intersections with each other somewhere. The UMAP result in~\citep{mcinnes2018umap} shows that the digit sets of 4 and 9 are probably connected, which makes sense given that 4 and 9 are homotopy-equivalent. The digit sets of 3, 5 and 8 also seem connected for similar reason. Therefore, the whole MNIST dataset may not be locally Euclidean at some of the intersections, which explains why it cannot be embedded in a 13-D latent space. It should be noted that it is generally hard to predict the embedding dimension of two intersecting manifolds of know dimensions, especially for high dimensional ones. 

    \item MNIST's latent set in the 19-D latent space should be much harder to sample using a single latent Gaussian approximation, since it is homeomorphic to the MNIST dataset and thus also consists of 10 balls that may intersect each other. We can see in Fig.~\ref{fig4:lvint_mnist} that many digits produced by the latent Gaussian approximation $\sP_z$ are invalid or distorted, which means $\sP_z(\supp\sP_z\cap\gZ) \ll \sP_z(\supp\sP_z \backslash \gZ)$, \ie, $\gZ$ only constitutes a small portion of the Gaussian ball $\supp\sP_z$. The plot of the MNIST dataset's latent code in Fig.~\ref{fig4:mnist_code} also shows that a single Gaussian should not be an accurate approximation of the entire MNIST's latent code distribution. 

    \item Because the latent space dimension is larger than all digit sets' embedding (or intrinsic) dimensions, each digit's latent set is not necessarily flattened in the latent space, provided that the autoencoder's flattening effect in \S\ref{sec:ldspace} does not exist here. Therefore, if each digit's latent set is curved, its distribution may not be well approximated by a Gaussian. Figure~\ref{fig4:mnist_individual} testifies to this hypothesis, in which we can see that each individual latent Gaussian sampling still produces some distorted and invalid digits, thus some latent codes sampled from the latent Gaussian approximations must fall outside $\gZ$.
\end{itemize}

Based on these findings, we may propose that by construction, autoencoders are not ideal to serve as generative models for \emph{datasets whose embedding dimensions are larger than their intrinsic dimensions}. This could partially explain why so far all purely-autoencoder-based generative models perform poorly on large scale, high dimensional datasets, as many of them could potentially be unions of intersecting manifolds of various intrinsic dimensions that satisfy Assumption~\ref{asp:smooth_manifold}.

\subsubsection{CelebA is a High Dimensional Rocky Planet}
\begin{figure}[bt!]
    \centering
    \begin{subfigure}[b]{\textwidth}
         \centering
         \includegraphics[width=\textwidth]{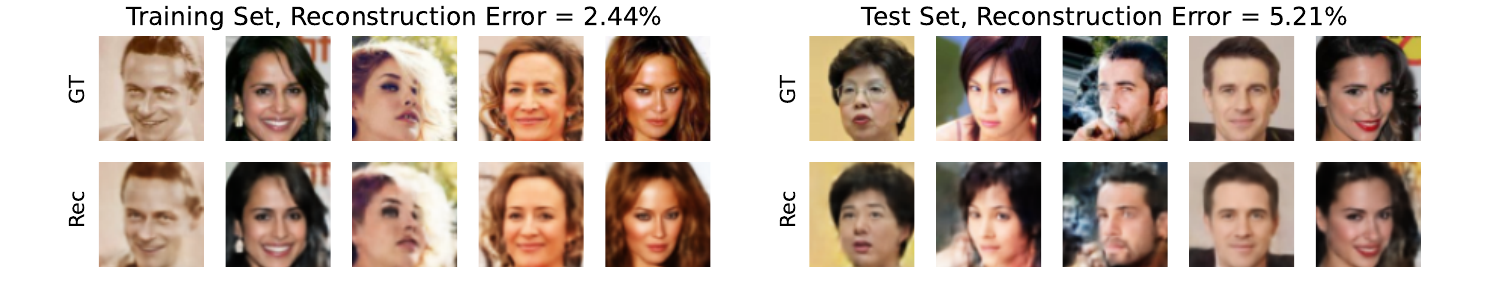}
     \end{subfigure}
     \begin{subfigure}[b]{\textwidth}
         \centering
         \includegraphics[width=\textwidth, trim={1cm 0 1.5cm 8cm},clip]{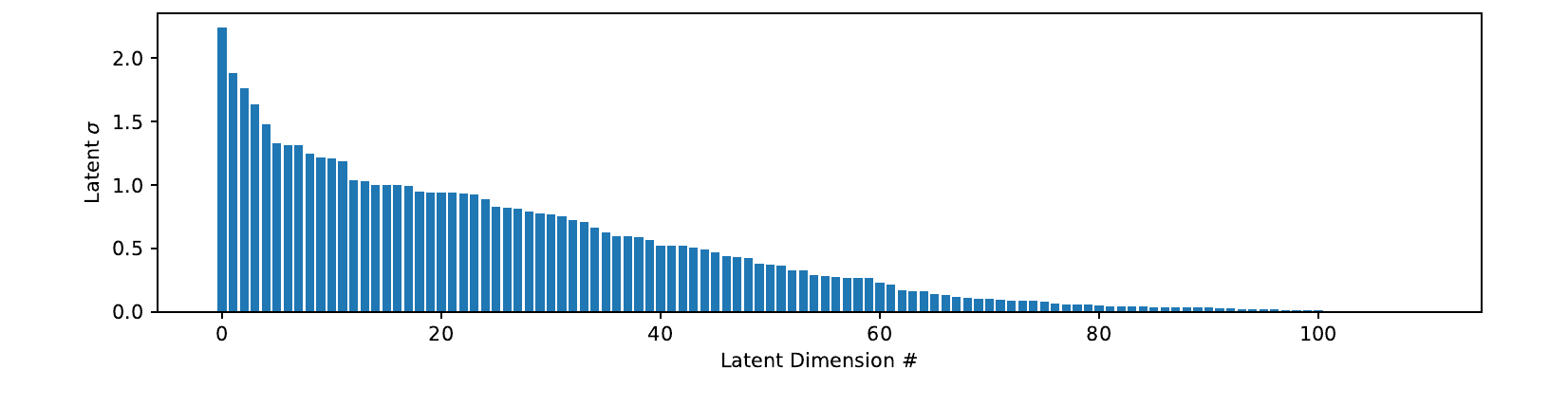}
     \end{subfigure}
    \caption{LVDP-AE's reconstruction of CelebA images and the distribution of 110 latent $\sigma$.} 
    \label{fig4:celeba_rec_lat}
\end{figure}

Next, we inspect LVAE-DP's results on the CelebA dataset~\citep{liu2015deep}, which is more complicated than MNIST. Figure~\ref{fig4:celeba_rec_lat} shows that LVAE-DP achieves great dimension reduction performance on it, reducing the autoencoder's latent dimension from 4000 all the way to 110, which might be an accurate estimate of the embedding dimension or even the intrinsic dimension of $64\times 64$ human face images. Its reconstruction error on both the training set and test set are also low enough to make the facial patterns well reconstructed, and there is no blurriness typically seen in the VAEs' results. 

\begin{figure}[bt!]
    \centering
    \begin{subfigure}[b]{0.48\textwidth}
         \centering
         \includegraphics[width=\textwidth, trim={0 0 0 0.5cm},clip]{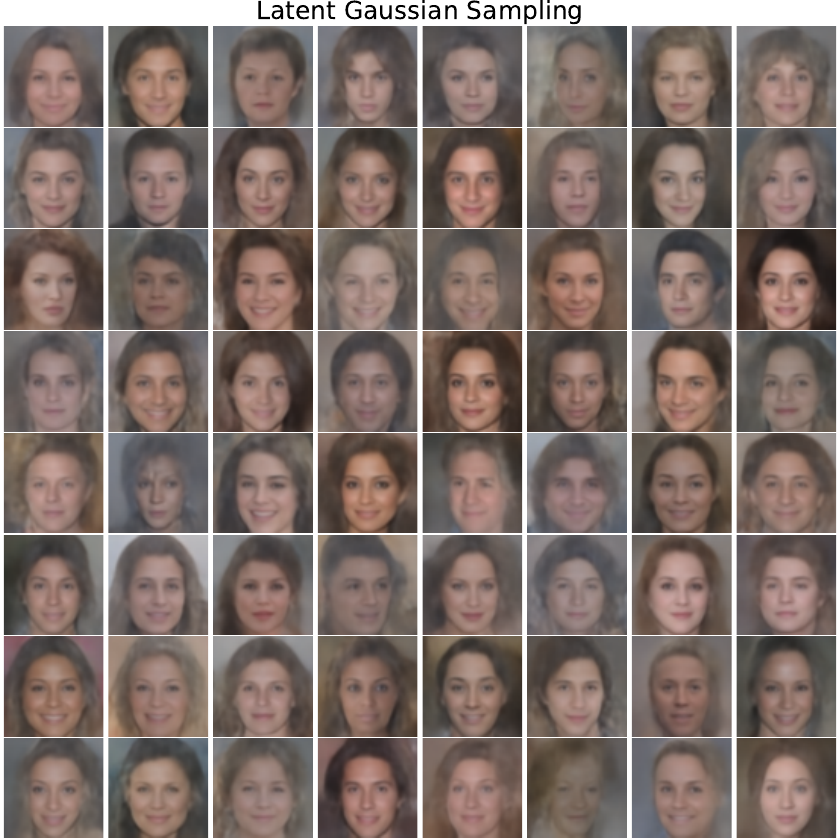}
         \caption{Latent Gaussian sampling with $\Sigma / 10$}
         \label{fig4:celeba_sample_0.3}
     \end{subfigure}
     \hfill
    \begin{subfigure}[b]{0.48\textwidth}
         \centering
         \includegraphics[width=\textwidth, trim={0 0 0 0.5cm},clip]{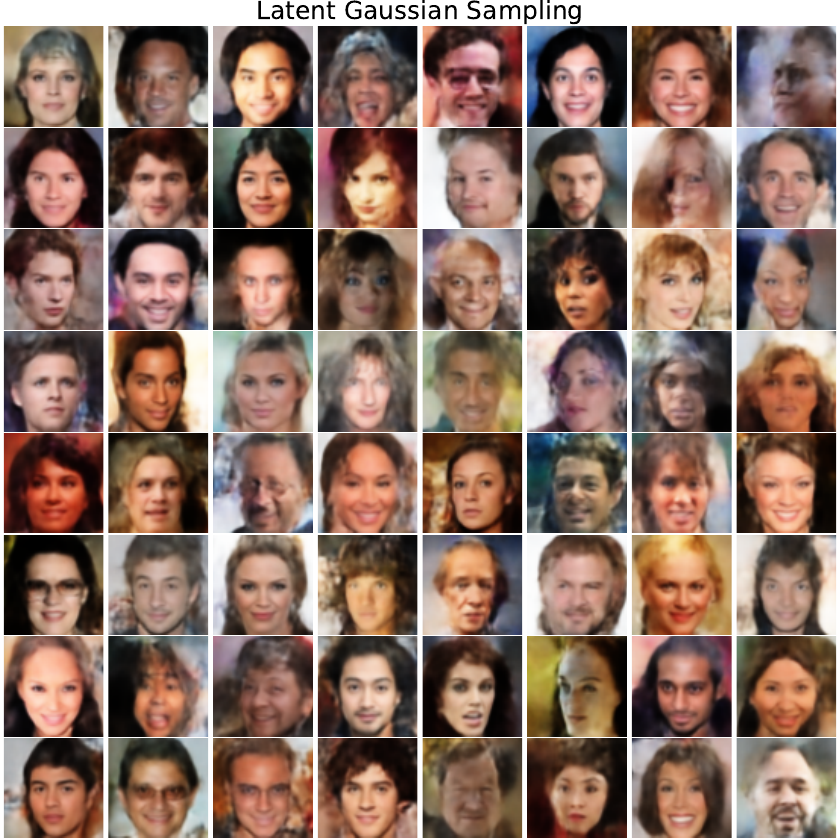}
         \caption{Latent Gaussian sampling with full $\Sigma$}
         \label{fig4:celeba_sample_1}
     \end{subfigure}
    \caption{Samplings in the CelebA LVAE-DP's latent space.} 
    \label{fig4:celeba_sampling}
\end{figure}

\begin{figure}[bt!]
    \centering     
     \begin{subfigure}[b]{\textwidth}
         \centering
         \includegraphics[width=0.32\textwidth]{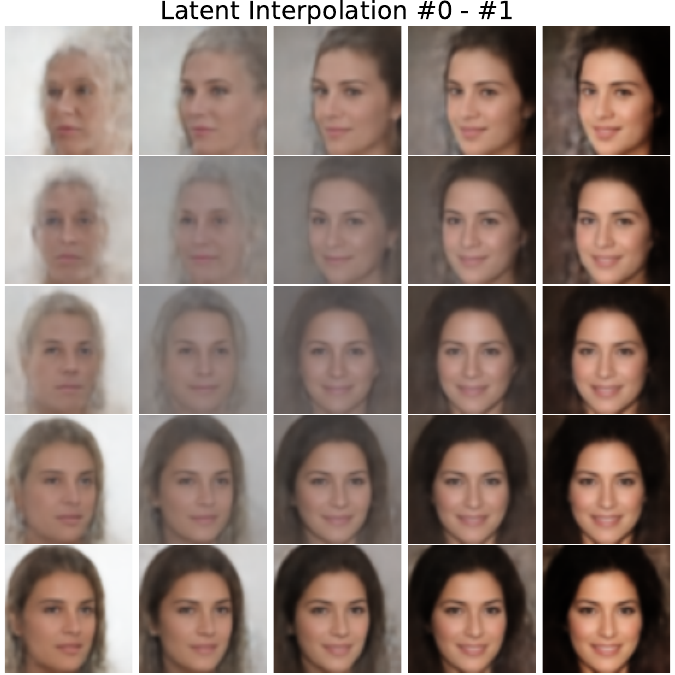}
         \includegraphics[width=0.32\textwidth]{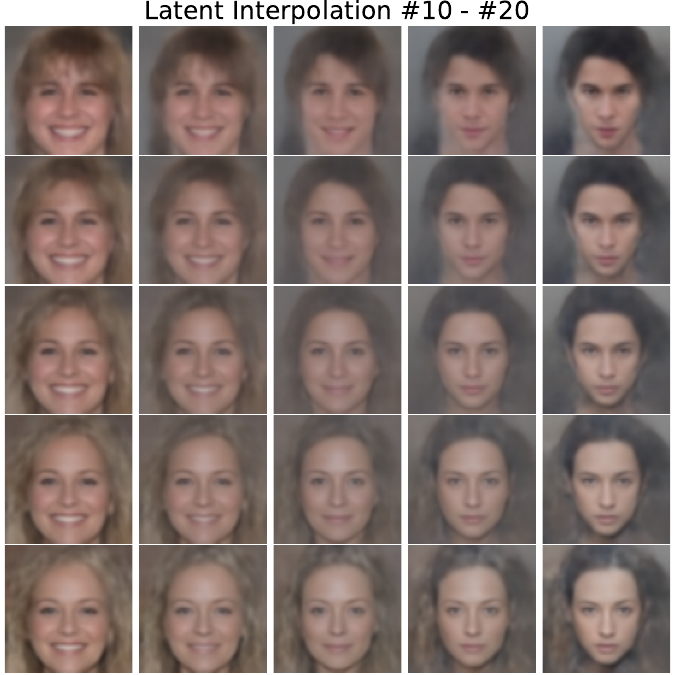}
         \includegraphics[width=0.32\textwidth]{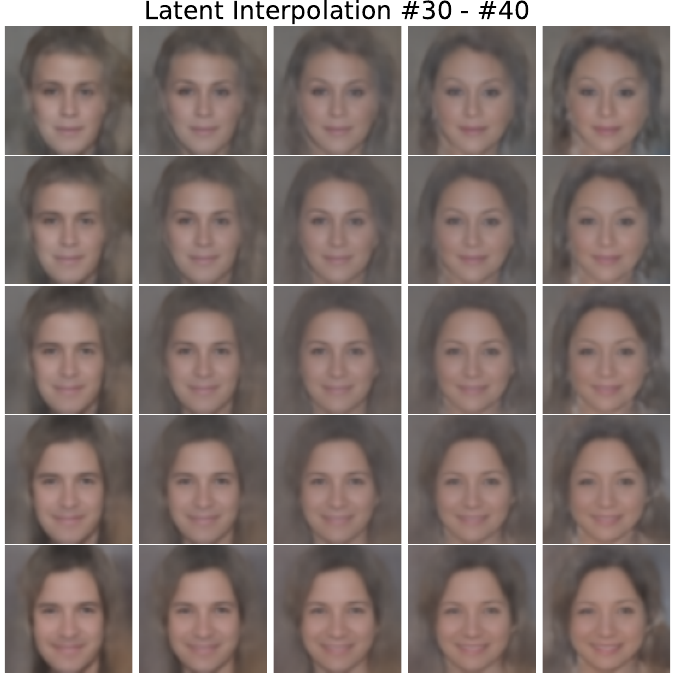}
         \caption{Latent interpolations around ``smiling lady''\textemdash the center of $\gZ$}
         \label{fig4:celeba_int_mean}
     \end{subfigure}

     \begin{subfigure}[b]{\textwidth}
         \centering
         \includegraphics[width=0.32\textwidth]{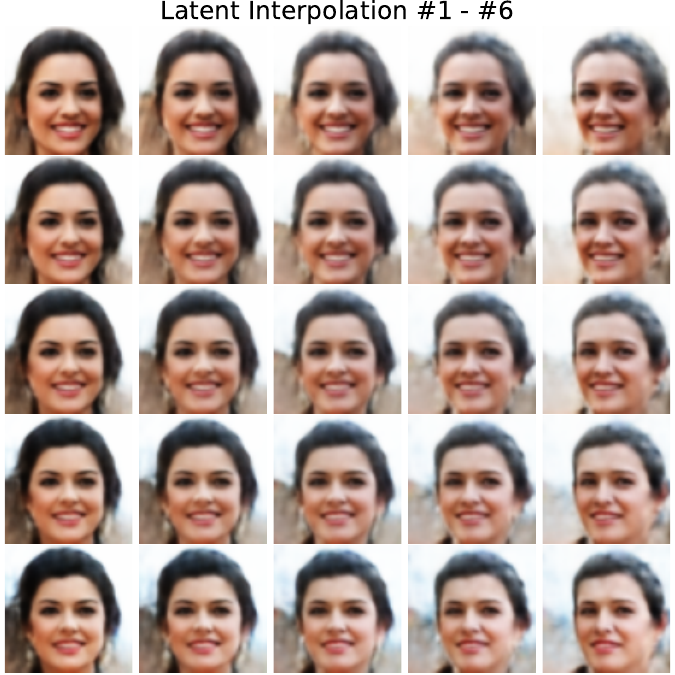}
         \includegraphics[width=0.32\textwidth]{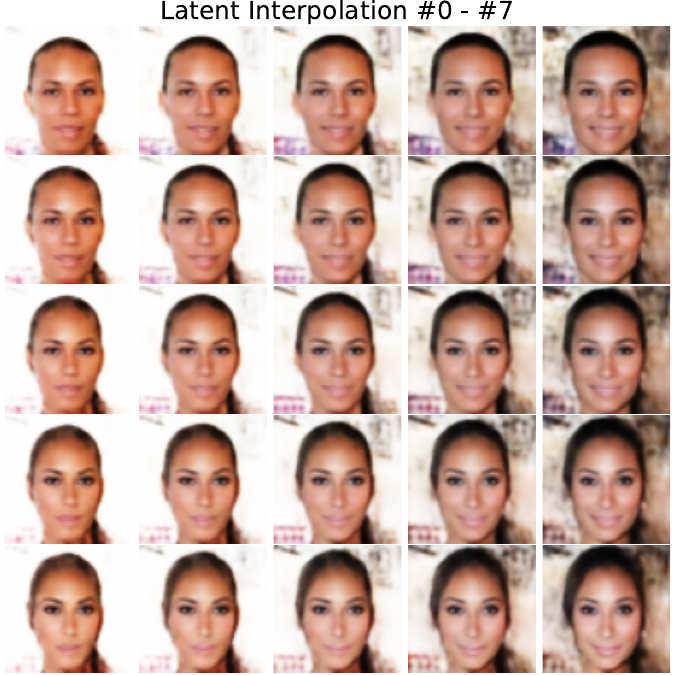}
         \includegraphics[width=0.32\textwidth]{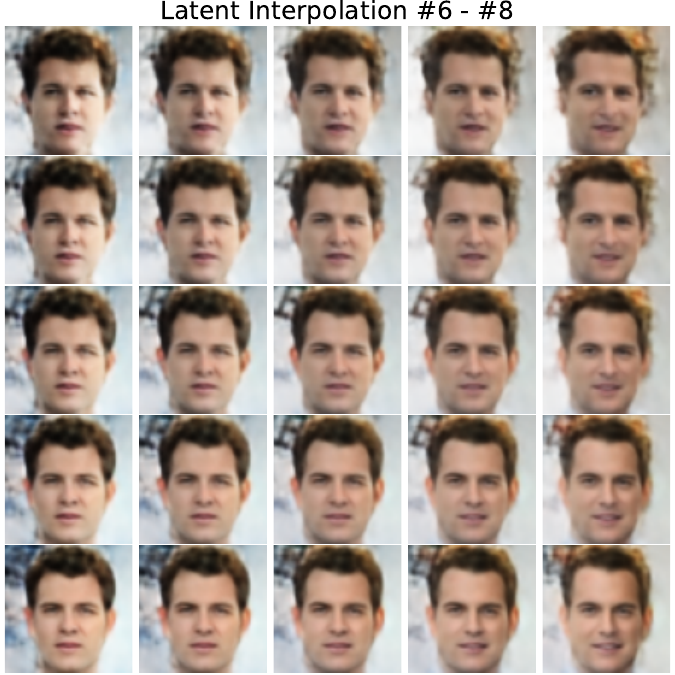}
         \caption{Latent interpolations around different data samples}
         \label{fig4:celeba_int_sample}
     \end{subfigure}

     \begin{subfigure}[b]{0.75\textwidth}
         \centering
         \includegraphics[width=\textwidth, trim={0 2cm 0 0},clip]{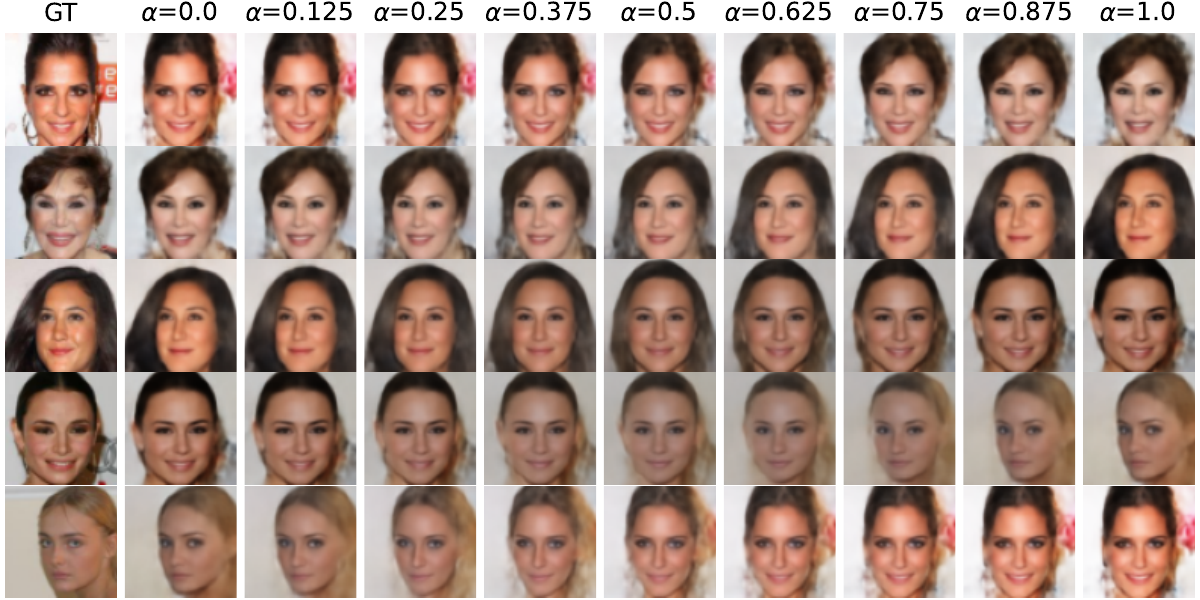}
         \caption{Latent interpolations between data sample pairs}
         \label{fig4:celeba_lat_int}
     \end{subfigure}
     \begin{subfigure}[b]{0.215\textwidth}
         \centering
         \includegraphics[width=\textwidth]{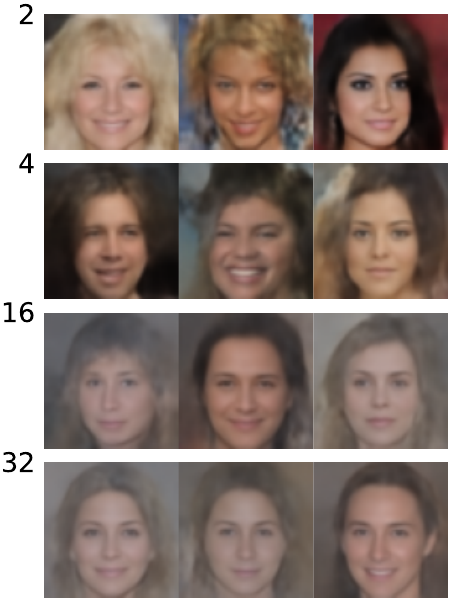}
         \caption{Multicombinations}
         \label{fig4:celeba_cvx}
     \end{subfigure}
    \caption{Interpolations in the CelebA LVAE-DP's latent space.} 
    \label{fig4:celeba_topo}
\end{figure}

As to the latent set's topological structure, we can observe the following phenomena:
\begin{enumerate}
    \item (Figure~\ref{fig4:celeba_sample_0.3} and~\ref{fig4:celeba_sample_1}) Latent Gaussian sampling produces many images with either distorted backgrounds or even distorted faces, which means the latent set may not have a simple topological structure upon which these latent codes easily fall. However, if we shrink down the latent Gaussian approximation's covariance matrix by 10, so that we only sample latent codes in the vicinity of the latent set's mean (or center), the decoder produces images with realist faces but grayish and blurry backgrounds. These stand in contrast to the much sharper reconstructions of true images in the dataset, suggesting that the decoder indeed has the ability to produce sharp images, but the sharp images' latent codes are uneasy to sample using a Gaussian distribution. 
    
    \item (Figure~\ref{fig4:celeba_int_mean} and~\ref{fig4:celeba_int_sample}) Latent interpolation at the latent set's mean produces face variations in a variety of factors, from primary ones like age, gender, pose and race, to minor ones like jaw opening, eye expression and mood. Not only that, the interpolation also works properly when we perform the interpolation at different images from the dataset. This suggests the latent set should in general have non-empty interior in the final latent space, and each data point very likely has an open neighborhood in the latent space.

    \item (Figure~\ref{fig4:celeba_lat_int} and~\ref{fig4:celeba_cvx}) The linear latent interpolation between an arbitrary pair of images usually leads to valid and smooth face transition, despite situations where sometimes the interpolation could yield distorted human faces along the way. However, as we increase the number of data samples to be convexly combined in the latent space (from 2 to 4 to 16 to 32, as shown in Fig.~\ref{fig4:celeba_cvx}), their convex combination gets closer to the latent set's center and looks more grayish back in the data space. In addition, although sometimes the latent convex combination of a few samples produces invalid images with distorted faces, this becomes much less frequent as we increase the number of samples to be combined.  
\end{enumerate}

One plausible explanation of these findings is that the CelebA dataset\textemdash or, more generally, the set of all human face images\textemdash is a high dimensional \emph{oval ball with rough and spiky surface} (like our mother Earth), in which most of the colorful and sharp face images locate near the surface, whereas the desaturated and blurry ones are around the center of the ball. This hypothesis can naturally explain the phenomena respectively as follows: 
\begin{enumerate}
    \item In high dimensional spaces, the Gaussian distributions are like soap bubbles, with most of their mass concentrated around the surfaces of their supports~\citep{vershynin2018high}\textemdash which are practically balls. This means if we perform the latent Gaussian sampling without any adjustment, we are almost always sampling latent points near the support ball's \emph{spherical surface}, and we can only effectively sample the interior of the ball by scaling down the covariance matrix. Therefore, if the latent set $\gZ$ is a ball with rugged and spiky surface, then the Gaussian distribution's surface-sampling may sometimes produce invalid samples falling between the ``valleys'' and  ``canyons'' on the surface and thus outside the latent set, as we can see in Fig~\ref{fig4:celeba_sample_1}. In contrast, sampling the interior always leads to valid samples inside the ball, as Fig~\ref{fig4:celeba_sample_0.3} shows. 

    \item A spiky and rugged ball is still topologically a ball, so the latent set $\gZ$ has a non-empty interior that can intersect a given 2-D latent plane and produce a latent intersection $\gC$ with non-empty interior on this plane, such that all the points inside $\gC$ can yield valid images and leads to the disentanglement effect, as discussed in \S\ref{sec:disentangle}.

    \item If we randomly pick two points on the surface of that rugged ``planet'' $\gZ$ and connect them using a straight line, then this line can intersect $\gZ$ and yield valid samples when 1) the points are far away from each other, \eg, in different hemispheres, or 2) the points are close to each other but both are at the ``sea level'', so the ``crust'' of $\gZ$ can still block\textemdash \ie, intersect\textemdash the straight line. On the other hand, the line circumvents $\gZ$ and produces invalid human faces when at least one data point is on a tall spiky ``mountain'' like Everest, such that no part of $\gZ$ blocks the straight line. 
\end{enumerate}

\begin{figure}[bt!]
    \centering
    \includegraphics[width=0.8\linewidth]{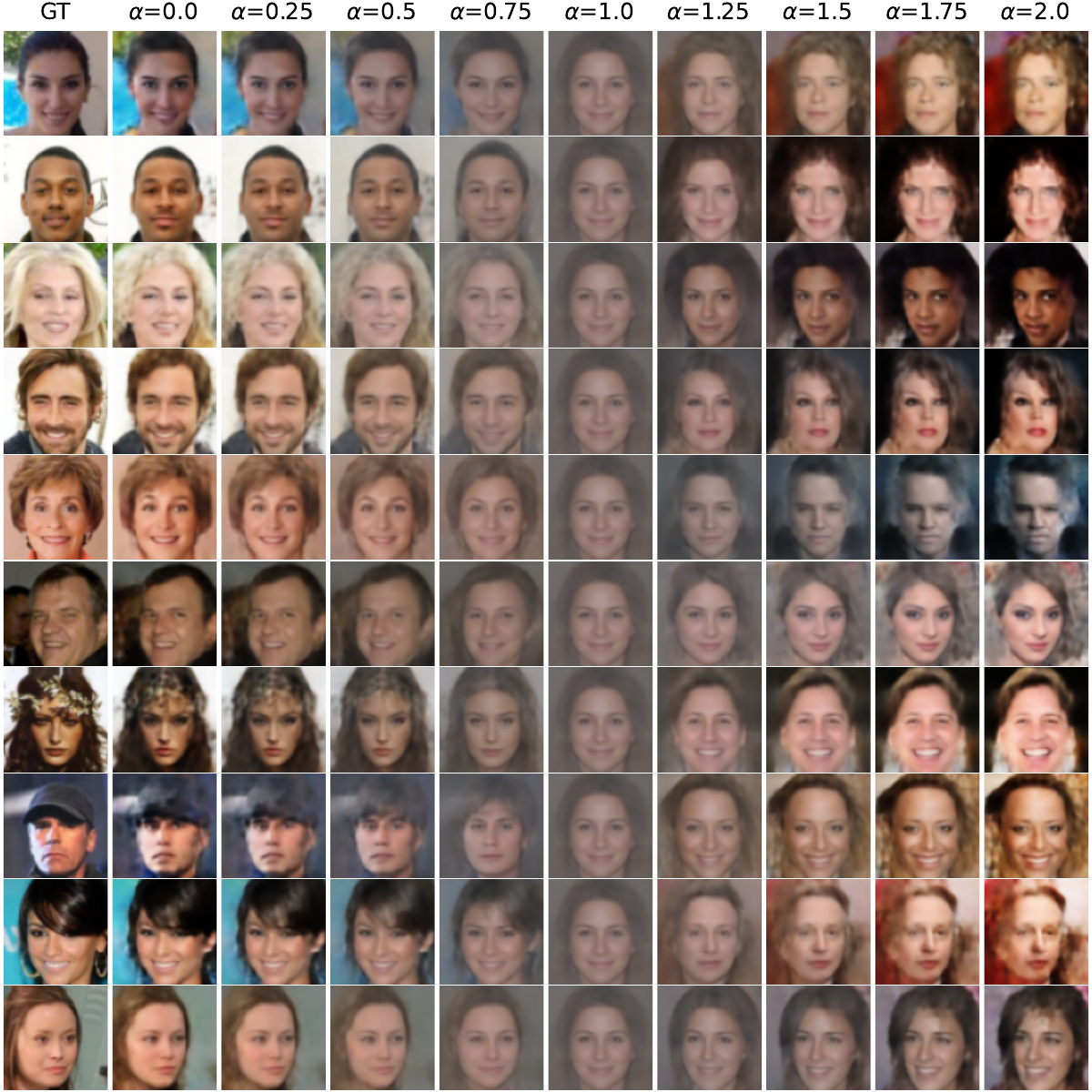}
    \caption{Latent antipodes test\textemdash the other side of the face image planet is still a face image.}
    \label{fig4:diameter_int}
\end{figure}

To further verify this hypothesis, we can perform the \emph{latent antipodes test}: We start from several data samples' latent codes $z_0$ and drill holes along the potential planet $\gZ$'s diameter all the way to its core\textemdash the mean $\mu(\gZ)$, then continue drilling for the same distance until we come out of the other side of $\gZ$, and inspect what we get along the way. In other words, we linearly interpolate in the latent space along the line $z(\alpha) = (1-\alpha)z_0 + \alpha \mu(\gZ)$ from $\alpha = 0$ to $\alpha=1$ then to $\alpha = 2$, and feed $z(\alpha)$ to the decoder $g$ to produce images. 
The result in Fig.~\ref{fig4:diameter_int} verifies that the other side of ``planet'' $\gZ$ is still a valid human face image, which further supports our hypothesis.

\section{Experiments: Supervised Least Volume Analysis}
\label{sec:exp_slva}

In this second primary experiment section, we test GLV on two datasets of different types of labels\textemdash the MNIST dataset with a discrete label set of digit categories, and the UIUC dataset with a connected label set of lift and drag coefficients. We investigate GLV's effect on the latent representations to verify our anticipations made at the end of \S\ref{sec:lvlabel}.

\subsection{MNIST with Categorical Labels}
We formulate the GLV problem for MNIST as per Corollary~\ref{corollary4:labellvmax}. The architecture of the GLV autoencoder for MNIST is not very different from the unlabelled one in \S\ref{sec:mnist_lv}, except for two modifications. First, we introduce an additional 1-Lipschitz label predictor $g^y_\theta: Z\to Y$, where the 10 labels are one-hot encoded to make their mutual distances equal to 1. When training the GLV autoenocder, we connect this predictor to softmax activation and train it with cross entropy as usual. 
Second, we increase the Lipschitz constant of the image decoder $g^x_\theta: Z\to X$ to $1024$ to scale down the latent set's length scale to around 1 (in contrast to the large length scale in Fig.~\ref{fig4:mnist_code} caused by the 1-Lipschitz $g^x_\theta$), so that the label predictor $g^y_\theta$ and the one-hot labels can adjust the distances between the latent codes noticeably. 

\subsubsection{Clustering Effect}
\begin{figure}
    \centering
    \includegraphics[width=\linewidth]{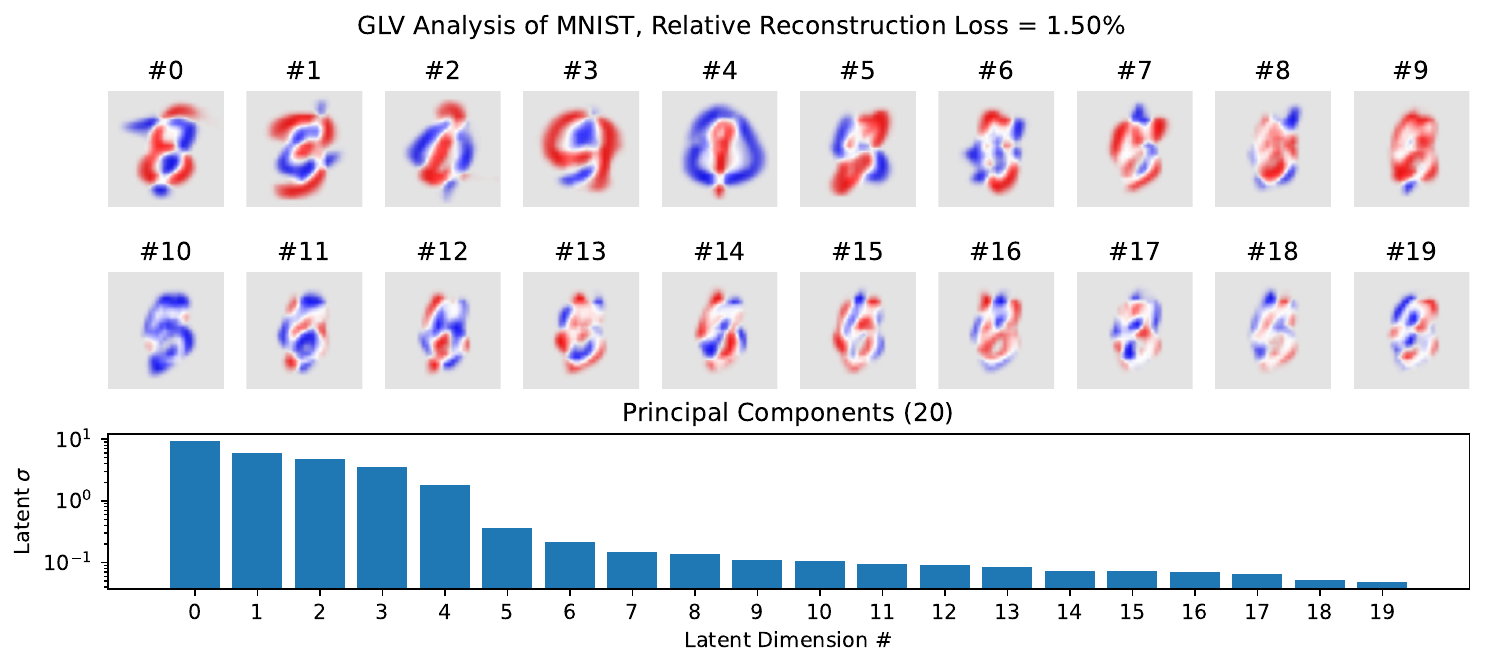}
    \caption{Latent dimensions of GLV autoencoder of MNIST.}
    \label{fig4:glv_mnist}
\end{figure}

\begin{figure}[bt!]
    \centering
    \includegraphics[width=0.7\linewidth]{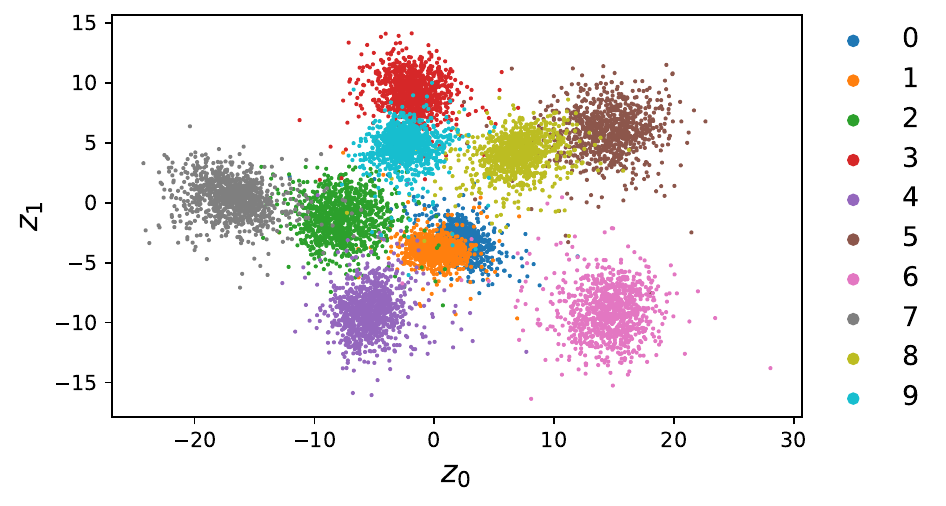}
    \caption{Distribution of MNIST digits in GLV latent space.}
    \label{fig4:glv_mnist_code}
\end{figure}

\begin{figure}[bt!]
    \centering
    \begin{subfigure}[b]{\textwidth}
        \includegraphics[width=0.46\textwidth]{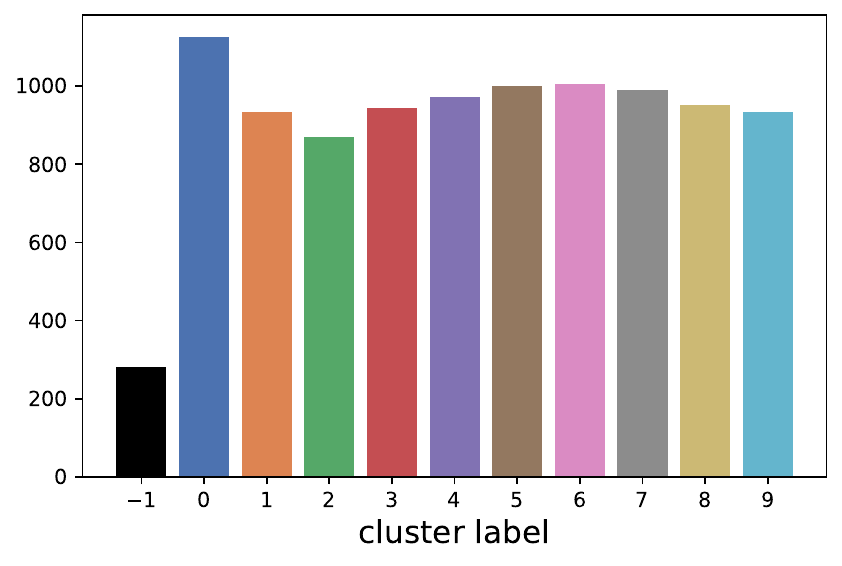}
        \includegraphics[width=0.5\textwidth]{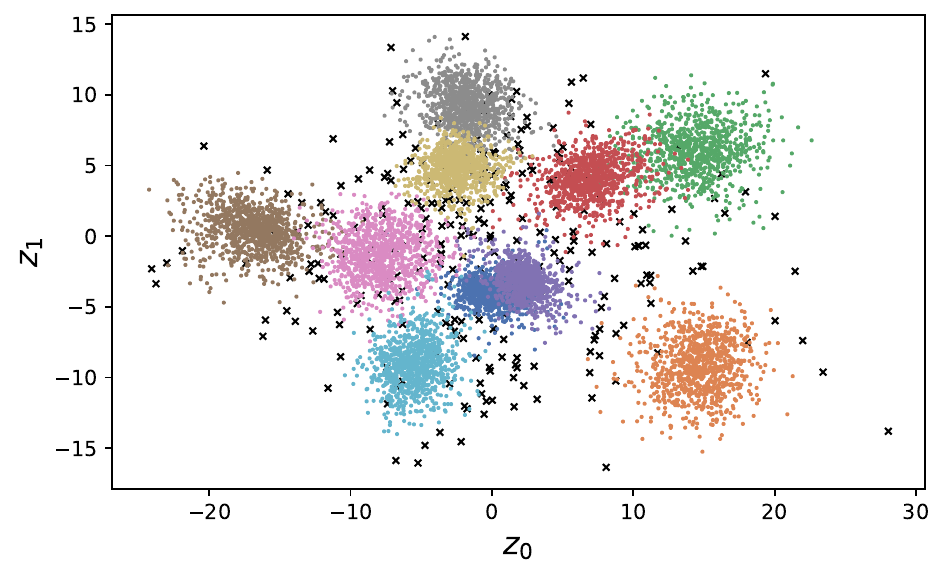}
        \caption{Distribution of HDBSCAN clusters}
    \end{subfigure}
    \begin{subfigure}[b]{0.8\textwidth}
        \includegraphics[width=\textwidth]{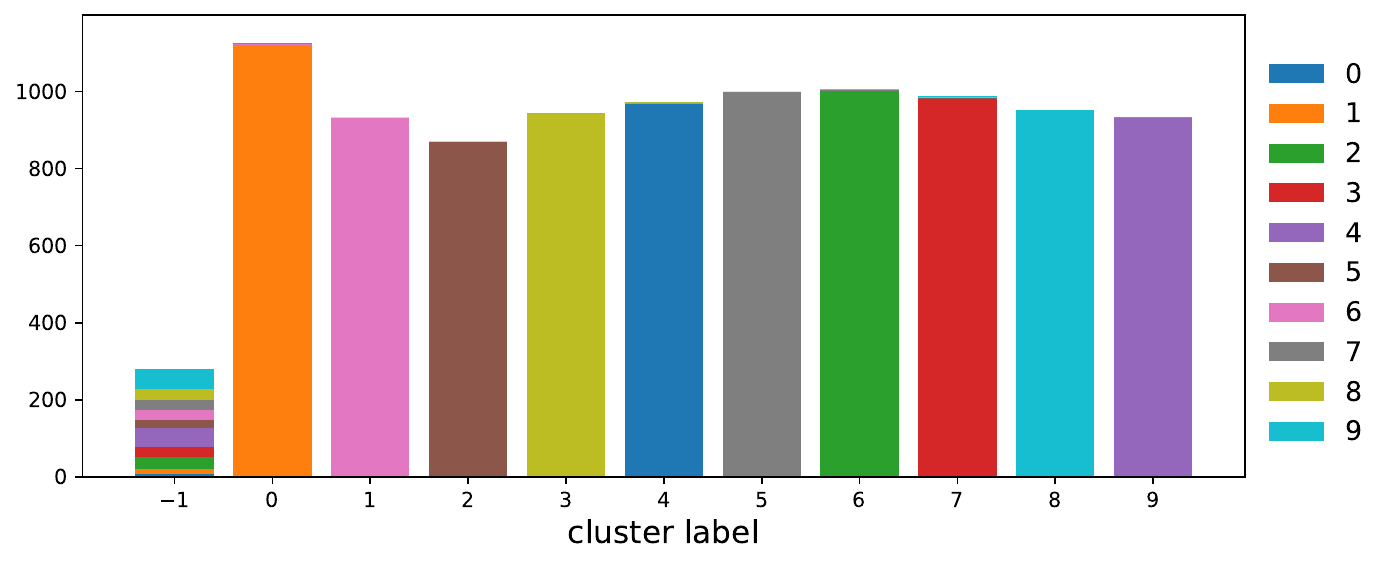}
        \caption{Contents of HDBSCAN clusters}
        \label{fig4:glv_mnist_cluster_digit}
    \end{subfigure}
    \caption{HDBSCAN result on MNIST's GLV representation.}
    \label{fig4:glv_mnist_hdbscan}
\end{figure}

Figure~\ref{fig4:glv_mnist} shows the 20 principal components extracted by GLV, which is only one more than the 19 components extracted by LV in \S\ref{sec:mnist_lv}. In contrast to Fig.~\ref{fig4:mnist_lv}, there are 5 major latent dimensions whose length scales are about one order of magnitude larger than the rest, which is possibly induced by the additional label information. If we project the latent codes into the 2-D subspace spanned by dimension \#0 and \#1 then plot them out in Fig.~\ref{fig4:glv_mnist_code}, we can notice a distinctive clustering pattern absent from Fig.~\ref{fig4:mnist_code}: the images from different classes converge into distinct clusters that look well-separated from one another. To see if they are truly well-separated (given that some clusters overlap in the 2-D space), we apply HDBSCAN~\citep{mcinnes2017hdbscan} on the latent set. We set its parameters $m_\text{clSize}$ to 50 and $m_\text{pts}$ to 5. Its result in Fig.~\ref{fig4:glv_mnist_hdbscan} shows that this density-based clustering algorithm can tell the clusters apart sufficiently, as each cluster (except the noise cluster \#-1) consists almost exclusively of a single MNIST digit (as shown in the stacked bar chart Fig.~\ref{fig4:glv_mnist_cluster_digit}). 
In comparison, HDBSCAN classifies around 70\% latent points in Fig.~\ref{fig4:mnist_code} as noise, showing no similar clustering effect in LV's representation. 
Therefore, GLV, when applied on a dataset with categorical labels, indeed has an effect resembling contrastive learning that makes the MNIST images of different labels stay apart from each other, while images of the same class cluster closely together in the latent space. This confirms our anticipation in \S\ref{sec:lvlabel}.

\subsubsection{Semi-supervised GLV}
Not always are all data samples in a dataset paired with labels. It is thus intriguing to inspect how GLV operates when only a few labelled samples are given while the rest of the dataset remains unlabelled. To this end, we perform the same experiment in Fig.~\ref{fig4:glv_mnist_hdbscan} three times but using different amounts of labelled data samples\textemdash 50\%, 10\% and 2\% of the training set\textemdash to train the predictor $g^y_\theta$. 

\begin{figure}
    \centering
    \begin{subfigure}[b]{0.62\textwidth}
        \includegraphics[width=0.49\textwidth]{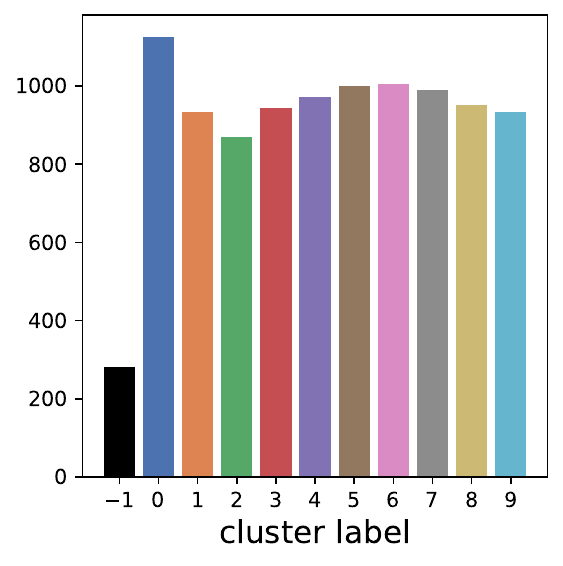}
        \includegraphics[width=0.49\textwidth]{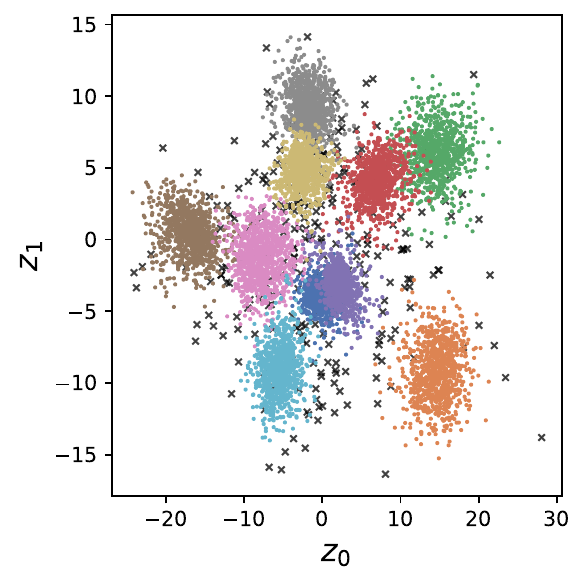}
        \caption{Distribution of HDBSCAN clusters (100\% labels)}
    \end{subfigure}
    \begin{subfigure}[b]{0.37\textwidth}
        \includegraphics[width=\textwidth]{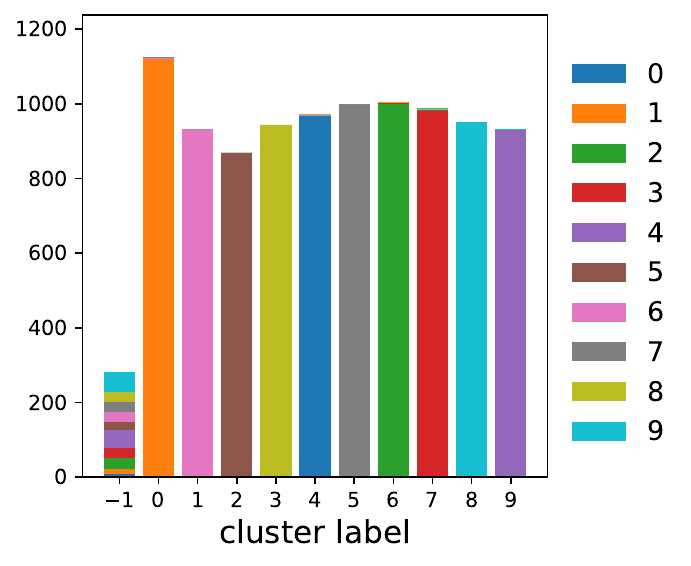}
        \caption{Contents (100\% labels)}
    \end{subfigure}
    \begin{subfigure}[b]{0.62\textwidth}
        \includegraphics[width=0.49\textwidth]{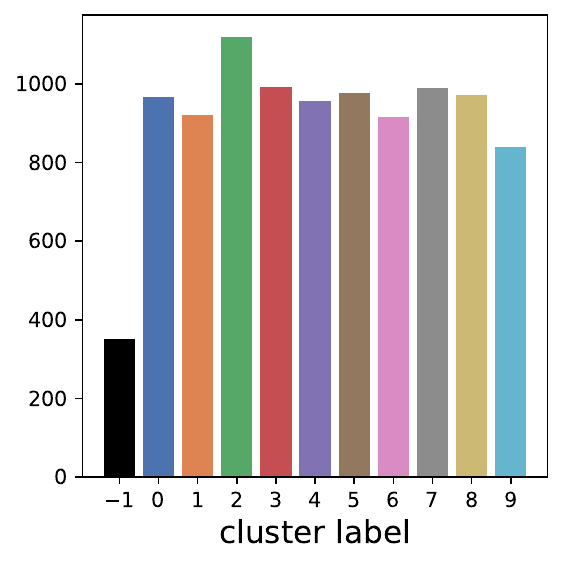}
        \includegraphics[width=0.49\textwidth]{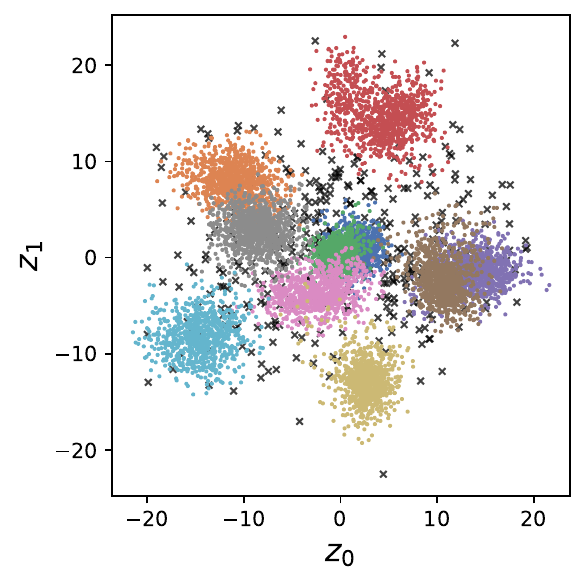}
        \caption{Distribution of HDBSCAN clusters (50\% labels)}
    \end{subfigure}
    \begin{subfigure}[b]{0.37\textwidth}
        \includegraphics[width=\textwidth]{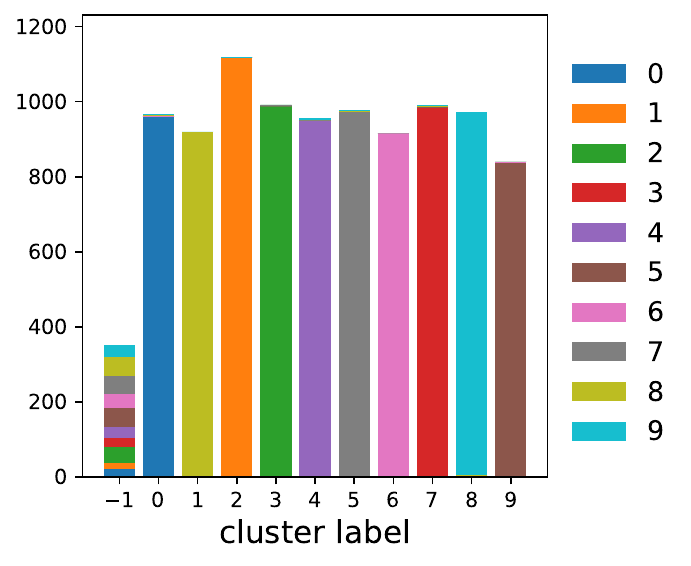}
        \caption{Contents (50\% labels)}
    \end{subfigure}
    \begin{subfigure}[b]{0.62\textwidth}
        \includegraphics[width=0.49\textwidth]{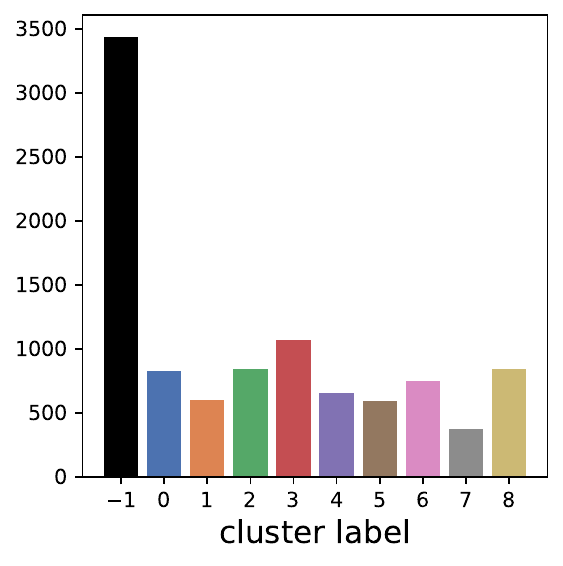}
        \includegraphics[width=0.49\textwidth]{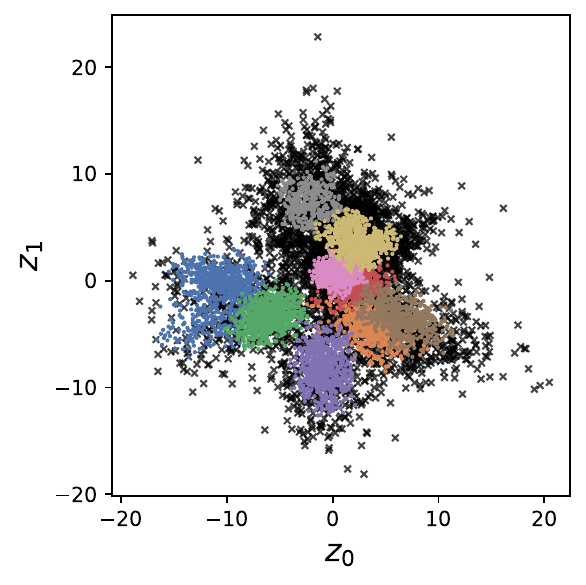}
        \caption{Distribution of HDBSCAN clusters (10\% labels)}
    \end{subfigure}
    \begin{subfigure}[b]{0.37\textwidth}
        \includegraphics[width=\textwidth]{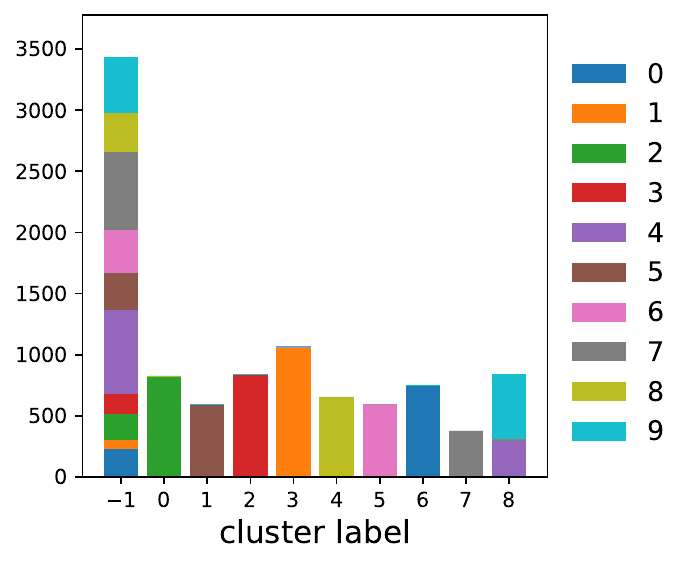}
        \caption{Contents (10\% labels)}
    \end{subfigure}
    \begin{subfigure}[b]{0.62\textwidth}
        \includegraphics[width=0.49\textwidth]{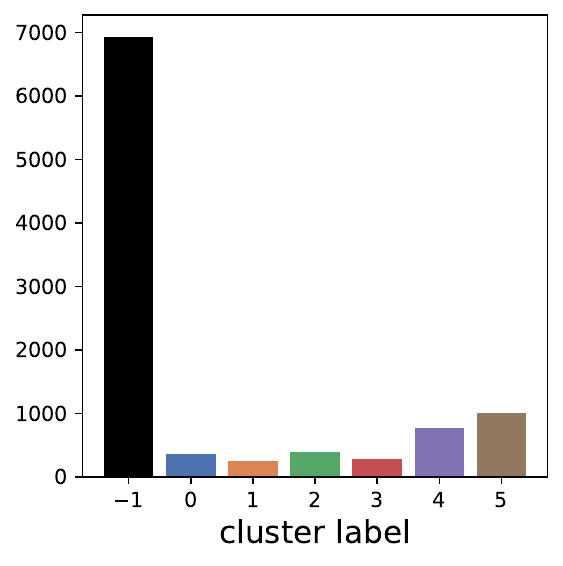}
        \includegraphics[width=0.49\textwidth]{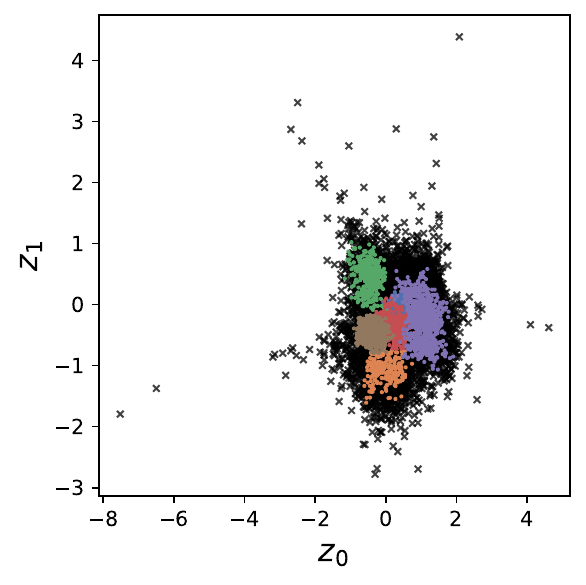}
        \caption{Distribution of HDBSCAN clusters (2\% labels)}
    \end{subfigure}
    \begin{subfigure}[b]{0.37\textwidth}
        \includegraphics[width=\textwidth]{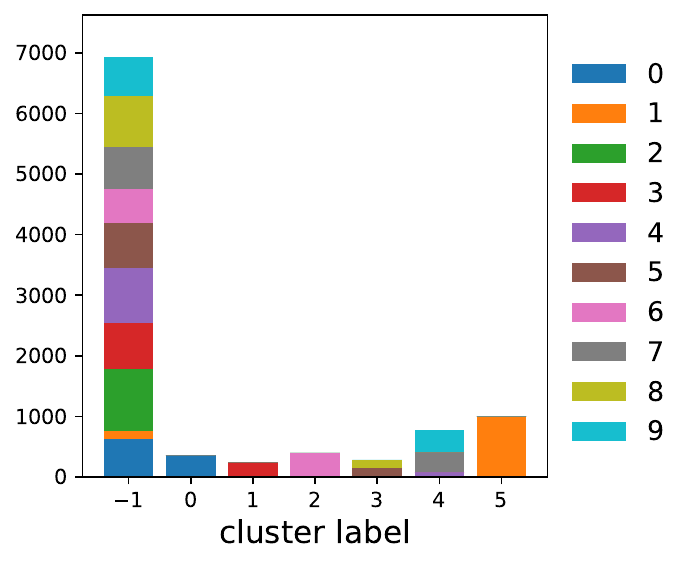}
        \caption{Contents (2\% labels)}
    \end{subfigure}
    \caption{HDBSCAN result on MNIST's GLV representation with different amount of labels.}
    \label{fig4:glv_mnist_hdbscan_semi}
\end{figure}

The results are shown in Fig.~\ref{fig4:glv_mnist_hdbscan_semi}. It is apparent that as we decrease the amount of labelled samples for training the GLV autoencoder, HDBSCAN starts to regard more latent codes as noise and more often allocate different digits into the same clusters, and these clusters get closer to each other. This happens when the latent codes fail to conglomerate to form clusters effectively. However, when the amount of labelled samples is 50\%, the result is comparable to the 100\% case in Fig.~\ref{fig4:glv_mnist_hdbscan}. Therefore, to make GLV induce the clustering effect on a labelled dataset, we do not necessarily need all data samples to be labelled, but the amount of labelled samples must be sufficient.

\subsection{UIUC Airfoils with Lift and Drag Coefficients}
\label{sec:experiments_slva_airfoil}

In airfoil optimization, we normally care about the airfoil's lift coefficient $C_L$ and drag coefficient $C_D$ under different environmental conditions. To investigate which parts of the airfoil shape contribute the most to these two crucial coefficients, we pass the UIUC airfoil dataset to a CFD simulator to generate a labelled airfoil dataset for GLV analysis. The dataset is created by first taking the Cartesian product between the airfoil shapes and a variety of Mach number (Ma), Reynolds number (Re) and Angle of Attack (AoA), then passing the samples from this product to a CFD simulator to derive the corresponding lift and drag coefficient. 
All airfoil simulations were run using the MACH-Aero framework\footnote{https://github.com/mdolab/MACH-Aero}. In the framework, structured volume meshes were extruded using pyHyp, a hyperbolic mesh generator~\citep{PyHyp}, which were then used to run Reynolds Averaged Navier Stokes (RANS) simulations in the open source ADflow solver~\citep{ADFLOW}. All simulations used the Spalart-Allmaras model to capture turbulent effects. Additionally the solver made use of the Newton–Krylov (ANK) method for improved convergence properties~\citep{ANK}. 
The resulting dataset has 46900 samples of the triplet form $(x, c, y)$, in which $x$ denotes airfoil shape, $c$ denotes boundary conditions (Ma, Re and AoA) and $y$ denotes performance ($C_L$ and $C_D$). Here Ma ranges from 0.4 to 0.9, Re ranges from 1e6 to 2e7, and AoA ranges from $0^\circ$ to $20^\circ$. 

\subsubsection{GLV Formulation}
The airfoil's GLV problem takes the form in Corollary~\ref{corollary4:labellvmax}, with a variation in the architecture of $g^y_\theta$. In this application, $g^y_\theta$ is a surrogate for predicting $C_L$ and $C_D$ given $x$ and $c$, so it has the form $g^y_\theta(z, c)$ instead of $g^y_\theta(z)$, taking the boundary condition $c$ as an additional input. Despite that, if we assume that the performance mapping $X\times C\to Y$ is continuous, then we can view $g^y_\theta: Z\times C\to Y$ equivalently as $g^{y}_\theta(c): Z \to Y$. Thus, we may generalize the constraints in Corollary~\ref{corollary4:labellvmax} by requiring them to hold under all $c$.
With this generalization, the reconstruction loss can now be reformulated as ${\E_{x,c,y} \|(g^x_\theta(e_\theta(x)), g^y_\theta(e_\theta(x), c)) - (x, y)\|_2}$, and the second part of the constraint (\ref{eq4:lipinfty}) becomes $\max_c\|g^y_\theta(z_1, c) - g^y_\theta(z_2, c)\|_2 \leq K \|z_1- z_2\|_2$. In other words, $g^{y}_\theta(z, c)$ should be $K$-Lipschitz \wrt $z$ at every $c$. Intuitively, enforcing this constraint in GLV should push two airfoils $x_1$ and $x_2$ far away from each other if there exists any condition $c$ under which they have very different coefficients $y_1$ and $y_2$, making the performance $y$ never change abruptly as $z$ varies.
To make $g^y_\theta$ only $K$-Lipschitz \wrt $z$, we can construct it as $g^y_\theta(z, c) = f_\theta(z, e^c_\theta(c))$ where $f_\theta$ is $K$-Lipschitz but $e^c_\theta$ is not. 

\subsubsection{Procrustes Analysis of Latent Airfoil Representation}
\label{sec:procrustes}
\begin{figure}[bt!]
    \centering
    \begin{subfigure}[b]{\textwidth}
         \centering
         \includegraphics[width=\textwidth]{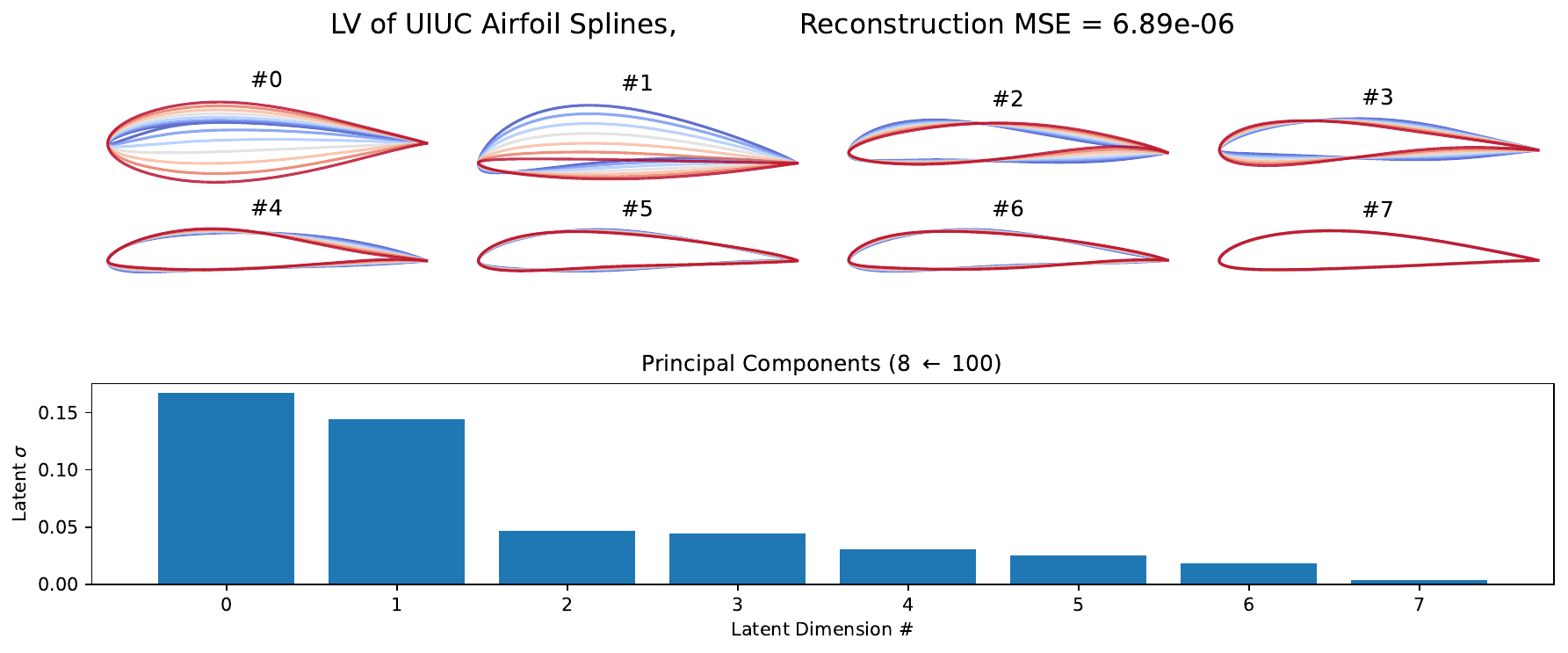}
     \end{subfigure}
     \begin{subfigure}[b]{\textwidth}
         \centering
         \includegraphics[width=\textwidth]{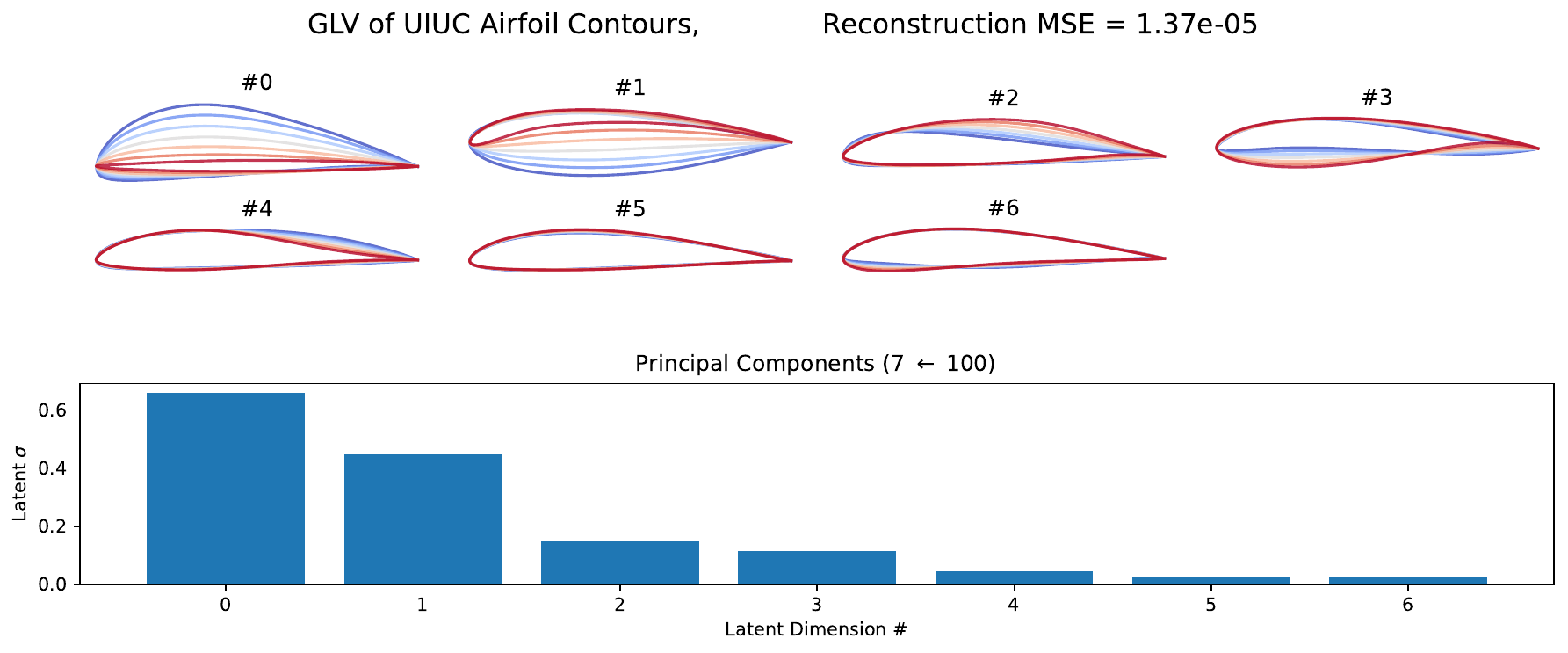}
     \end{subfigure}
    \caption{LV and GLV results on UIUC airfoils. To illustrate the airfoil shape variation that each latent dimension controls, after obtaining the latent set's mean $\vmu$, for each latent dimension $i$ with STD $\sigma_i$ and standard basis $\ve_i$, we superimpose the airfoils $g^x_\theta(\vmu + \alpha \cdot \sigma_i \ve_i)$ for $\alpha\in\{-3, -2, -1, 0, 1, 2, 3\}$ using hues shifting from blue to red.} 
    \label{fig4:airfoil_glv_lv}
\end{figure}

Our results in Fig.~\ref{fig4:airfoil_glv_lv} indicate that GLV produces a representation distinct from that derived by the unlabelled LV. Although it is not an exact match, we can still notice that the latent dimension \#0 and \#1 of LV's result are seemingly swapped in GLV's result, suggesting that that the shape variation controlled by \#1 (of LV) should have more contribution to the change in lift and drag coefficient than \#0. Apart from that, GLV's dynamic pruning also manages to eliminate one more latent dimension than LV, although LV's last dimension can also be safely removed manually since it has marginal STD and negligible contribution to shape variation. In addition, we can see that in both cases, the latent dimensions with the largest STDs in general induces the largest shape variations. This means the lift and drag coefficients are not very sensitive to shape variation, \ie, a small airfoil shape variation generally cannot induce a significant performance variation.

\begin{figure}[bt!]
    \centering
    \begin{subfigure}[b]{0.48\textwidth}
         \centering
         \includegraphics[width=\textwidth]{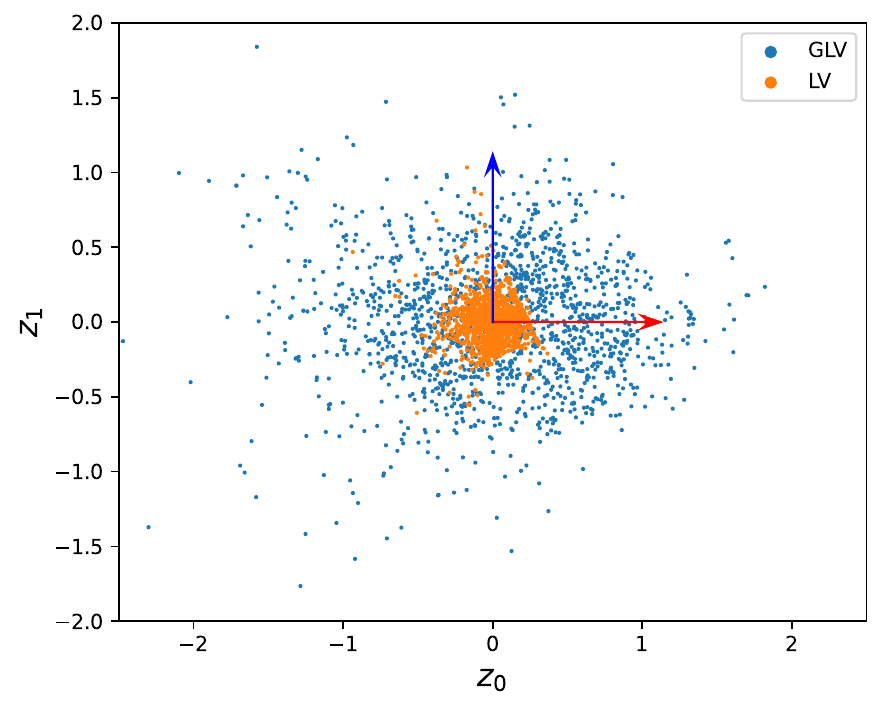}
         \caption{Latent sets after centering}
         \label{fig4:airfoil_prin}
     \end{subfigure}
     \begin{subfigure}[b]{0.48\textwidth}
         \centering
         \includegraphics[width=\textwidth]{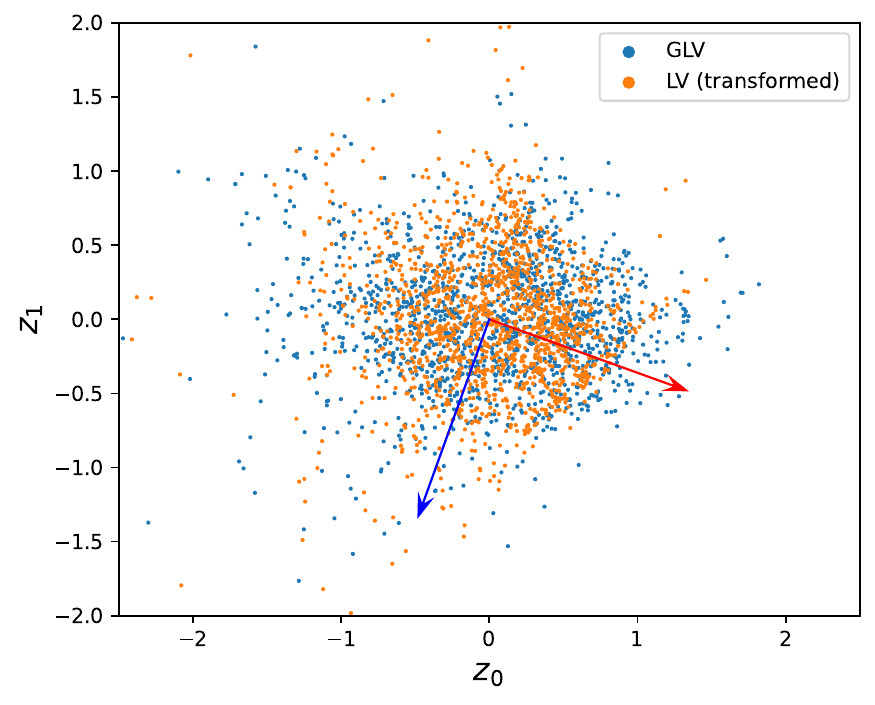}
         \caption{Latent sets aligned by Procrustes Analysis}
         \label{fig4:airfoil_procrustes}
     \end{subfigure}
     \begin{subfigure}[b]{0.48\textwidth}
         \centering
         \includegraphics[width=\textwidth]{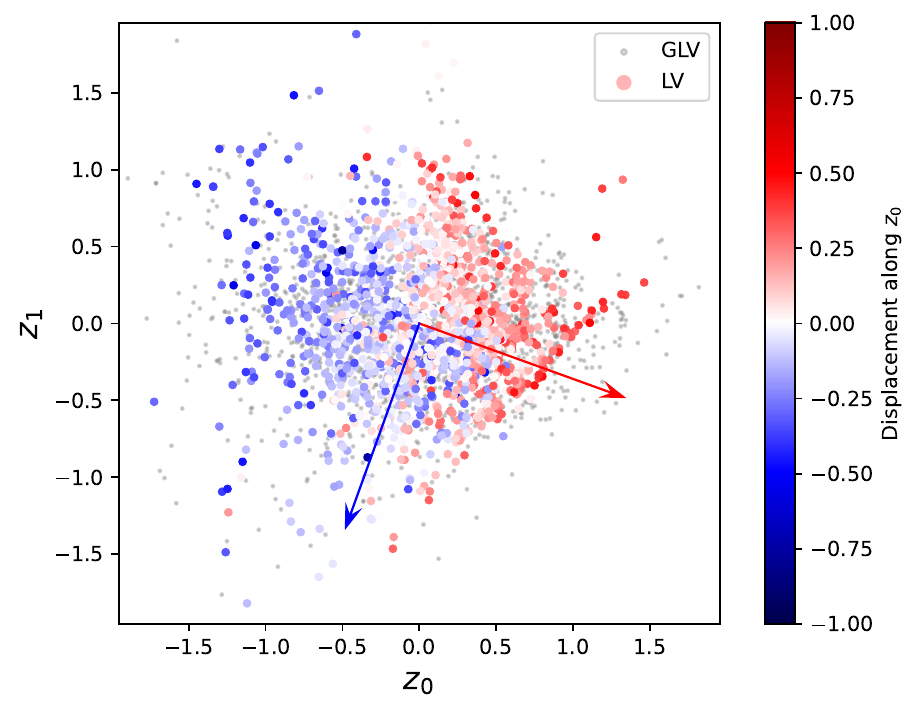}
         \caption{Displacement of LV latent codes along $z_0$}
         \label{fig4:airfoil_procrustes_dis0}
     \end{subfigure}
     \begin{subfigure}[b]{0.48\textwidth}
         \centering
         \includegraphics[width=\textwidth]{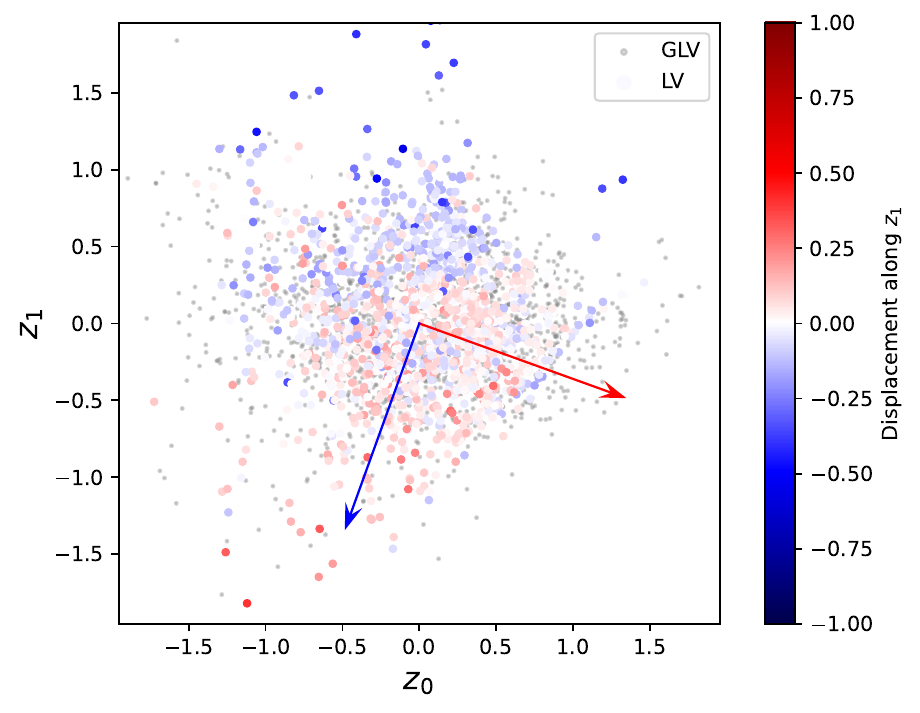}
         \caption{Displacement of LV latent codes along $z_1$}
         \label{fig4:airfoil_procrustes_dis1}
     \end{subfigure}
    \caption{Distribution of LV and GLV latent codes of UIUC airfoils.} 
    \label{fig4:airfoil_lat_code}
\end{figure}

It is also intriguing to inspect how the overall distribution of the UIUC airfoils' latent codes changes after introducing their performance information into the LV problem. For this purpose, we employ Procrustes Analysis~\citep{gower1975generalized} to rotate, flip and uniformly scale up the LV latent set, such that it agrees as much as possible with the GLV latent set in terms of minimizing the MSE, then we superimpose the two transformed latent sets for comparison. In other words,  we ignore the LV latent set's homogeneous transformation and only focus on the heterogeneous one induced by incorporating the performance information. 
The results are illustrated in Fig.~\ref{fig4:airfoil_lat_code}. Figure~\ref{fig4:airfoil_prin} first shows what the two latent sets look like before the Procrustes transformation. The latent codes are projected onto the latent subspace spanned by dimension \#0 and \#1. We can see that that the LV latent set is about 3 to 4 times smaller than the GLV latent set, suggesting that the $K$-Lipschitz surrogate $g^y_\theta$ scales up the latent set's overall length scale. This also agrees with the latent $\sigma$ values in Fig.~\ref{fig4:airfoil_glv_lv}. 
Figure~\ref{fig4:airfoil_procrustes} displays the two latent sets after Procrustes analysis, where the LV latent set is transformed while the GLV latent set is fixed. Figure~\ref{fig4:airfoil_procrustes_dis0} and~\ref{fig4:airfoil_procrustes_dis1} demonstrates the displacement of the LV latent codes along dimension \#0 and \#1 (of GLV). It reveals the LV latent set's stretching along $z_{0}$, shrinking along $z_{1}$, and rotation in the $z_0$--$z_1$ subspace, which explains why the LV latent set's dimension \#0 and \#1 looks swapped in GLV's result. 

\begin{figure}
    \centering
    \includegraphics[width=\linewidth]{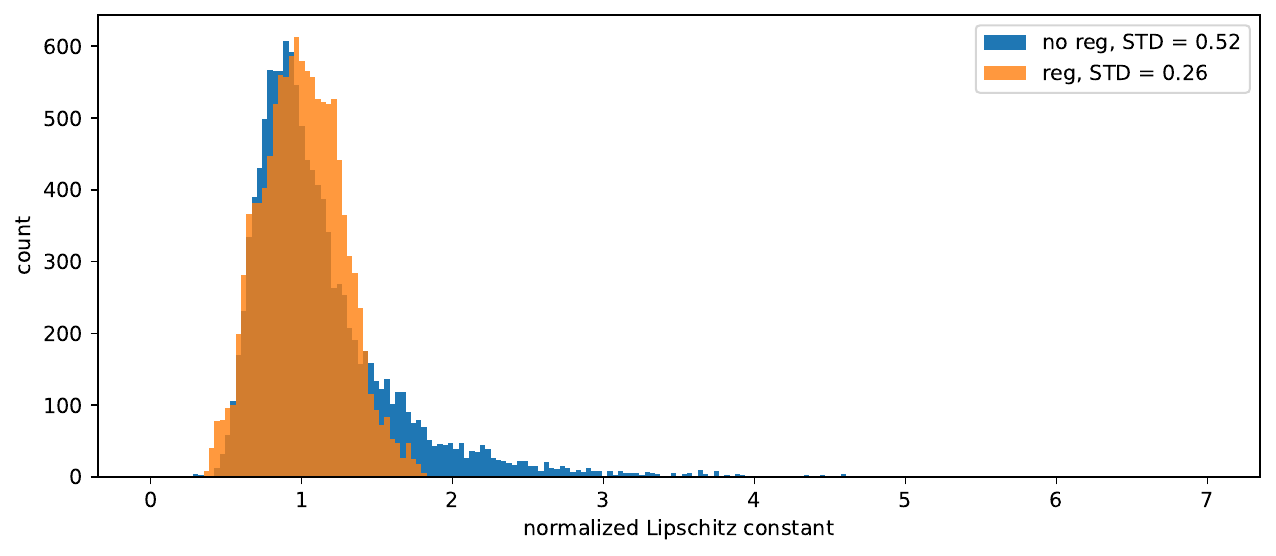}
    \caption{Distribution of normalized Jacobian spectral norm, or local Lipschitz constant.}
    \label{fig4:airfoil_lipschitz}
\end{figure}

Solving the GLV problem should also derive an airfoil representation against which the lift and drag coefficients change more \emph{uniformly}. Specifically, let $f:X\times C\to Y$ denote the airfoil performance function, we should expect $f(g^x_\theta(z), c)\approx g^y_\theta(z, c)$ to have more similar local (best) Lipschitz constant \wrt $z\in\gZ$ everywhere. This is because $g^y_\theta(z, c)$ is constrained to be $K$-Lipschitz \wrt $z$ at all $c$, while it is also encouraged to not have unnecessarily low local Lipschitz constant, as otherwise this will stretch $\gZ$ and increase the volume penalty. To verify this, we can inspect the distribution of the Jacobian $\nabla_z  g^y_\theta(z, c)$'s spectral norm $\|\nabla_z g^y_\theta(z, c)\|_2$ at different $z \in \gZ$ and $c$, as this equals the local Lipschitz constant when $g^y_\theta$ is differentiable. Since LV does not come with the performance surrogate $g^y_\theta$ in its formulation, for comparing LV's result to GLV's result, we additionally train a separate surrogate to approximate $f({g}_\theta(z), c)$ for the LV decoder ${g}_\theta: Z\to X$. Figure~\ref{fig4:airfoil_lipschitz} shows the distribution of the spectral norms normalized by their median, \ie, $\|\nabla_z g^y_\theta(z, c)\|_2 / \textrm{med}(\|\nabla_z g^y_\theta(z, c)\|_2)$. It is obvious that GLV has more values concentrated around 1, and thus our anticipation is verified. 

\subsubsection{Optimization}
\begin{figure}[bt!]
    \centering
    \begin{subfigure}[b]{0.8\textwidth}
         \centering
         \includegraphics[width=\textwidth]{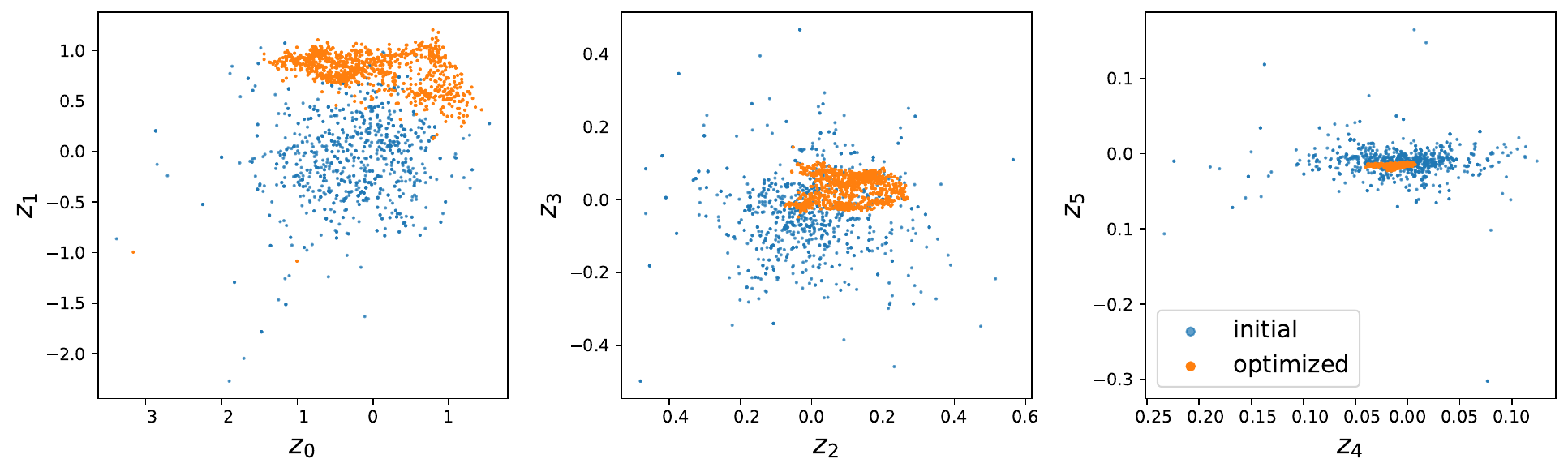}
         \caption{GLV: Latent codes before and after optimization}
         \label{fig4:airfoil_opt_code_glv}
     \end{subfigure}
     \begin{subfigure}[b]{\textwidth}
         \centering
         \includegraphics[width=\textwidth]{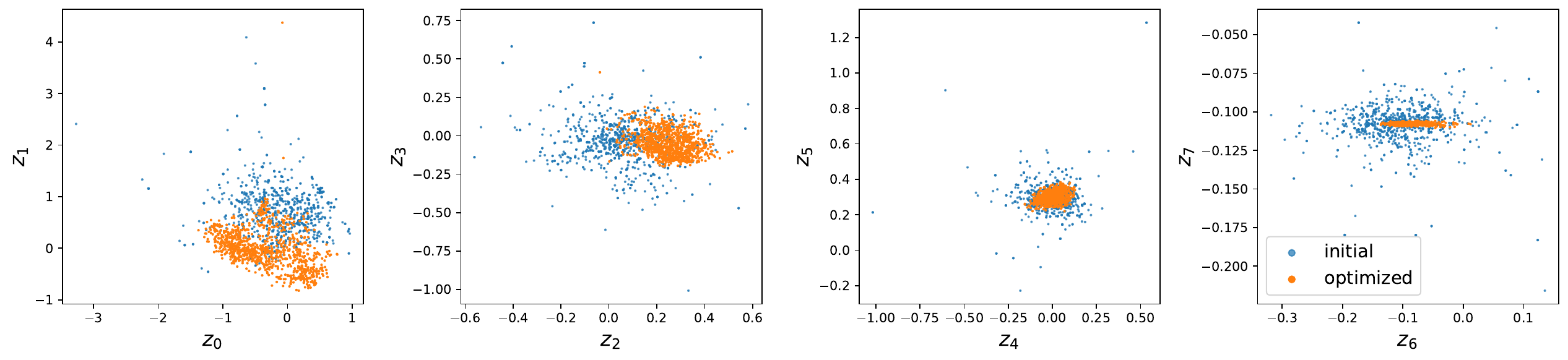}
         \caption{LV: Latent codes (scaled) before and after optimization}
         \label{fig4:airfoil_opt_code_lv}
     \end{subfigure}
     \begin{subfigure}[b]{0.45\textwidth}
         \centering
         \includegraphics[width=\textwidth]{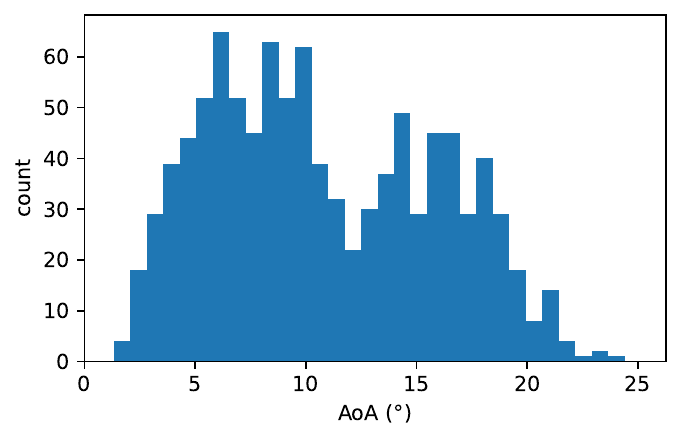}
         \caption{GLV: AoAs after optimization}
         \label{fig4:airfoil_opt_aoa_glv}
     \end{subfigure}
     \hspace{1cm}
     \begin{subfigure}[b]{0.45\textwidth}
         \centering
         \includegraphics[width=\textwidth]{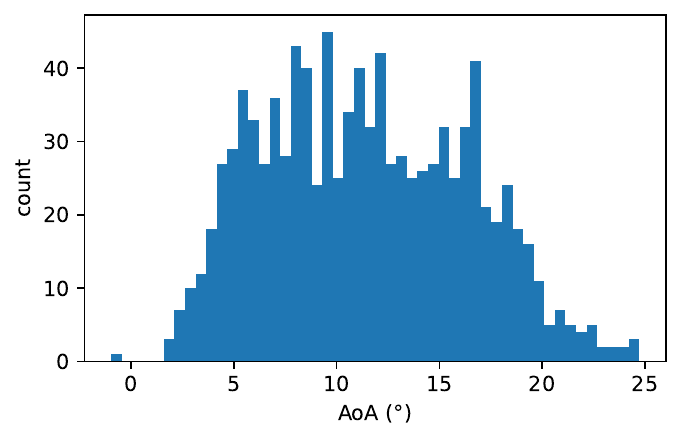}
         \caption{LV: AoAs after optimization}
         \label{fig4:airfoil_opt_aoa_lv}
     \end{subfigure}
     \begin{subfigure}[b]{\textwidth}
         \centering
         \includegraphics[width=\textwidth]{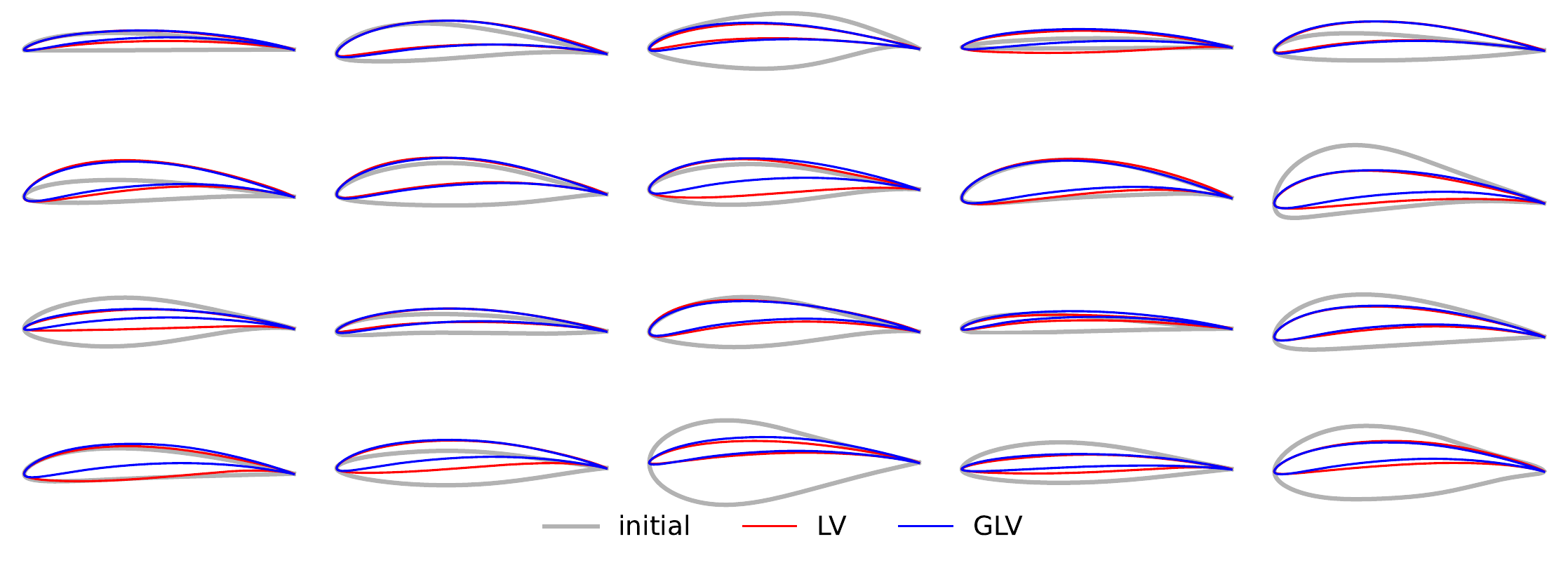}
         \caption{Airfoil shapes before and after optimization}
         \label{fig4:airfoil_opt_shape}
     \end{subfigure}
    \caption{Airfoil design variables before and after optimization.} 
    \label{fig4:airfoil_opt_var}
\end{figure}

\begin{figure}
    \centering
    \includegraphics[width=0.8\linewidth]{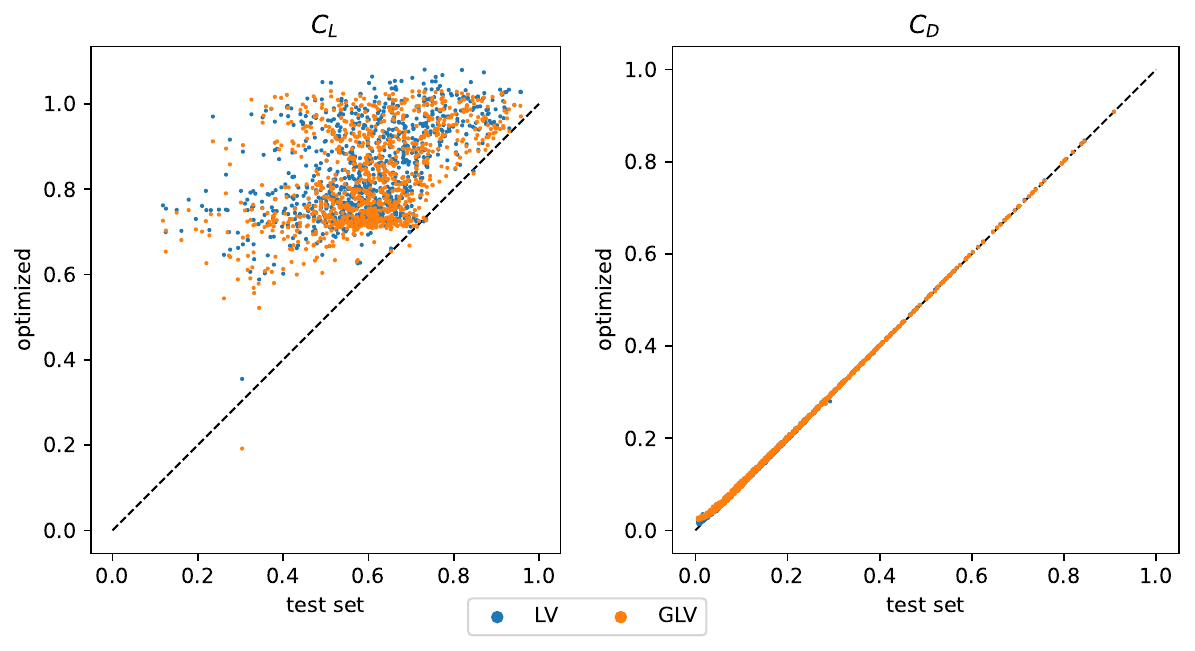}
    \caption{$C_L$ and $C_D$ after optimization.}
    \label{fig4:airfoil_opt_perf}
\end{figure}

The last property just shown in \S\ref{sec:procrustes} might be beneficial to gradient-based optimization in the latent space, as intuitively it makes the performance function $f$ less rugged \wrt $z$ and thus should help stabilize the optimization. Since it is nontrivial to integrate machine learning models into conventional CFD toolkits like XFOIL and MACH-Aero, we instead use the surrogates $g^y_\theta(z, c)$ to approximately investigate this assumption and leave the more complete CFD-toolkit-based investigation to future work. 

For each of the LV and GLV autoencoders, given the surrogate $g^y: z\times c \mapsto y$ where $c = a\times m \times r$ consists of AoA($a$), Ma($m$) and Re($r$) and $y = C_L\times C_D$, together with the specified boundary conditions $m$ and $r$ and drag constraint $c_d$, we formulate the test optimization problem as follows to maximize the lift/drag ratio in the low dimensional latent space $Z$:
\begin{align}
    \argmax_{z\in Z, a\in\sR} \quad& C_L = g^y_1(z, a, m, r) 
    \\
    \text{s.t.} \quad& C_D = g^y_2(z, a, m, r) = c_d
\end{align}
For the ease of implementation, we use penalty method to turn it into an unconstrained problem with Tikhonov regularization:
\begin{align}
    \argmax_{z\in Z, a\in\sR} \quad& g^y_1(z, a, m, r) + w_1(g^y_2(z, a, m, r) - c_d)^2 + w_2\cdot \log p(z)
\end{align}
where we set $w_1 = 100$, $w_2 = 0.02$ and let $p(z)$ be the probability density function of the diagonal Gaussian approximation of $\gZ$, which, when combined with $\log$, leads to a Tikhonov regularization term $\propto \|D (z-\mu(\gZ))\|_2^2$, where $D$ is a diagonal matrix whose $i$th diagonal element equals $\sigma^{-1}_i(\gZ)$. This $\log p$ regularization helps keep $z$ within or at least close to $\gZ$, thus preventing $z$ from converging to airfoil codes far away from $\gZ$ that correspond to invalid airfoil designs. We employ the Adam optimizer with a learning rate of $0.01$ for both LV and GLV representation, and to make the learning rate equally applicable in both cases, we scale up the LV latent set by the scaling factor $s=3.5$ retrieved via Procrustes analysis, and accordingly make LV's surrogate become $g^y_\text{new}(z, a, m, r) = g^y(z/s, a, m, r)$, so that both the LV and GLV latent sets have the same length scale. For each test case $(x, a, m, r, c_l, c_d)$ drawn out of the test set, we let LV and GLV retrieve the airfoil $x$'s latent codes respectively as their optimizations' initial $z$, and set the initial values of all AoAs $a$ to $10^\circ$. We optimize all the 1000 test cases for 1000 iterations.

The optimized design variables $z$ and $a$ along with the corresponding airfoils retrieved by the decoder $g^x_\theta(z)$ are displayed in Fig.~\ref{fig4:airfoil_opt_var}. We can see that in many cases, the airfoils are able to morph their shapes dramatically from the initial designs throughout the optimization. This is different from our experience in traditional airfoil optimizations without machine learning, in which the CFD optimizer usually focuses much more on changing the AoAs of airfoils, such that the optimized shapes look similar to the initial shapes. Figure~\ref{fig4:airfoil_opt_perf} further compares the predicted $C_L$ and $C_D$ of optimized airfoils with the values in the test set. It indicates that the optimization for both LV and GLV can improve $C_L$ greatly while having the constraint $C_D = c_d$ satisfied.

\begin{figure}[bt!]
    \centering
     \begin{subfigure}[b]{0.32\textwidth}
        \includegraphics[width=\linewidth]{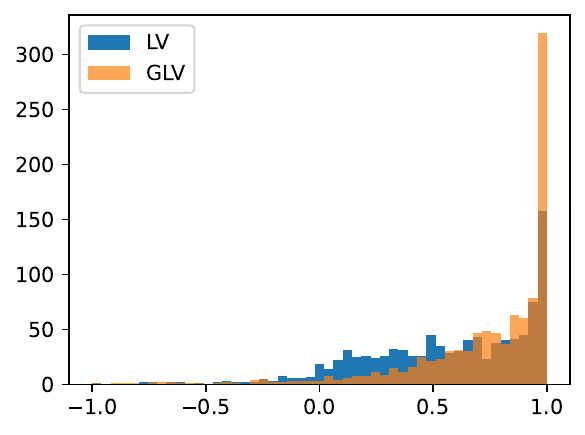}
        \caption{Spearman's rank correlation}
    \label{fig4:airfoil_opt_spr}
    \end{subfigure}
    \begin{subfigure}[b]{0.32\textwidth}
        \includegraphics[width=\linewidth]{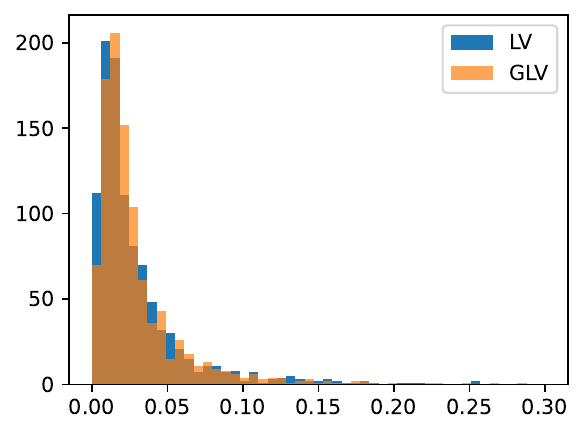}
        \caption{Cumulative regret}
    \label{fig4:airfoil_opt_regret}
    \end{subfigure}
    \begin{subfigure}[b]{0.32\textwidth}
        \includegraphics[width=\linewidth]{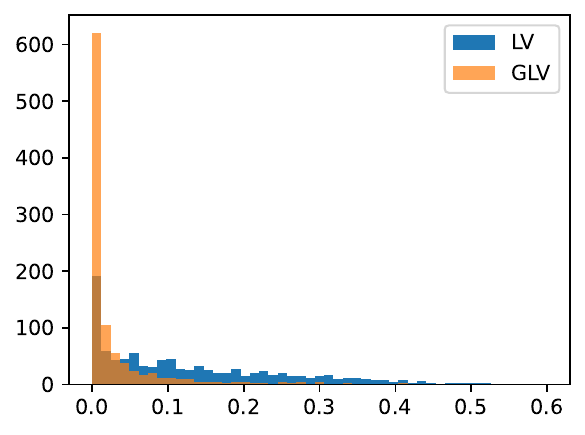}
        \caption{Mean deviation}
    \label{fig4:airfoil_opt_meandev}
    \end{subfigure}
    \caption{Histograms of the Spearman's rank correlation, cumulative regret and mean deviation of 1000 test cases' optimization trajectories.} 
    \label{fig4:airfoil_opt_curve}
\end{figure}

To compare the LV and GLV design representations in terms of their influence on airfoil optimization, and see if the GLV representation is superior in some aspects, we investigate and characterize their \emph{optimization histories} numerically\textemdash \ie, the objective's value $y_i$ against the optimization iteration $i$. We focus on three aspects:
\begin{enumerate}
    \item \emph{Stability}: Whether most of the optimization curves are monotonically increasing during most iterations. We quantify this aspect using the \emph{Spearman's rank correlation} between the objective value and the iteration number~\citep{spearman1961proof}. The more a curve is monotonically increasing, the closer its Spearman's correlation is to 1. The histogram in Fig.~\ref{fig4:airfoil_opt_spr} verifies that the airfoil optimizations of GLV are stabler, as its Spearman's correlations concentrate much more around 1.
    
    \item \emph{Convergence Rate}: How fast the optimization converges. We quantify it using \emph{cumulative regret}, which is defined as $\frac{1}{N}\sum^N_{i=1} (y^\star - y_i)$ where $y_i$ refers to the objective value at iteration $i$. In other words, it is the average area enclosed by the optimization curve and the horizontal line at the optimal objective value. The smaller the better. Figure~\ref{fig4:airfoil_opt_regret} shows that both LV and GLV have comparable cumulative regrets, so GLV does not bring any advantage in convergence rate to our test cases. 

    \item \emph{Robustness}: How sensitive the optimization's final solution is to the initial guess. To investigate this, we repeat the same 1000 optimization test cases for 5 times, each time we keep all the variables and parameters fixed except resampling the dataset for a different airfoil to obtain a new initial latent code $z$. After the optimizations, for each test case we can get 5 different resulting variables $(z, a)_i$. Then for each case, we first obtain the mean $(\bar{z}, \bar{a})$ of the 5 solutions, then evaluate the $L_2$ deviation $\epsilon_i = \|(z, a)_i - (\bar{z}, \bar{a})\|_2$, and finally evaluate the mean of $d_i$ as the \emph{mean deviation} to quantify how similar these 5 solutions are to each other. The distributions of these 1000 mean deviations in Fig~\ref{fig4:airfoil_opt_meandev} show that GLV's representation can greatly improve the optimization robustness compared to LV.
\end{enumerate}

In summary, the integration of performance information into airfoil's LV problem may help derive a latent representation that can improve the optimization tasks' stability and robustness by making the objective function less rugged. However, this result has to be re-verified with actual CFD solvers after our machine learning models are integrated in the toolkit.

    


\section{Conclusion}
This work introduces \emph{Least Volume} regularization for \emph{continuous} autoencoders. By formulating the autoencoder training as a optimization problem \wrt the \emph{volume preorder}, it both compresses the latent dimension to the dataset's embedding dimension while preserving the dataset's topological structure, and makes the latent dimensions indicate their importance via their standard deviations\textemdash similar to PCA.
We further extended the LV problem to \emph{Generalized Least Volume} problem, which is applicable to both labeled datasets and datasets defined over non-Euclidean metric spaces. Unlike LV, it helps incorporate non-Euclidean information into the latent representation, so that the importance of each data dimension\textemdash\wrt the new distance metric or label information\textemdash can be better understood. Alongside the theoretical formulations, we developed the \emph{Dynamic Pruning} algorithm to effectively solve these least volume problems.

Through theoretical analysis, we showed that minimizing the volume is equivalent to minimizing an upper bound of the latent code's generalized variance, and that using a $K$-Lipschitz decoder prevents the latent set's isotropic shrinkage. This Lipschitz constraint further guarantees the reliability of the Dynamic Pruning algorithm, making the training seamless. Moreover, we proved that PCA is a linear special case of the least volume problem, which suggests that a similar importance ordering effect should be observable on the nonlinear autoencoders regularized by LV or GLV. 

We conducted several experiments to validate the effectiveness and properties of Least Volume. The (na\"ive) volume penalty achieves better dimension reduction results than several $L_1$ distance-based counterparts on multiple benchmark image datasets. It was also verified that the standard deviation of each latent dimension indicates its importance in terms of its contribution to reducing the reconstruction error. Furthermore, the dynamic pruning algorithm outperforms the na\"ive volume reduction thanks to its construction. Leveraging knowledge from Topology and the great dimension reduction capability of least volume autoencoders trained with dynamic pruning, we reveal insights into various important aspects of representation learning. We list the key findings below:

\begin{itemize}
    \item Reducing the autoencoder's latent space dimension to the dataset's embedding dimension is beneficial to sampling and analyzing the latent set (or equivalently the dataset), especially when the dataset's intrinsic dimension equals the embedding dimension. 
    
    \item ``Disentangled representation'' is just a high dimensional version of strict monotonicity, and can appear even in an ordinary autoencoder without regularization, as long as it is continuous, injective, and has the right latent dimension that allows us to inspect the cross section of the latent set (\eg, the latent dimension equals the dataset's intrinsic dimension). 
    
    \item Different datasets have different topological complexities. For instance, the MNIST dataset consists of intersecting balls of different intrinsic dimensions, the UIUC airfoil dataset is a $7\sim8$ dimensional ball, and the CelebA dataset lives on a sphere of $100\sim 110$ dimensions. This topological difference might account for the varying challenges faced in different tasks.

    \item Regarding labeled datasets, the GLV exhibits an effect similar to contrastive learning, separating data samples with different labels while bringing together those with the same label. When performed on the MNIST dataset of categorical labels, different digits form different clusters in the latent space, and this effect persists even when 50\% of the labels are missing. When applied to the UIUC dataset of continuous labels (lift and drag coefficient), it results in a low dimensional airfoil representation where small perturbations lead to limited variation in fluid dynamics performance, which stabilizes adjoint optimizations conducted on it. 
\end{itemize}

It should be stressed that the LV and GLV formulations are established on the bridge between \emph{continuous} autoencoders and homeomorphisms and the manifold hypothesis on many datasets. These assumptions fail in two primary cases, making LV and GLV inapplicable:
\begin{itemize}
    \item \textit{When the autoencoder is discontinuous.} 
    For instance, in recent years VQ-VAE~\citep{van2017neural} has attracted a lot of attention. Though it is constructed with continuous encoder and decoder, its encoding operation is in general discontinuous, because mapping the continuous encoder's output to the nearest embedding vector is a discontinuous operation in most cases. Specifically, any finite non-singleton set of embedding vectors is discrete and thus totally disconnected. Because continuous functions preserve connectedness, if a connected subset of a dataset that satisfies Assumption~\ref{asp:smooth_manifold} is mapped onto more than one embedding vectors, the encoding must be discontinuous.
    Another common example is the integration of discontinuous layers, such as discontinuous activation functions. 

    \item 
    \textit{When no subset of the dataset is a manifold.} 
    If the dataset is defined in a discrete space then it cannot satisfy Assumption~\ref{asp:smooth_manifold}. This also holds for the product of discrete spaces. For instance, the dataset of human language sentences for training LLMs is not a manifold. It is essentially a subset of the infinite product of discrete vocabulary spaces, which is totally disconnected and thus cannot be locally Euclidean\textemdash required by being a manifold. The topological dimension in these cases might become zero or infinite, which makes the LV dimension reduction pointless.
\end{itemize}

It is intriguing to develop equivalent representation compression methods for these constrained settings in the future. Meanwhile, we should keep these caveats in mind and carefully make assumptions about the datasets when applying Least Volume.


\acks{This work was conducted while the authors were at the University of Maryland, College Park, USA. We acknowledge the support from the National Science Foundation through award \#1943699 as well as ARPA-E award DE-AR0001216. We also appreciate Prof.~Peter W. Chung's support with computational resources.}


\newpage

\appendix








\vskip 0.2in
\bibliography{sample}

\end{document}